\documentclass[letterpaper,twocolumn,10pt]{article}
\usepackage{usenix-2020-09}

\usepackage{algorithm}
\usepackage[noend]{algorithmic}
\usepackage{subfloat}
\usepackage{amsfonts}
\usepackage{amsthm}
\usepackage{booktabs}
\usepackage{multirow}
\usepackage{multicol}
\usepackage{subfigure}
\usepackage{comment}
\usepackage{color}
\usepackage{xcolor}

\usepackage[shortlabels]{enumitem}
\usepackage{thm-restate}

\usepackage{graphicx}
\usepackage{overpic} 
\usepackage{amsmath} 
\usepackage{bm}      

\usepackage{diagbox}
\usepackage{cite}
\usepackage{float}
\usepackage{textpos}
\usepackage{bbm}
\usepackage{soul}

\usepackage{fancyhdr}

\newcommand{\name}{Inf$^2$Guard}

\newtheorem{definition}{Definition}
\newtheorem{lemma}{Lemma}

\begin{document}

\title{Inf$^2$Guard: An Information-Theoretic Framework for Learning Privacy-Preserving Representations against Inference Attacks}

\author{
{\rm Sayedeh Leila Noorbakhsh$^{1,*}$, Binghui Zhang$^{1,*}$, Yuan Hong$^2$, Binghui Wang$^1$}\\
$^1$Illinois Institute of Technology, $^2$University of Connecticut, $^*$Equal contribution
}

\maketitle
\begin{abstract}
Machine learning (ML) is vulnerable to inference (e.g., membership inference, property inference, and data reconstruction) attacks that aim to infer the private information of training data or dataset. Existing defenses are only designed for one specific type of attack and sacrifice significant utility or are soon broken by adaptive attacks. We address these limitations by proposing an information-theoretic defense framework, called \texttt{\name}, against the three {major} types of inference attacks. 
Our framework, inspired by the success of representation learning, posits that learning shared representations not only saves time/costs but also benefits numerous downstream tasks.  
Generally, \texttt{\name} involves two mutual information objectives, 
for privacy protection and utility preservation, respectively. 
\texttt{\name} exhibits many merits: it facilitates the design of customized objectives against the specific inference attack; it provides a general defense framework which can treat certain existing defenses as special cases; and importantly, it aids in deriving theoretical results, e.g., inherent utility-privacy tradeoff and guaranteed privacy leakage.  
Extensive evaluations validate the effectiveness of \texttt{\name} for learning privacy-preserving representations against inference attacks and demonstrate the superiority over the baselines.\footnote{Source code: \url{https://github.com/leilynourbakhsh/Inf2Guard}.}

\end{abstract}

\section{Introduction}
\label{sec:intro}

Machine learning (ML) models (particularly deep neural networks) are vulnerable to inference attacks, which aim to infer sensitive information about the training data/dataset that are used to train the models. There are three well-known types of inference attacks on training data/dataset: membership inference attacks (MIAs)~\cite{shokri2017membership,yeom2018privacy,carlini2022membership}, property inference attacks (PIAs) (also called distribution inference attacks)~\cite{ateniese2015hacking,ganju2018property,suri2022formalizing}, and data reconstruction attacks (DRAs) (also called model inversion attacks)~\cite{hitaj2017deep,balle2022reconstructing}. 
Given an ML model, in MIAs, an adversary aims to infer whether a particular data sample was in the training set, while in PIAs, an adversary aims to infer statistical properties of the 
training dataset used to train the targeted ML model. 
Furthermore, an adversary aims to directly reconstruct the training data in DRAs. 
Leaking the data sample or information about the dataset raises serious privacy issues. 
For instance, by performing MIAs, an adversary is able to identify users included in sensitive medical datasets, which itself is a privacy violation~\cite{homer2008resolving}. 
By performing PIAs, an adversary can determine whether or not machines that generated the bitcoin logs were patched for Meltdown and Spectre attacks~\cite{ganju2018property}. 
More seriously, DRAs performed by an adversary leak all the information about the training data.

To mitigate the privacy risks, various 
defenses have been proposed against MIAs~\cite{srivastava2014dropout,shokri2017membership,salem2018ml,nasr2018machine,jia2019memguard,song2021systematic,shejwalkar2021membership,xu2022neuguard} and DRAs~\cite{pathak2010multiparty,hamm2016learning,wei2020federated,zhu2019deep,sun2020provable,gao2021privacy,lee2021digestive,scheliga2022precode}\footnote{To our best knowledge, there exist no effective defenses against PIAs. 
\cite{hartmann2023distribution} analyzes sources of information leakage to cause PIAs, but their solutions are difficult to be tested on real-world datasets due to lack of generality.}. However, there are two fundamental limitations in existing defenses: 1) They are designed against only one specific type of attack; 
2) Provable defenses (based on differential privacy~\cite{dwork2006differential,abadi2016deep}) incur significant utility losses to achieve reasonable defense performance against inference attacks~\cite{jayaraman2019evaluating,shejwalkar2021membership} since {the design of} such randomization-based defenses did not consider specific inference attacks (also see Section~\ref{sec:eval}); and empirical defenses are soon broken by stronger/adaptive attacks~\cite{choquette2021label,song2021systematic,balunovic2022bayesian}.

We aim to address these limitations and consider the 
question: 
\emph{1) Can we design a unified privacy protection framework against these inference attacks, that also maintain utility? 2) Under the framework, can we further theoretically understand the utility-privacy tradeoff and the privacy leakage against the inference attacks?} 
To this end, we propose an information-theoretic defense framework, termed {\texttt{\name}}, against inference attacks 
through the lens of \emph{representation learning}~\cite{bengio2013representation}. 
Representation learning has been one of the biggest successes in modern ML/AI so far (e.g., it plays an important role in today's large language models such as ChatGPT~\cite{chatgpt:online} and PaLM2~\cite{palm2:online}). 
Particularly, rather than training large models from scratch, which requires huge computational costs and time  (e.g., GPT-3 has 175 billion parameters),
learning shared representations (or pretrained encoder)\footnote{
\emph{Pretrained encoder as a service} has been widely deployed by industry, e.g., OpenAI's GPT-4 API~\cite{chatgpt:online} and Clarifai's Embedding API~\cite{clarifai_demo}. 
We will interchangeably use the pretrained encoder and learnt representations.} presents an economical alternative. For instance, the shared representations can be directly used or further fine-tuned with different purposes, achieving considerable savings in time and cost. 

More specifically, we formulate \texttt{\name} 
via two mutual information (MI)\footnote{In information theory, MI is a measure of shared information between random variables, and offers a metric to quantify the ``amount of information" obtained about one random variable by observing the other random variable.} 
objectives in general, for privacy protection and utility preservation, respectively. 
Under this framework, we can design customized MI objectives to defend against each inference attack. 
For instance, to defend against MIAs, we design one MI objective to 
learn representations that contain as less information as possible about the membership of the training data---thus protecting membership privacy, while the other one to ensure the learnt representations include as much information as possible about the training data labels---thus maintaining utility. 
However, directly solving the MI objectives for each inference attack is challenging, since calculating an MI between arbitrary variables is often infeasible \cite{peng2018variational}. 
To address it, 
we are inspired by the MI neural estimation~\cite{alemi2017deep,belghazi2018mutual,oord2018representation,poole2019variational,hjelm2019learning,cheng2020club}, which transfers the intractable MI calculations to the tractable variational MI bounds. Then, we are capable of parameterizing each bound with a (deep) neural network, and train neural networks to approximate the true MI and learn representations against the inference attacks. 
Finally, we can derive theoretical results based on our MI objectives: we obtain an inherent utility-privacy tradeoff, and guaranteed privacy leakage against each inference attack. 

We extensively evaluate \texttt{\name} and compare it with the existing defenses against the inference attacks on multiple benchmark datasets. Our experimental results validate that \texttt{\name} obtains a promising utility-privacy tradeoff and significantly outperforms the existing defenses. For instance, under the same defense performance against MIAs, \texttt{\name} has a 30\% higher testing accuracy than the DP-SGD~\cite{abadi2016deep}. Our results also validate the privacy-utility tradeoffs obtained by  \texttt{\name}\footnote{A recent work \cite{salem2023sok} formulates defenses against inference attacks under a privacy game framework, but it does not propose concrete defense solutions.}. 

Our main contributions are summarized as below:
\begin{itemize}
\vspace{-2mm}
    \item {\bf Algorithm:} We design the first unified framework \texttt{\name} to defend against  the three well-known types of inference attacks via information theory. Our framework can 
 instantiate many existing defenses as special cases, e.g., AdvReg~\cite{nasr2018machine} against MIAs (See Section~\ref{sec:MI_MIA}) and Soteria~\cite{sun2020provable} against DRAs (See Section~\ref{sec:MI_DRA}). 
\vspace{-2mm}
    \item {\bf Theory:} Based on our formulation, we can derive novel theoretical results, e.g., the inherent tradeoff between utility and privacy, and guaranteed privacy leakage against all the considered inference attacks. 
\vspace{-2mm}
    \item {\bf Evaluation:} Extensive 
    evaluations verify the effectiveness of 
    \texttt{\name}  
    for learning privacy-preserving representations against inference attacks.
\end{itemize}

\section{Background and Problem Definition}
\label{sec:background}

\noindent {\bf Notations:} We use ${s}$, ${\bf s}$, {\bf S}, and $\mathcal{S}$ to denote (random) scalar, vector, matrix, and space, respectively. Accordingly, $\textrm{Pr}(s)$, $\textrm{Pr}({\bf s})$, and $\textrm{Pr} ({\bf S})$ are the probability distribution over $s$,  ${\bf s}$, and ${\bf S}$. 
$ I({\bf x}; {\bf r})$ and $H({\bf x}, {\bf r})$ are the mutual information and cross entropy between a pair of random variables $({\bf x}, {\bf r})$, respectively, and $H({\bf x}) = I({\bf x}; {\bf x})$ as the entropy of ${\bf x}$.
${KL}(p || q)$ is the KL-divergence between two distributions $p$ and $q$.  
We denote $\mathcal{D}$ as the underlying  distribution that data are sampled from. 
A data sample is denoted as $({\bf x}, {y}) \sim \mathcal{D}$, where ${\bf x} \in \mathcal{X}$ is data features, ${y} \in \mathcal{Y}$ is the label, and $\mathcal{X}$ and $\mathcal{Y}$ are the data space and label space, respectively. 
We further denote a dataset as $D= \{{\bf X}, {\bf y}\}=\{({\bf x}_i, {y_i})\}$, 
that consists of a set of data samples $({\bf x}_i, {y_i}) \sim \mathcal{D}$, and 
will interchangeably
use $D$ and $\{{\bf X}, {\bf y}\}$. 
We let $u \in \mathcal{U}$ be the private attribute within the attribute space $\mathcal{U}$. For instance, in MIAs, $u \in \mathcal{U}=\{0,1\}$ means a binary-valued private membership; in PIAs, $u \in \mathcal{U}=\{1,2,\cdots, K\}$ indicates a $K$-valued private dataset property; and $u \in \mathcal{U}=\mathcal{X}$ indicates the private data itself in DRAs. 
The composition function 
of two functions $f$ and $g$ is denoted as $(g\circ f) (\cdot) = g(f({\cdot}))$. 

\subsection{Formalizing Privacy Attacks}
\label{sec:bg-privattk}

\noindent We denote a classification model\footnote{In this paper, we focus on classification models for simplicity. } 
$F_\theta: \mathcal{X} \rightarrow \mathcal{Y}$ 
as a function, parameterized by $\theta$, 
that maps a data sample ${\bf x} \in \mathcal{X}$ to a label $y \in \mathcal{Y}$.  Given a training set $D \sim \mathcal{D}$, we denote $F \leftarrow \mathcal{T}(D)$ as learned by running a training algorithm $\mathcal{T}$ on 
the dataset $D$. 

\vspace{+0.02in} 
\noindent  {\bf Formalizing MIAs:}
Assume a data sample $({\bf x}, y) \sim \mathcal{D}$  
with a private membership $u$ that is chosen uniformly at random from $\{0,1\}$, where $u=1$ means $({\bf x}, y)$ is a member of 
$D$, and 0 otherwise. 
An MIA ${A}_{MIA}$ has access to $\mathcal{D}$ and $F$, takes $({\bf x}, y)$ as input,  
and outputs a binary ${A}_{MIA}^{\mathcal{D},F}({\bf x},y)$. We omit $\mathcal{D},F$ for notation simplicity.  
Then, the attack performance of an MIA ${A}_{MIA}$ is defined as $\textrm{Pr}_{({\bf x}, y, u)}({A}_{MIA}({\bf x},y) = u)$.

\vspace{+0.02in} 
\noindent {\bf Formalizing PIAs:} 
PIAs define a private property on a dataset. 
Given a dataset $D_u \sim \mathcal{D}$ with a private property $u$ chosen uniformly at random from $\{1, 2, \cdots, K\}$. 
A PIA ${A}_{PIA}$ has access to $\mathcal{D}$ and $F$, 
and outputs a $K$-valued ${A}_{PIA}(D_u)$.
Then, the attack performance of a PIA ${A}_{PIA}$ is defined as $\textrm{Pr}_{(D_u,u)}({A}_{PIA}(D_u) = u)$.  

\vspace{+0.02in} 
\noindent {\bf Formalizing DRAs:} 
Given a random data $({\bf x},y) \in D$, DRAs aim to reconstruct the private ${\bf x}$. 
A DRA ${A}_{DRA}$ 
has access to $\mathcal{D}$ and $F$, and outputs a reconstructed $\hat{\bf x} = {A}_{DRA}({\bf x},y)$. 
The DRA performance is measured by the similarity/difference between $\hat{\bf x}$ and ${\bf x}$.
For instance, \cite{balle2022reconstructing} introduces the $(\eta,\gamma)$-reconstruction metric defined as $\textrm{Pr}_{({\bf x},y)}(\| \hat{\bf x} - {\bf x} \|_2 \leq \eta) \geq \gamma$, where a smaller $\eta$ and a larger $\gamma$ imply a more severe DRA.

\subsection{Threat Model and Problem Formulation}
\label{sec:problem}

{We have three roles: \emph{task learner}, \emph{defender}, and  
\emph{attacker}. The task learner (i.e., data owner) aims to learn an accurate classification model on its training data. 
The defender (e.g., data owner or a trusted service provider)
aims to protect the training data privacy---it designs a
defense framework by learning \emph{shared data representations}
that are robust against inference attacks. 
The attacker can \emph{arbitrarily} use data representations 
to perform the inference attack. 
The attacker is also assumed to know 
the underlying data distribution,  
but cannot access the internal encoder (e.g., deployed as an API~\cite{chatgpt:online,clarifai_demo}). 
}

Formally, we denote 
$f_\Theta: \mathcal{X} \rightarrow \mathcal{Z}$
as the \emph{encoder},
parameterized by $\Theta$,  
that maps a data sample ${\bf x} \in \mathcal{X}$ (or a dataset ${\bf X} \in \mathcal{X}$) to its representation vector ${\bf r} = f({\bf x})\in \mathcal{Z}$ (or representation matrix ${\bf R} = f({\bf X}) \in \mathcal{Z}$), where $\mathcal{Z}$ is the representation space. 
Moreover, we let $C: \mathcal{Z} \rightarrow \mathcal{Y}$ be the \emph{classification model} on top of the representation  ${\bf r}$ or encoder $f$,  which predicts the data  
label $y$ (or dataset labels ${\bf y}$).  
We further let $A: \mathcal{Z} \rightarrow \mathcal{U}$ be the \emph{inference model}, which infers the private attribute $u$ using the learnt  representations ${\bf r}$ or ${\bf R}$. 
Then, our defense goals are: 

\begin{itemize}[leftmargin=*]
\vspace{-2mm}
\item {\bf Defend against MIAs:}
Given a random sample $({\bf x}, y, u) \in \mathcal{D}$, we expect to learn $f$ such that the MIA performance $\textrm{Pr}(A_{MIA}(f({\bf x}), y)=u)$  is low, and the utility loss/risk, i.e., $\textrm{Risk}_{MIA}(C \circ f) = \textrm{Pr}(C \circ f({\bf x}) \neq y)$, is also small.   

\vspace{-2mm}
\item {\bf Defend against PIAs:}
Given a random dataset $({\bf X}, {\bf y}, u) \in \mathcal{D}$, we expect to learn $f$ with low PIA performance $\textrm{Pr}(A_{PIA}(f({\bf X}), {\bf y})=u)$, and also a small  utility loss/risk, i.e., $\textrm{Risk}_{PIA}(C \circ f) = 
\frac{1}{|{\bf y}|}\sum_{({\bf x}, y) \in \{{\bf X}, {\bf y}\}} \textrm{Pr}(C\circ f({\bf x}) \neq y)$.

\vspace{-2mm}
\item {\bf Defend against DRAs:}
Given a random sample $({\bf x}, y) \in \mathcal{D}$, we expect to learn $f$ with low DRA performance, 
i.e., 
$\textrm{Pr}_{({\bf x},y)}(\| \hat{\bf x} - {\bf x} \|_2 \geq \eta) \geq \gamma$ with a large $\eta$ and $\gamma$ ({flipping the inequality direction on $\eta$ for DRAs}), and 
also a small utility risk $\textrm{Risk}_{DRA}(C \circ f) = \textrm{Pr}(C \circ f({\bf x}) \neq y)$.

\vspace{-2mm}
\end{itemize}

\section{Design of \texttt{\name}}
\label{sec:MIGuard}

\subsection{\texttt{\name} against MIAs}
\label{sec:MI_MIA}

\subsubsection{MI objectives}
Given a data sample ${\bf x} \sim \mathcal{D}$, from the training set $D$ (i.e., $u=1$) or not  (i.e., $u=0$), the defender learns the representation ${\bf r}= f({\bf x})$ that satisfies the following two goals: 

\begin{itemize}[leftmargin=*]
\vspace{-2mm}
\item \textbf{Goal 1: Membership protection.} 
${\bf r}$ contains as less information as possible about 
the private membership  
$u$. Ideally, when ${\bf r}$ does not include information about 
$u$ (i.e., ${\bf r} \perp {u}$), it is impossible to infer $u$ from ${\bf r}$. 
Formally, we quantify the membership protection using the MI objective as follows: 
{
\setlength{\abovedisplayskip}{4pt}
\setlength{\belowdisplayskip}{4pt}
\begin{align}
    & \min\limits_f I({\bf r}; u),  \label{eqn:MIA_minpriv}
\end{align}
}%
where we minimize such MI to maximally reduce the correlation between ${\bf r}$ and $u$.

\vspace{-2mm}
\item \textbf{Goal 2: Utility preservation.} 
${\bf r}$ should be effective for predicting the label $y$ of the \emph{training} data (i.e., $u=1$), thus preserving utility. 
Formally, we quantify the utility preservation using the below MI
objective:
{
\setlength{\abovedisplayskip}{4pt}
\setlength{\belowdisplayskip}{4pt}
\begin{align}
    \max\limits_f I(y; {\bf r}|u=1), 
       \label{eqn:MIA_maxutil_known} 
\end{align}
}%
where we maximize such MI to make ${\bf r}$ accurately predict the training data label $y$ during training.  
\vspace{-1mm}
\end{itemize}

\subsubsection{Estimating MI via tractable bounds}

The key challenge of solving the above two MI objectives is that calculating an MI between two arbitrary random variables is likely to be infeasible \cite{peng2018variational}. 
Inspired by the existing 
MI neural estimation methods~\cite{alemi2017deep,belghazi2018mutual,oord2018representation,poole2019variational,hjelm2019learning,cheng2020club}, we convert the intractable exact MI calculations to the tractable variational MI bounds. 
Specifically, we first obtain an MI upper bound for membership protection and an MI lower bound for utility preserving via introducing two auxiliary posterior distributions, respectively. 
Then, we parameterize each auxiliary distribution with a neural network, and approximate the true MI by minimizing the upper bound  and  maximizing the  lower bound through training the involved neural networks. 
{\emph{We emphasize we do not design novel MI neural estimators, but adopt existing ones to assist our MI objectives for learning privacy-preserving representations. Note that, though the estimated MI bounds may not be tight (due to the MI estimators or auxiliary distributions learnt by neural networks)~\cite{cheng2020club,hjelm2019learning}, they have shown promising performance in practice. It is still an active research topic to design better MI estimators that lead to tighter MI bounds (which is orthogonal to this work).}}

\vspace{+0.05in} 
\noindent {\bf Minimizing the upper bound MI in Equation (\ref{eqn:MIA_minpriv}).} 
We adapt the variational upper bound 
proposed in~\cite{cheng2020club}. Specifically, 

{
\vspace{-4mm}
\small
\begin{align*}
        & I({\bf r};u ) \leq I_{vCLUB}({\bf r};u )  
        = \mathop{\mathbb{E}}_{p({\bf r}, u )} [\log q_{\Psi}(u |{\bf r}) ] - \mathop{\mathbb{E}}_{ p({\bf r}) p(u )} [\log q_{\Psi}(u |{\bf r}) ], 
\end{align*}
}%
where 
$q_{\Psi}(u  | {\bf r})$ is an auxiliary posterior distribution of $p(u  | {\bf r})$ needing to satisfy the 
below condition on KL divergence: $KL (p({\bf r}, u ) ||  q_{\Psi} ({\bf r}, u )) \leq KL (p({\bf r})p(u ) || q_{\Psi} ({\bf r}, u ))$.
To achieve this, we
thus 
minimize:

{
\vspace{-4mm}
\small
\begin{align}
    & \min_{\Psi}  KL (p({\bf r}, u ) ||  q_{\Psi} ({\bf r}, u ))
    =\min_{\Psi}  KL (p(u  | {\bf r}) ||  q_{\Psi} (u  | {\bf r})) \notag \\
    & = \min_{\Psi}  \mathop{\mathbb{E}}_{p({\bf r}, u )} [\log p(u | {\bf r})] -  \mathop{\mathbb{E}}_{p({\bf r}, u )} [\log q_{\Psi}(u | {\bf r}))] \notag \\
    & \Longleftrightarrow \max_{\Psi} \mathop{\mathbb{E}}_{p({\bf r}, u )} [\log q_{\Psi}(u | {\bf r})],  \label{eqn:KL_conv} 
\end{align}
}%
where we note that 
$\mathop{\mathbb{E}}_{    p({\bf r}, u )} [\log p(u | {\bf r})]$ is irrelevant to $\Psi$.   
{\cite{cheng2020club} proved when $q_\Psi(u|r)$ is parameterized by a neural network with high expressiveness (e.g., deep neural network), the condition 
is satisfied almost surely by maximizing Equation (\ref{eqn:KL_conv}).  
} 
Finally, our {\bf Goal 1} for privacy protection 
is reformulated as solving the below 
\emph{min-max} objective function: 
\begin{align}
        \min \limits_f \min \limits_{\Psi}  I_{vCLUB}({\bf r};u ) 
        \Longleftrightarrow & \min \limits_f \max \limits_{\Psi} 
        \mathop{\mathbb{E}}_{    p({\bf r}, u )} [\log q_{\Psi}(u | {\bf r})]
        \label{eqn:prot_MIA}
\end{align}
\emph{Remark.} 
Equation (\ref{eqn:prot_MIA}) can be interpreted as an \emph{adversarial game} between an adversary $q_{\Psi}$ (i.e., a membership inference classifier) 
who aims to infer the membership $u$ from 
${\bf r}$; and the encoder $f$ who aims to protect 
$u$  from being inferred. 

\vspace{+0.05in} 
\noindent {\bf Maximizing the lower bound MI in Equation (\ref{eqn:MIA_maxutil_known}).} We adopt the 
MI estimator proposed in~\cite{nowozin2016f} to estimate the lower bound of Equation (\ref{eqn:MIA_maxutil_known}).
Specifically, we have 

{
\vspace{-4mm}
\small
\begin{align*}
    & I(y ; {\bf r} | u=1 ) 
    = H(y|u=1) - H(y | {\bf r}, u=1) \notag \\ 
    & = H(y|u=1) + \mathop{\mathbb{E}}_{p(y , {\bf r}, u)} [\log p(y | {\bf r}, u=1))] \notag \\
    & = H(y|u=1) + \mathop{\mathbb{E}}_{p(y , {\bf r}, u)} [\log q_\Omega (y | {\bf r}, u=1))] \notag \\
    & \qquad + \mathop{\mathbb{E}}_{p(y , {\bf r}, u)} [KL(p(\cdot|{\bf r}, u=1) || q_\Omega (\cdot | {\bf r}, u=1))]  \notag \\
    & \geq H(y|u=1) + \mathop{\mathbb{E}}_{p(y , {\bf r}, u)} [\log q_\Omega (y | {\bf r}, u=1))],  
\end{align*}
}%
where $q_{\Omega}$ is an \emph{arbitrary} auxiliary posterior distribution that aims to 
accurately 
predict the training data label $y$ from the representation  ${\bf r}$. 
Hence, our {\bf Goal 2} for utility preservation 
can be rewritten as the following \emph{max-max} objective function:

{
\vspace{-4mm}
\small
\begin{align}
    & \max_f I(y ; {\bf r}| u=1) 
    \Longleftrightarrow
    \max_f \max_{\Omega} \mathop{\mathbb{E}}_{p(y , {\bf r}, u)} \left[\log {q_{\Omega}}[(y  | {\bf r}, u=1) \right]
    \label{eqn:max_jsd_MIA}
\end{align} 
}%
\emph{Remark.} 
Equation (\ref{eqn:max_jsd_MIA}) can be interpreted as a \emph{cooperative game} between the encoder $f$ and 
$q_{\Omega}$ (e.g., a label predictor) that aims to preserve the utility collaboratively. 

\vspace{+0.05in} 
\noindent {\bf Objective function of \texttt{\name} against MIAs.} By combining Equations (\ref{eqn:prot_MIA}) and (\ref{eqn:max_jsd_MIA}), 
 our objective function of learning privacy-preserving representations against MIAs is: 
 
 {
 \vspace{-4mm}
\small 
\begin{align}
    & \max_f\Big(\lambda  \min_{\Psi} -\mathop{\mathbb{E}}_{p({\bf x}, u)}\left[\log q_{\Psi}(u  | f({\bf x} )) \right] 
    \notag \\ & \quad 
    + (1-\lambda ) \max_{\Omega} \mathop{\mathbb{E}}_{p({\bf x}, y, u)} \left[\log {q_{\Omega}}(y | f({\bf x}), u=1)\right] \Big), \label{eqn:task_known_MI_Guard} 
\end{align}
}%
where $\lambda  \in [0,1]$ tradeoffs privacy and utility. That is, a larger $\lambda $ indicates a stronger membership privacy protection, while a smaller $\lambda $ indicates a better utility preservation.

\subsubsection{Implementation in practice}
In practice, 
we solve Equation (\ref{eqn:task_known_MI_Guard}) 
via training three parameterized neural networks (i.e., encoder $f$,  membership protection network $g_{\Psi}$ associated with the posterior distribution $q_{\Psi}$, and utility preservation 
network  $h_{\Omega}$ associated with the posterior distribution $q_{\Omega}$) using data samples from the  underlying data distribution.
Specifically, we first collect two datasets $D_{1}$ and $D_{0}$ from a (larger) dataset,  
and they include the members and non-members, respectively.  
Then, $D_{1}$ is used for training the utility  
network $h_\Omega$ (i.e., predicting labels for training data $D_{1}$) and the encoder $f$; and 
both $D_{1}$ and $D_{0}$ are used for training the membership protection network $g_\Psi$ 
(i.e., inferring whether a data sample from $D_{1}$/$D_{0}$ is a member or not) and the encoder $f$. 
With it, we can approximate the expectation terms in Equation (\ref{eqn:task_known_MI_Guard}) and use them to train the neural networks.

\vspace{+0.05in} 
\noindent {\bf Training the membership protection network $g_\Psi$:}
We approximate the first expectation w.r.t. $q_\Psi$ as\footnote{We omit the sample size $|D_{1}|$, $|D_{0}|$ for description brevity.}
\begin{small}
\begin{align*}
& \mathop{\mathbb{E}}_{p({\bf x}, u)}\log q_{\Psi}(u |f({\bf x} )) 
\approx
 - \sum_{({\bf x}_j, u_j) \in D_1 \cup D_0} H(u_j,g_{\Psi}(f({\bf x} _j))),  
 \end{align*}
\end{small}%
where $H(a,b)$ is the cross-entropy loss between $a$ and $b$. 
{Take a single data ${\bf x}$ with private $u$ for example. The above equation is obtained by: $ -H(u, g_\Psi(f({\bf x}))) = 
\log g_\Psi(f({\bf x}))_u  =  \log q_\Psi(u| f({\bf x})) $, where $g_\Psi(f({\bf x}))_i$ indicates $i$-th entry probability, and $q_\Psi(u| f({\bf x}))$ means the probability of inferring ${\bf x}$’s member $u$.}
The adversary maximizes this expectation aiming to enhance the membership inference performance. 
 
\vspace{+0.05in} 
\noindent {\bf Training the utility preservation network $h_\Omega$:}
We approximate the second expectation w.r.t. $q_{\Omega}$ as: 
 
 {
 \vspace{-4mm}
 \small 
 \begin{align*}
 \mathop{\mathbb{E}}_{p({\bf x}, y, u)}\log q_{\Omega}(y |f({\bf x}), u=1) 
\approx
 - \sum_{({\bf x}_j, y_j) \in D_1} H(y_j,h_{\Omega}(f({\bf x} _j))). 
 \end{align*}
}%
We maximize this expectation to enhance the utility.

\vspace{+0.05in} 
\noindent {\bf Training the encoder $f$:} With the updated $g_\Psi$ and $h_\Omega$, the defender performs gradient ascent on Equation (\ref{eqn:task_known_MI_Guard}) to update $f$, which can learn representations that protect membership privacy and further enhance the utility.

We iteratively train the three networks until reaching 
predefined 
maximum rounds. 
Figure \ref{fig:MIAs} illustrates our \texttt{\name} against MIAs. Algorithm~\ref{alg:MIGuard_MIA} in Appendix details the training. 

\begin{figure}[!t]
\vspace{-2mm}
	\centering
{\centering\includegraphics[width=0.8\linewidth]{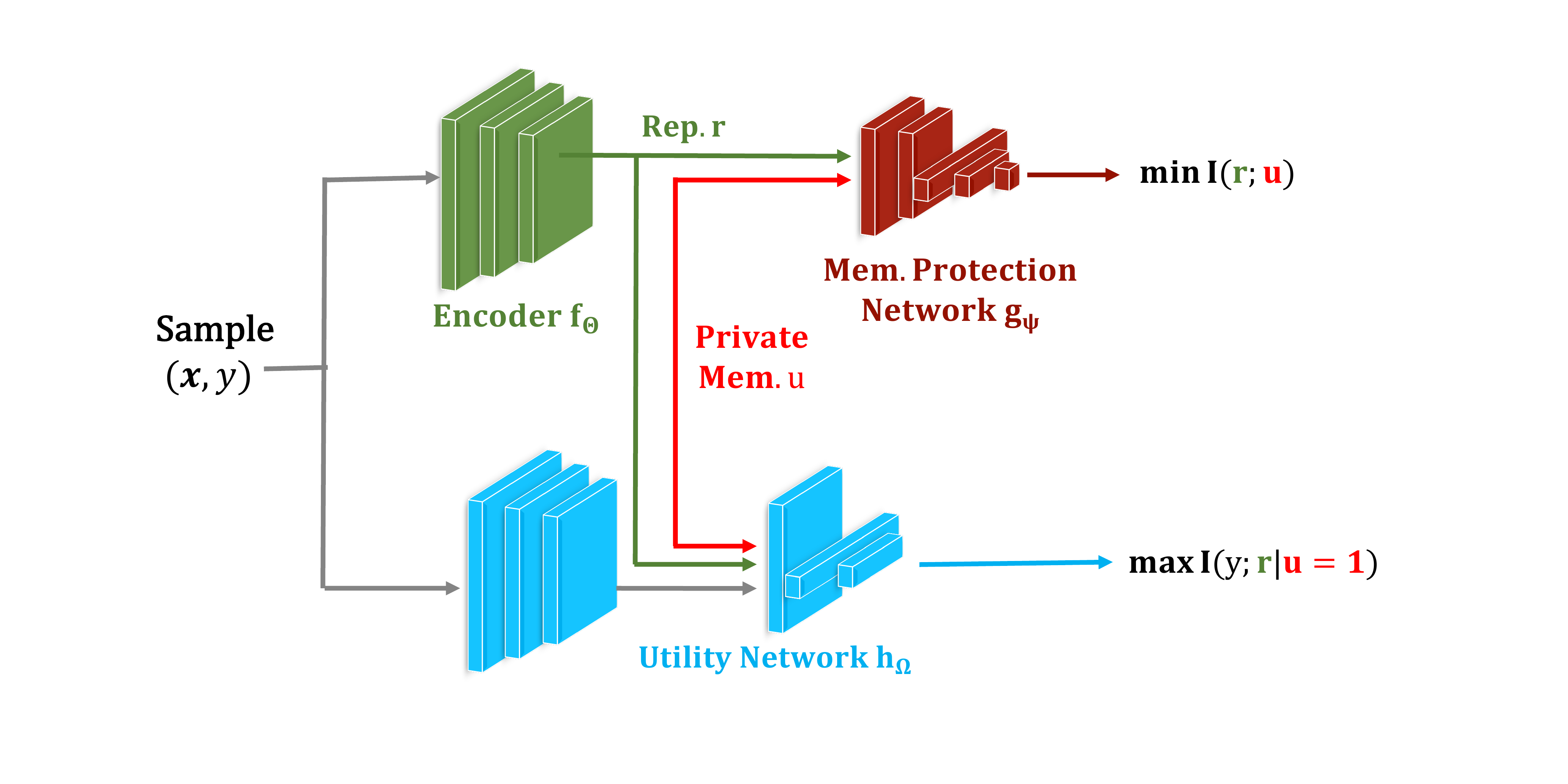}}
	\vspace{-2mm}
	\caption{\texttt{\name} against MIAs.} 
	\label{fig:MIAs}
	\vspace{-4mm}
\end{figure}

\vspace{+0.05in} 
\noindent {\bf Connection with AdvReg~\cite{nasr2018machine}.}  
We observe that AdvReg is a special case of \texttt{\name}.
Specifically, the objective function of AdvReg can be rewritten as: 

{
\vspace{-2mm}
\footnotesize
\begin{align*}
    & \max_f\Big(\lambda  \min_{\Psi} \sum_{({\bf x}_j, u_j) \in D_1 \cup D_0} H(u_j,g_{\Psi}(f({\bf x} _j)))
- (1-\lambda ) \sum_{({\bf x}_j, y_j) \in D_1} H(y_j,f({\bf x} _j)) \Big), 
\end{align*}
}%
where $f: \mathcal{X} \rightarrow [0,1]^{|\mathcal{Y}|}$ now outputs a sample's probabilistic confidence score and $g_\Psi$ is a membership inference model aiming to distinguish between members and non-members.

\subsection{\texttt{\name} against PIAs}
\label{sec:MI_PIA}

Different from MIAs, 
PIAs leak the training data properties at the \emph{dataset-level}. 
To align this, instead of using a random sample $({\bf x}, y)$, we consider a random dataset $({\bf X}, {\bf y})$ in PIAs.
Specifically, let ${\bf X} = \{{\bf x}_i\}$ consist 
of a set of independent data samples and ${\bf y} = \{y_i\}$ the corresponding data labels that are sampled from the underlying data distribution $\mathcal{D}$; and ${\bf X}$ is associated with a private (dataset) property $u$.

\subsubsection{MI objectives}
Given a dataset ${\bf X} \sim \mathcal{D}$ with a property $u$,  the defender learns a dataset representation ${\bf R} = f({\bf X})$ that satisfies two goals\footnote{For notation simplicity, we use the same $f$ to indicate the encoder. Similar for subsequent notations such as $g_\Psi$, $h_\Psi$, $q_\Omega$, $h_\Omega$, etc.}: 

\begin{itemize}[leftmargin=*]
\vspace{-2mm}
\item \textbf{Goal 1: Property protection.} 
${\bf R}$ contains as less information as possible about 
the private dataset property
$u$. Ideally, when 
${\bf R}$ does not include information about 
$u$ (i.e., ${\bf R} \perp {u}$),  it is impossible to infer 
$u$ from ${\bf R}$. 
Formally, we quantify the property protection using the below MI objective:
{
\setlength{\abovedisplayskip}{4pt}
\setlength{\belowdisplayskip}{4pt}
\begin{align}
    \min\limits_f I({\bf R}; u).  
    \label{eqn:PIA_minpriv}
\end{align}
}%

\vspace{-2mm}
\item \textbf{Goal 2: Utility preservation.} 
${\bf R}$ 
includes as much information as possible about predicting 
${\bf y}$. 
Formally, we quantify the utility preservation using the MI objective as below:
{
\setlength{\abovedisplayskip}{4pt}
\setlength{\belowdisplayskip}{4pt}
\begin{align}
    &  \max\limits_f I({\bf y}; {\bf R}).
    \label{eqn:PIA_maxutil_known} 
\end{align}
}
\end{itemize}

\subsubsection{Estimating MI via tractable bounds}

We estimate the 
bounds of Equations~\ref{eqn:PIA_minpriv} and \ref{eqn:PIA_maxutil_known} as below.

\vspace{+0.05in} 
\noindent {\bf Minimizing the upper bound MI in Equation (\ref{eqn:PIA_minpriv}).} 
Following membership protection, {\bf Goal 1} 
is reformulated as solving the below 
min-max objective function: 
\setlength{\abovedisplayskip}{4pt}
\setlength{\belowdisplayskip}{4pt}
\begin{align}
        & \min \limits_f I({\bf R};u) 
        \Longleftrightarrow \min \limits_f \max \limits_{\Psi} 
        \mathop{\mathbb{E}}_{p({\bf R}, u)} [\log q_{\Psi}(u| {\bf R})],
        \label{eqn:prot_PIA}
\end{align}
where 
$q_{\Psi}(u  | {\bf R})$ is an arbitrary posterior distribution. 

\vspace{+0.05in} 
\noindent \emph{Remark.} Similarly, Equation (\ref{eqn:prot_PIA}) can be interpreted as an \emph{adversarial game} between a property inference adversary $q_{\Psi}$ 
who aims to infer 
$u$ from the dataset representations 
${\bf R}$ and the encoder $f$ who aims to protect 
$u$  from being inferred.

\vspace{+0.05in} \noindent {\bf Maximizing the lower bound MI in Equations (\ref{eqn:PIA_maxutil_known}).} 
Similarly, we adopt the 
MI estimator \cite{nowozin2016f} to estimate the lower bound MI in our {\bf Goal 2}, which 
can be rewritten as the following max-max objective function:
{
\begin{align}
& \max_f I({\bf y}; {\bf R}) 
 \Longleftrightarrow
   \max_f \max_{\Omega} \mathop{\mathbb{E}}_{p({\bf y}, {\bf R})} \left[\log {q_{\Omega}}({\bf y} | {\bf R})\right], 
\label{eqn:max_jsd_known0} 
\end{align} 
}%
where $q_{\Omega}$ is an \emph{arbitrary} posterior distribution that aims to 
predict each label ${y} \in {\bf y}$ from the data representation ${\bf r} \in {\bf R}$.

\vspace{+0.05in} 
\noindent \emph{Remark.} 
Equation (\ref{eqn:max_jsd_known0})
can be interpreted as a \emph{cooperative game} between 
$f$ and $q_{\Omega}$
to preserve the utility collaboratively. 

\vspace{+0.05in} 
\noindent {\bf Objective function of \texttt{\name} against PIAs.} By combining Equations (\ref{eqn:prot_PIA}) and (\ref{eqn:max_jsd_known0}), 
 our objective function of learning privacy-preserving representations against PIAs is: 

{
\vspace{-2mm}
\footnotesize
\begin{align}
    &  
    \max_f\Big(\lambda  \min_{\Psi } -\mathop{\mathbb{E}}_{p({\bf X}, u)}\left[\log q_{\Psi }(u  | f({\bf X} )) \right] 
    + (1-\lambda) \max_{\Omega} 
    \mathop{\mathbb{E}}_{p({\bf X}, {\bf y})} \left[\log {q_{\Omega}}({\bf y} | f({\bf X}))\right] \Big), 
    \label{eqn:task_known_PIA_Guard} 
\end{align}
}%
where $\lambda  \in [0,1]$ tradeoffs between privacy and utility. That is, a larger/smaller $\lambda $ indicates less/more dataset property can be inferred through the learnt dataset representation.

\begin{figure}[!t]
\vspace{-2mm}
	\centering
{\centering\includegraphics[width=0.8\linewidth]{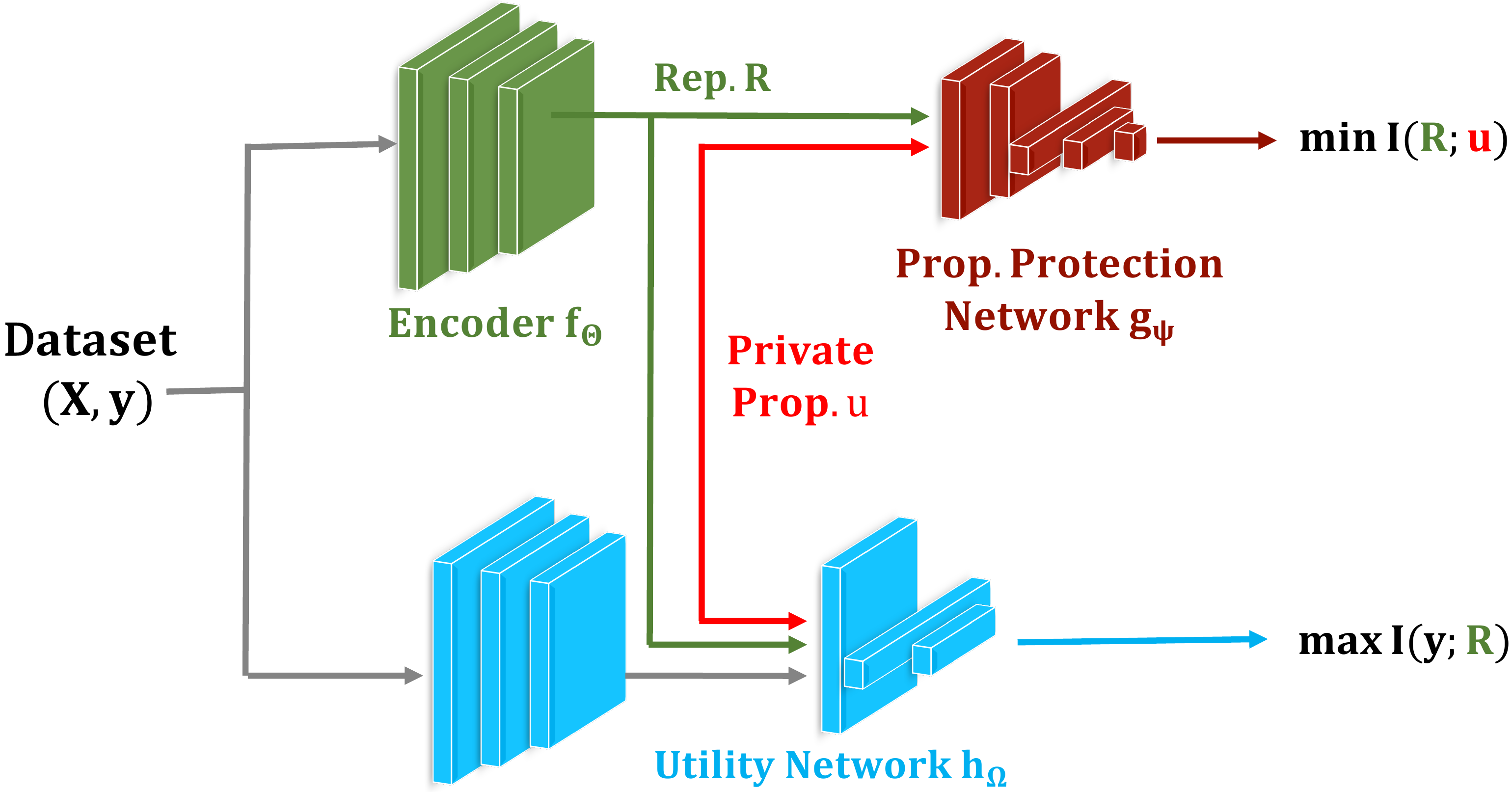}}
	\vspace{-2mm}
	\caption{\texttt{\name} against PIAs.} 
	\label{fig:PIAs}
	\vspace{-4mm}
\end{figure}

\subsubsection{Implementation in practice}
Equation (\ref{eqn:task_known_PIA_Guard}) is solved 
via three parameterized neural networks (i.e., the encoder $f_\Theta$, the property protection network $g_{\Psi}$ associated with $q_{\Psi}$, and the utility preservation network  $h_{\Omega}$ associated with $q_{\Omega}$) using a set of \emph{datasets} sampled from  a data distribution. 
Specifically, we first collect a large reference dataset $D_r$. Then, we randomly 
generate a set of small datasets $\{D_j = ({\bf X}_j, {\bf y}_j) \}_j$ from $D_r$. 
We denote the dataset property value for each $D_j$ as $u_j$.  
With it, we can approximate the expectation terms in Equation (\ref{eqn:task_known_PIA_Guard}). 

\vspace{+0.05in} 
\noindent {\bf Training the property inference network $g_\Psi$:} We approximate the first expectation w.r.t. 
$q_{\Psi}$ 
as  
{
\setlength{\abovedisplayskip}{4pt}
\setlength{\belowdisplayskip}{4pt}
\small
\begin{align}
& \mathop{\mathbb{E}}_{p({\bf X}, u)}\log q_{\Psi}(u |f({\bf X} )) \approx
 - \sum_{\{{\bf X}_j \in D_j\}} H(u_j,g_{\Psi}(f({\bf X}_j))),
 \end{align}
}%
where $f({\bf X}_j)$ is  the aggregated representation of a dataset ${\bf X}_j$, i.e., $f({\bf X}_j) = \text{Agg}(\{f({\bf x})\}_{{\bf x} \in {\bf X}_j})$. We will discuss the aggregator $\text{Agg}(\cdot)$ in Section~\ref{sec:results_PIAs}.  
The adversary maximizes this expectation to enhance the property inference performance.

\vspace{+0.05in} 
\noindent {\bf Training the utility preservation network $h_\Omega$:}
Similarly, we approximate the second expectation w.r.t. $q_{\Omega}$ as: 
\setlength{\abovedisplayskip}{4pt}
\setlength{\belowdisplayskip}{4pt}
\begin{small}
  \begin{align}
\mathop{\mathbb{E}}_{p({\bf X}, {\bf y})} \log q_{\Omega}({\bf y} |f({\bf X}))  
 \approx
 - \sum_{\{D_j\}} \sum_{({\bf x}_i, {y}_i) \in D_j}  H({y}_i,h_{\Omega}(f({\bf x}_i))),
 \label{eqn:sumall}
 \end{align}
 \end{small}%
 where we maximize this expectation to enhance the utility.

\vspace{+0.05in} 
\noindent {\bf Training the encoder $f$:} The defender then performs gradient ascent on Equation (\ref{eqn:task_known_PIA_Guard}) to update $f$, which  mitigates the PIA and further enhances the utility.  

We iteratively train the three networks until reaching 
maximum rounds. 
Figure \ref{fig:PIAs} illustrates our \texttt{\name} against PIAs. Algorithm~\ref{alg:MIGuard_PIA} in Appendix details the training process. 

 \subsection{\texttt{\name} against DRAs}
\label{sec:MI_DRA}

Different from MIAs and PIAs, DRAs aim to \emph{directly} recover the training data from the learnt representations. 
A recent defense~\cite{sun2020provable}   
shows perturbing the latent representations 
can somewhat protect the data from being reconstructed. However, this defense is broken by an advanced attack~\cite{balunovic2022bayesian}. 
One key reason 
is  the defense 
perturbs representations in a \emph{deterministic} fashion for already  \emph{trained} models. 
We 
address the issues and 
propose an information-theoretic defense to learn \emph{randomized} representations against 
the 
DRAs in an end-to-end \emph{learning} fashion. 
{Our core idea is to learn a \emph{deterministic} encoder and a \emph{randomized} perturbator that ensures learning the perturbed representation in a controllable manner.}

\vspace{-2mm} 
\subsubsection{MI objectives}
 
Given a data sample ${\bf x} \sim \mathcal{D}$ with a label $y$, 
 the defender
 learns a representation ${\bf r} = f({\bf x})$ such that when ${\bf r}$ is perturbed by certain perturbation (denoted as $\bm{\delta}$), the {shared} perturbed representation ${\bf r} + \bm{\delta}$ cannot be used to well recover ${\bf x}$, but is effective for predicting $y$, from the information-theoretic perspective. 
Then we aim to achieve the following two goals:

\begin{itemize}[leftmargin=*]
\vspace{-2mm}
\item \textbf{Goal 1: Data reconstruction protection.} 
${\bf r} + \bm{\delta}$ contains as less information as possible about ${\bf x}$. 
Moreover, the perturbation $\bm{\delta}$ should be effective enough.
Hence, we require 
$\bm{\delta}$ can cover all directions of ${\bf x}$, 
and force the entropy of 
$\bm{\delta}$ 
to be as large as possible.
Formally, we quantify the data reconstruction protection using the below MI objective:
{
\setlength{\abovedisplayskip}{4pt}
\setlength{\belowdisplayskip}{4pt}
\begin{align}
    & \min\limits_{f, p(\bm{\delta}) \sim \mathcal{P}} I({\bf r} + \bm{\delta}; {\bf x}) - H({\bm{\delta}}),
    \label{eqn:DRA_minpriv}
\end{align}
}%

\vspace{-2mm}
\item \textbf{Goal 2: Utility preservation.} 
To ensure 
${\bf r}$ be useful, it should be effective for predicting the  label $y$.  Further, as we will share the perturbed representation 
${\bf r} + \bm{\delta} $, it should be also effective for predicting 
$y$. 
Formally, we quantify the utility preservation using the MI objective as follows:
{
\setlength{\abovedisplayskip}{4pt}
\setlength{\belowdisplayskip}{4pt}
\begin{align}
       & \max\limits_{f, p(\bm{\delta}) \sim \mathcal{P}} I({\bf r} + \bm{\delta}; y) + I({\bf r}; {y}),  
        \label{eqn:DRA_maxutil}
\end{align}
}%
\vspace{-4mm}
\end{itemize}

\subsubsection{Estimating MI via tractable bounds} 

\noindent {\bf Minimizing the upper bound MI in Equation (\ref{eqn:DRA_minpriv}).}
Similarly, we adapt the variational upper bound in~\cite{cheng2020club}. Our {\bf Goal 1} for data reconstruction protection 
can be reformulated as the below 
\emph{min-max} objective function:
\setlength{\abovedisplayskip}{4pt}
\setlength{\belowdisplayskip}{4pt}
\begin{small}
\begin{align}
        & \min \limits_{f, p(\bm{\delta}) \sim \mathcal{P}} I({\bf r} + \bm{\delta};{\bf x}) - \alpha H(\bm{\delta}) \label{eqn:prot_DRA}
       \\ 
        & \Longleftrightarrow \min \limits_{f,p(\bm{\delta}) \sim \mathcal{P}} \big( \max \limits_{\Psi} 
        \mathop{\mathbb{E}}_{p({\bf r}, \bm{\delta}, {\bf x})} [\log q_{\Psi}({\bf x}| {\bf r} + \bm{\delta})] -\alpha H(\bm{\delta}) \big).
         \notag 
\end{align}
\end{small}%

\noindent \emph{Remark.} Equation (\ref{eqn:prot_DRA}) can be interpreted as an \emph{adversarial game} between an adversary $q_{\Psi}$ (i.e., data reconstructor) who aims to infer ${\bf x}$ from 
${\bf r} + \bm{\delta}$; and the encoder $f$ who aims to protect 
${\bf x}$ from being inferred via carefully perturbing ${\bf r}$.

\vspace{+0.05in} \noindent {\bf Maximizing the lower bound MI in Equation (\ref{eqn:DRA_maxutil}).}
Based on~\cite{poole2019variational}, we can produce a lower bound on the MI $I({\bf r};y)$ due to the non-negativity of the KL-divergence: 

{
\vspace{-4mm}
\small
\begin{align}
I({\bf r};{y}) & = \mathop{\mathbb{E}}_{p({y} , {\bf r})} \left[ \log {q_{\Omega}({y}|{\bf r})}/{p({y})}\right] 
+ 
\mathop{\mathbb{E}}_{p({\bf r})}[KL(p({y}|{\bf r})||q_{\Omega}({y}||{\bf r}))] \notag \\ 
& \ge 
\mathop{\mathbb{E}}_{p({y} , {\bf r})} \left[\log {q_{\Omega}}({y}  | {\bf r})\right] + H({y}),
\end{align}
}%
where $q_{\Omega}$ is an \emph{arbitrary} posterior distribution that predicts the label $y$ from ${\bf r}$ and 
the entropy 
$H({y})$ is a constant.

We have a similar form for the MI $I({\bf r} + \bm{\delta}; y)$ as below 
{
\vspace{-4mm}
\small
\begin{align}
I({\bf r} + \bm{\delta}; y) \ge 
    \max_{p(\bm{\delta})\sim \mathcal{P}} \mathop{\mathbb{E}}_{p(y , {\bf r}, \bm{\delta})} \left[\log {q_{\Omega}}(y  | {\bf r}+\bm{\delta}) \right] + H(y),
    \end{align}
}%
where we use the same $q_{\Omega}$ 
to predict the label $y$ from the perturbed representation ${\bf r} + \bm{\delta}$.

Then, our {\bf Goal 2} for utility preservation can be rewritten as the following max-max objective function:

{
\vspace{-4mm}
\small
\begin{align}
    & \max_{f, p(\bm{\delta})\sim \mathcal{P}} I({\bf r} + \bm{\delta}; y)  +   I({\bf r};{y})  
        \label{eqn:max_jsd_DRA} \\
    &    \Longleftrightarrow \max_{f, \Omega} \big(\max_{p(\bm{\delta}) \sim \mathcal{P} } \mathop{\mathbb{E}}_{p(y , {\bf r}, \bm{\delta})} \left[ \log {q_{\Omega}}(y  | {\bf r} + \bm{\delta}) \right] 
    +   \mathop{\mathbb{E}}_{p({y} , {\bf r})} \left[ \log {q_{\Omega}}({y}  | {\bf r}) \right]\big).  \notag
\end{align} 
}%
\emph{Remark.} Equation (\ref{eqn:max_jsd_DRA}) can be interpreted as a \emph{cooperative game} between the encoder $f$
and the label prediction network $q_{\Omega}$, who aim to preserve the utility collaboratively.

\vspace{+0.05in} 
\noindent {\bf Objective function of \texttt{\name} against DRAs.} 
By combining Equations (\ref{eqn:prot_DRA})-(\ref{eqn:max_jsd_DRA}), 
 our objective function of learning privacy-preserving representations against DRAs is:
 
 {
\vspace{-2mm}
\footnotesize
\begin{align}
    & \max_{f,p(\bm{\delta}) \sim \mathcal{P} }\Big(\lambda  \big( \min_{\Psi} - \mathop{\mathbb{E}}_{p({\bf x}, \bm{\delta})}\left[\log q_{\Psi}({\bf x}  | f({\bf x}) + \bm{\delta}) \right] + \alpha H(\bm{\delta}) \big) 
    \label{eqn:task_known_DRA_Guard}   \\ 
    & 
    + (1-\lambda ) \big(  \max_{\Omega} \mathop{\mathbb{E}}_{p({\bf x}, \bm{\delta}, y)} \left[ \log {q_{\Omega}}(y  | f({\bf x}) + \bm{\delta}) \right] 
     + 
      \mathop{\mathbb{E}}_{p({\bf x}, y)} \left[\log {q_{\Omega}}({y} | f({\bf x})) \right] \big) \Big), \notag
\end{align}
}%
where $\lambda  \in [0,1]$ tradeoffs privacy and utility. A larger $\lambda $ implies less data features can be inferred through the perturbed representation, while a smaller $\lambda$ implies the 
shared perturbed representation is easier for predicting the label. 

\subsubsection{Parameterizing perturbation distributions} 
 
The key of our defense lies in defining the perturbation distribution 
$p(\bm{\delta})$ in Equation (\ref{eqn:task_known_DRA_Guard}). 
Directly specifying the optimal perturbation distribution is challenging. 
Motivated by variational inference~\cite{kingma2014auto}, 
we propose to parameterize 
$p(\bm{\delta})$ with trainable parameters, e.g., $\Phi$. 
Then the optimization problem w.r.t. the perturbation $\bm{\delta}$ can be converted to be w.r.t. the parameters $\Phi$, which can be solved via back-propagation. 

A natural way to model the perturbation  
around a representation 
is using a distribution with an {explicit density function}. 
Here we adopt the method in \cite{kingma2014auto} by transforming $\bm{\delta}$ such that the reparameterization trick can be used in training. 
For instance, when considering $p_{\Phi}(\bm{\delta})$ as a Gaussian distribution $\mathcal{N}(\bm{\mu},\bm{\sigma}^2)$, we can reparameterize $\bm{\delta}$ (with a scale $\epsilon$) as: 
{
\begin{align}
\label{eqn:paratrans}
\bm{\delta} = \epsilon \cdot \textrm{tanh}({\bf u}), \quad {\bf u} \sim \mathcal{N}(\bm{\mu}, \textrm{diag}(\bm{\sigma}^2)),
\end{align} 
}%
That is, it first samples ${\bf u}$  from a diagonal Gaussian 
with a mean vector $\bm{\mu}$ and standard deviation vector $\bm{\sigma}$, and $\bm{\delta}$ is obtained by compressing ${\bf u}$ to be $[-1,1]$ via the $\textrm{tanh}(\cdot)$ function and multiplying 
$\epsilon$. $\Phi=(\bm{\mu}, \bm{\sigma})$ are the parameters to be learnt.

\begin{figure}[!t]
\centering
\vspace{-2mm}
{\centering\includegraphics[width=0.8\linewidth]{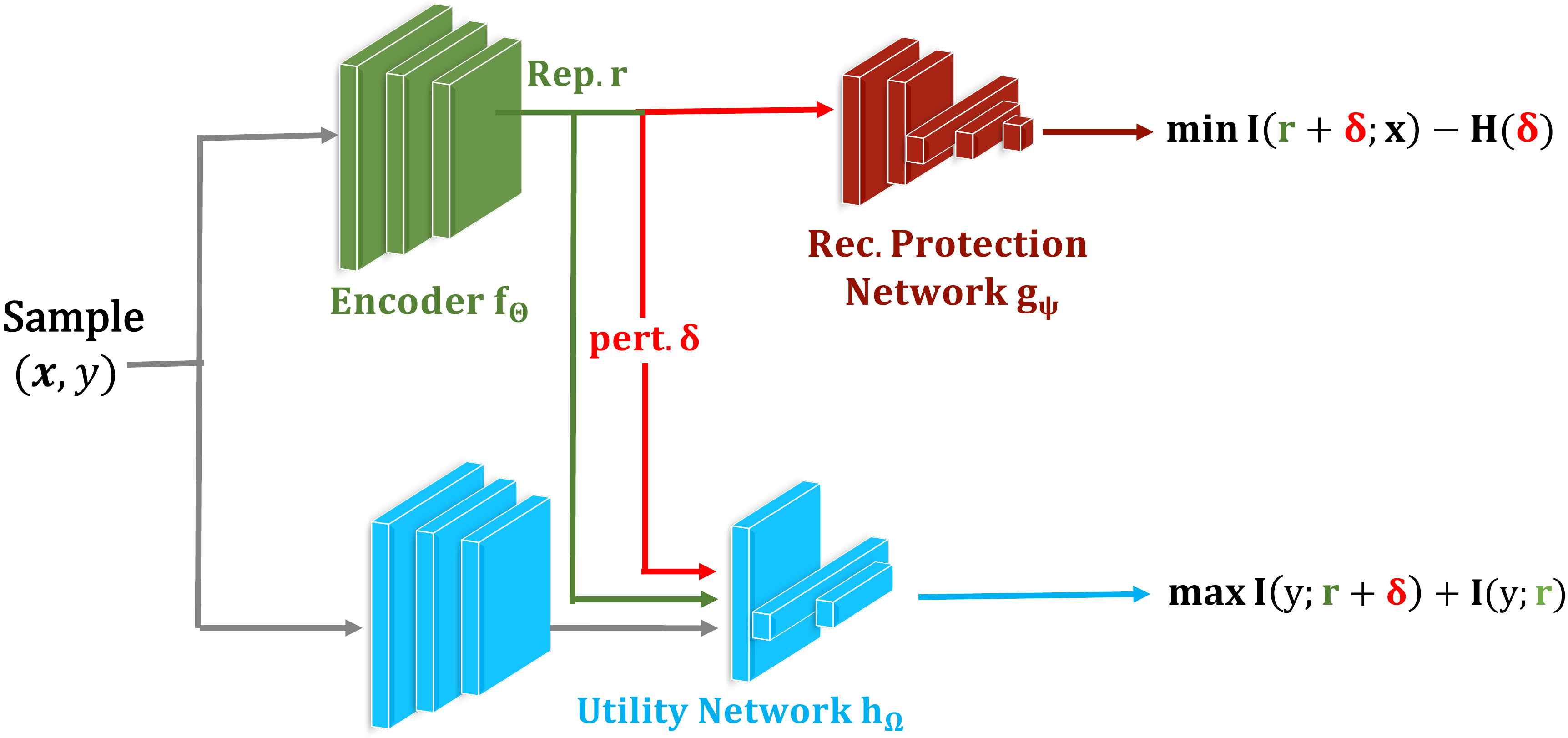}}
	\vspace{-2mm}
	\caption{\texttt{\name} against DRAs.} 
	\label{fig:DRAs}
	\vspace{-4mm}
\end{figure}

\subsubsection{Implementation in practice}

We train three neural networks (i.e., the encoder $f$,  reconstruction protection network $g_{\Psi}$,  and utility preservation network $h_{\Omega}$) using data samples from certain data distribution. 
Suppose we are given a set of data samples $D=\{{\bf x}_j, y_j\}$. 

\vspace{+0.05in} 
\noindent
{\bf Learning the data reconstruction network $g_\Psi$:} 
As ${\bf x}$ and its representation ${\bf r}$ are often high-dimensional, the previous MI estimators are inappropriate in this setting. 
To address it, we use the \textit{Jensen-Shannon} divergence (JSD)  \cite{hjelm2019learning} specially for high-dimensional MI estimation. 
Assume we have updated $\Phi$. 
We can approximate the expectation w.r.t. $g_\Psi$ as

{
\vspace{-4mm}
\small
  \begin{align}
 & \mathop{\mathbb{E}}_{p({\bf x}), p_\Phi(\bm{\delta})}\log q_{\Psi}({\bf x} |f({\bf x}) + \bm{\delta}) 
= 
 I^{(JSD)}_{\Theta, \Psi} ({\bf x} ; f_\Theta({\bf x})+ \bm{\delta}) \notag \\  
 & \approx \sum_{{\bf x}_j \in D, \bm{\delta}_j \sim p_\Phi(\bm{\delta}) }[-\textrm{sp}(-h_{\Psi}({\bf x}_j ,f_\Theta({\bf x}_j) + \bm{\delta}_j))] \notag \\
 & \quad - \sum_{({\bf x}_j, {\bf x}'_j) \in D, \bm{\delta}_j \sim p_\Phi(\bm{\delta})}[\textrm{sp}(h_{\Psi}({\bf x}'_j, f_\Theta({\bf x}_j) + \bm{\delta}_j)], 
 \label{eqn:MI_JSD}
 \end{align}
}%
where ${\bf x}'_j$ is an independent and random sample from the same distribution as ${\bf x}_j$, 
and $\textrm{sp}(z) = \log(1+\exp(z))$ is the softplus function. 
We maximize 
$ I^{(JSD)}_{\Theta, \Psi}$ to update $g_\Psi$.

\vspace{+0.05in} 
\noindent
{\bf Learning the utility preservation network $h_\Omega$:}
We first estimate the below expectation: 

{
\vspace{-3mm}
\footnotesize
\begin{align}
 & \mathop{\mathbb{E}}_{p({\bf x}, y) p_\Phi(\bm{\delta})}\log q_{\Omega}(y |f({\bf x})+ \bm{\delta})  
 \approx
 - \sum_{\substack{({\bf x}_j, y_j) \in D, \bm{\delta}_j \sim p_\Phi(\bm{\delta})}} H(y_j,h_{\Omega}(f({\bf x} _j)+\bm{\delta}_j)) \label{eqn:CE_pert}. 
 \end{align} 
 }%
 Similarly, we can approximate the third expectation as:

{
\vspace{-3mm}
\footnotesize
 \begin{align}
 & \mathop{\mathbb{E}}_{p({\bf x}, y)}\log q_{\Omega}(y |f({\bf x})) \approx
 - \sum_{({\bf x}_j, y_j) \in D} H(y_j,h_{\Omega}(f({\bf x} _j))) \label{eqn:CE_clean}. 
 \end{align} 
}%
We minimize the two cross entropy losses to update $h_\Omega$. 

\vspace{+0.05in} 
\noindent {\bf Updating the distribution parameter $\Phi$:} 
Due to the reparameterization trick,  the gradient can be back-propagated from each $\bm{\delta}_j$ to the  parameters $\Phi$. 
For simplicity, 
 we do not consider the JSD term in Equation (\ref{eqn:MI_JSD}) 
due to its complexity. 
Then we have the terms relevant to $\Phi$ as below: 

{
\vspace{-3mm}
\footnotesize
\begin{align}
& \mathop{\mathbb{E}}_{\bm{z} \sim \mathcal{N}(0,1)} \sum_{({\bf x}_j, y_j)}  H(y_j,h_{\Omega}(f({\bf x} _j)+ \epsilon \cdot \text{tanh}(\bm{\mu}+\bm{\sigma z}))) 
-  \beta \cdot H(\epsilon \cdot \text{tanh}(\bm{\mu}+\bm{\sigma z})), 
\label{eqn:genet}
\end{align}
}%
where $\beta=\lambda \alpha / (1-\lambda)$. The first term is the cross entropy loss, while the second term is the entropy. 
The gradient w.r.t. $\Phi$ in each term can be  calculated. 
In practice, we approximate the expectation on $\bm{z}$ with (e.g., 5) Monte Carlo samples, and perform 
\emph{stochastic gradient descent} to update $\Phi$. 
Details on 
updating $\Phi$ 
are in Algorithm~\ref{alg:update_perdist}. 
With $\Phi$, we use it to generate 
$\bm{\delta}$ and add it to ${\bf r}$ to produce the  perturbed representation.  

\vspace{+0.05in} \noindent 
{\bf Learning the encoder $f$.} Finally, after updating $g_\Psi$, $h_\Omega$, and 
$\Phi$, we can perform gradient ascent to update $f$.

We iteratively train the networks until reaching a predefined maximum round. Figure~\ref{fig:DRAs} illustrates 
\texttt{\name} against DRAs. Algorithm~\ref{alg:MIGuard_DRA} in Appendix details the  training process.

\section{Theoretical Results}
\label{sec:theory}
  
{We mainly show the guaranteed privacy leakage under \texttt{\name}, with proofs in Appendix~\ref{app:PrivGuarantees}. 
We also derive an inherent utility-privacy tradeoff of \texttt{\name}, 
which requires a binary classification task,  and binary-valued dataset property in PIAs. Details and proofs are deferred to Appendix~\ref{app:UPTradeoff}.}

\vspace{+0.02in}
\noindent {\bf Guaranteed privacy leakage of MIAs:} 
Let $\mathcal{A}_{MIA}$ be the set of all MIAs $\mathcal{A}_{MIA}=\{A_{MIA}: \mathcal{Z} \rightarrow u \in \{0,1\}\}$ that have access to the representations ${\bf r}$ by querying $f$ with data ${\bf x}$ from the distribution $\mathcal{D}$.
The MIA accuracy 
is bounded as below:

\begin{restatable}[]{theorem}{provprivacymi}
\label{thm:provprivacy_MI}
\vspace{-2mm}
Let $f$ be the  learnt encoder   
by Equation (\ref{eqn:task_known_MI_Guard}) over a  data distribution $\mathcal{D} \subset \mathcal{X}$. For a random data sample ${\bf x} \sim \mathcal{D}$ with the learnt representation ${\bf r} = f({\bf x})$ and membership $u$, we have 
$\textrm{Pr}({A}_{MIA}({\bf r}) = u) \leq 1 - \frac{H(u | {\bf r})}{2 \log_2 ({6}/{H(u | {\bf r})})}, \forall A_{MIA} \in \mathcal{A}_{MIA}$. 
\vspace{-2mm}
\end{restatable}
\noindent \emph{Remark.} Theorem~\ref{thm:provprivacy_MI} shows if  
$H(u | {\bf r})$ is larger, the bounded 
MIA accuracy 
is smaller. 
Note $H(u | {\bf r}) = H(u) - I(u;  {\bf r})$ and $H(u)$ is a constant.  Achieving a large $H(u | {\bf r})$ implies obtaining a small 
$I(u;  {\bf r})$, which is our {\bf Goal 1} in Equation (\ref{eqn:MIA_minpriv}) does. 
{In practice, once the encoder $f$ is learnt on a dataset from 
$\mathcal{D}$, $I(u;{\bf r})$ can be estimated, then the bounded MIA accuracy can be calculated. A better encoder $f$ or/and better MI estimator of $I(u;{\bf r})$ can yield a smaller MIA performance.
}

\vspace{+0.02in}
\noindent {\bf Guaranteed privacy leakage of PIAs:} Let $\mathcal{A}_{PIA}$ be the set of all PIAs that have access to the representations ${\bf R}$ of a dataset ${\bf X} = \{ {\bf x}_i \} $ sampled from the data distribution $\mathcal{D}$, i.e., $\mathcal{A}_{PIA}=\{A_{PIA}: \mathcal{Z}
\rightarrow u=\{0,1\}\}$. The PIA accuracy 
is bounded as:

\begin{restatable}[]{theorem}{provprivacypi}
\label{thm:provprivacy_PI}
\vspace{-2mm}
Let $f$ be the  learnt encoder   by Equation (\ref{eqn:task_known_PIA_Guard}) over a data distribution $\mathcal{D}$. For a random dataset ${\bf X} \sim \mathcal{D}$ with the learnt representation ${\bf R} = f({\bf X})$ and dataset property $u$, we have 
$\textrm{Pr}({A}_{PIA}({\bf R}) = u) \leq 1 - \frac{H(u | {\bf R})}{2 \log_2 ({6}/{H(u | {\bf R})})}, \forall A_{PIA} \in \mathcal{A}_{PIA}$. 
\vspace{-4mm}
\end{restatable} 
\noindent \emph{Remark.} 
Theorem~\ref{thm:provprivacy_PI} shows  when  $H(u | {\bf R})$ is larger, the 
PIA accuracy is smaller, i.e., less dataset property is leaked. 
Also, 
a large $H(u | {\bf R})$ indicates a small $I(u;  {\bf R})$---This is exactly our {\bf Goal 1} in Equation (\ref{eqn:PIA_minpriv}) aims to achieve.

\vspace{+0.02in}
\noindent {\bf Guaranteed privacy leakage of DRAs:} Let $\mathcal{A}_{DRA}$ be the set of all DRAs that have access to the perturbed data representations, i.e., $\mathcal{A}_{DRA}=\{A_{DRA}: {\bf r} + \bm{\delta} \in \mathcal{Z} \times \mathcal{P} \rightarrow {\bf x} \in \mathcal{X}\}$. 
{An $\ell_p$-norm ball centered at 
a point ${\bf v}$ with a radius $\rho$ is denoted as 
$\mathcal{B}_p({\bf v},\rho)$, i.e., $\mathcal{B}_p({\bf v},\rho) = \{ {\bf v}': \|{\bf v}' -  {\bf v}\|_p \le \rho \}$. For a space $\mathcal{S}$, we denote its boundary as $\partial \mathcal{S}$, whose volume is denoted as $\textrm{Vol}(\partial \mathcal{S})$.  
Then the} reconstruction error (in terms of $\ell_p$ norm difference)
incurred by any DRA is bounded as below:

\begin{restatable}[]{theorem}{provprivacydrappr}
\label{thm:provprivacy_DR_appr}
\vspace{-1mm}
Let $f$ be the encoder learnt 
by Equation (\ref{eqn:task_known_DRA_Guard}) over a  data distribution $\mathcal{D} \subset \mathcal{X}$ and $\bm{\delta}$ be the perturbation 
 for a random sample $\bf x \sim \mathcal{X}$.  
Then, 
$\textrm{Pr}( \| {A}_{DRA}({\bf r} + \bm{\delta}) - {\bf x} \|_p \geq \eta) \geq 1 - \frac{I({\bf x}; {\bf r}+ \bm{\delta}) + \log 2}{\log \text{Vol}(\partial \mathcal{X}) - \log \text{Vol}(\partial \mathcal{X}(\eta))}, \forall {A}_{DRA} \in \mathcal{A}_{DRA} $, where 
$\text{Vol}(\partial \mathcal{X}(\eta)) = {\max_{{\bf x} \in \mathcal{X}} \textrm{Vol}(\partial \mathcal{B}_p({\bf x},\eta)\cap \mathcal{X})}$. 
\vspace{-1mm}
\end{restatable}
\noindent \emph{Remark.} Theorem~\ref{thm:provprivacy_DR_appr} shows a lower bound error achieved by the strongest 
DRA. Given an $\eta$, when $I({\bf x}; {\bf r}+ \bm{\delta})$ is smaller,  the  lower bound 
data reconstruction 
error is larger, meaning the privacy of the data itself is better protected. 
Moreover, minimizing $I({\bf x}; {\bf r}+ \bm{\delta})$ is exactly our {\bf Goal 1} in Equation (\ref{eqn:DRA_minpriv}).

\begin{figure*}[!t]
\vspace{-2mm}
\centering
\subfigure[CIFAR10]{\centering\includegraphics[width=0.29\linewidth]{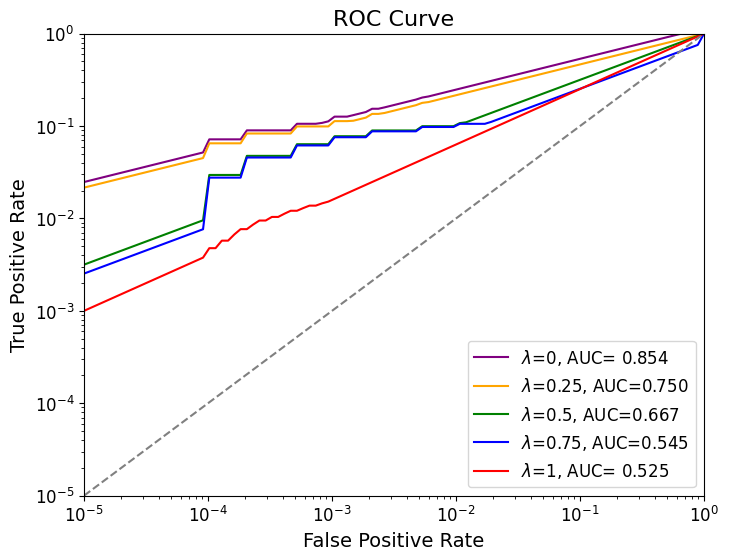}}
\subfigure[Purchase100]{\centering\includegraphics[width=0.29\linewidth]{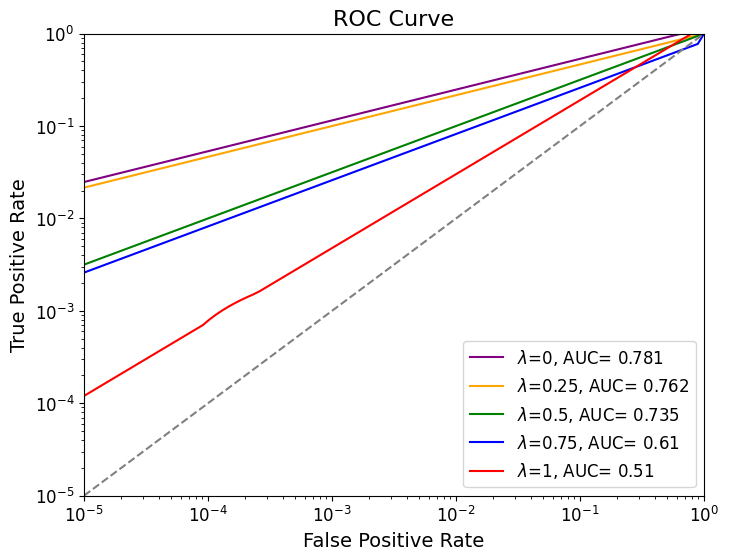}}
\subfigure[Texas100]{\centering\includegraphics[width=0.29\linewidth]{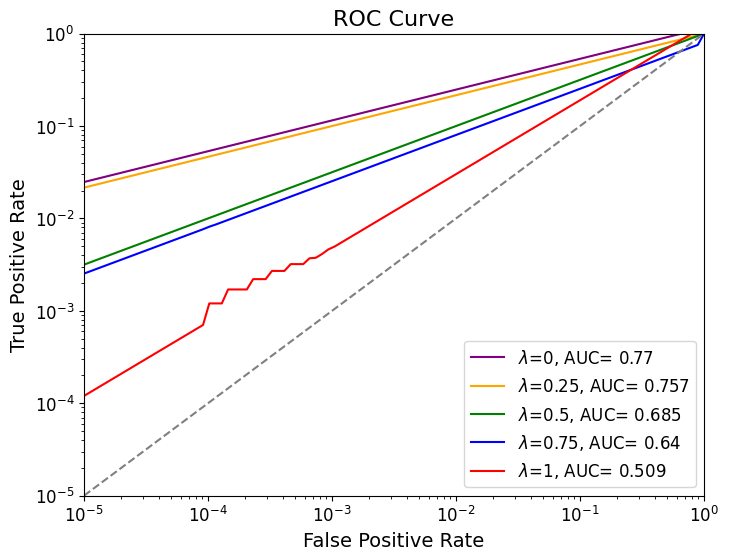}}
	\vspace{-4mm}
	\caption{TPR vs FPR of \texttt{\name} against LiRA on different $\lambda$'s.} 
	\label{fig:LiRA}
	\vspace{-4mm}
\end{figure*}

\section{Evaluations}
\label{sec:eval}

In this section, we will evaluate \texttt{\name} against the MIAs, PIAs, and DRAs on benchmark datasets. 
{\texttt{\name} involves training the encoder, the privacy protection network, and the utility preservation network. The detailed dataset description and architectures of the networks are given in Appendix~\ref{appendix:Datasets}.}

\subsection{Defense Results on MIAs}

\subsubsection{Experimental setup}

{\bf Datasets:} Following existing works~\cite{nasr2018machine,jia2019memguard}, we use the CIFAR10~\cite{Krizhevsky09learningmultiple}, Purchase100~\cite{nasr2018machine}, and Texas100~\cite{shokri2017membership} datasets, to evaluate \texttt{\name} against MIAs.

\vspace{+0.01in}
\noindent {\bf Defense/attack training and testing:} The 
training sets and test sets 
are listed in Table~\ref{tbl:dataset_MIAs} in Appendix~\ref{app:setup}. 
For instance, in CIFAR10, we use 50K samples in total and split it into two halves, where 25K samples are used as the \emph{utility training set} ("members") and the other 25K samples as the \emph{utility test set} ("non-members"). We select 80\% of the members and non-members as the \emph{attack training set} and the remaining members and non-members as the \emph{attack test set}. 

\begin{itemize}[leftmargin=*]
\vspace{-2mm}
\item {\bf Defense training:} We use the utility training set and attack training set to train the encoder, utility preservation network, and membership protection network simultaneously. Then, the learnt encoder is frozen and published as an API. 

\vspace{-2mm}
\item {\bf Attack training:} 
To mimic the strongest possible MIA, we let the attacker know the exact  membership protection network and attack training set used in defense training. Specifically, s/he feeds the attack training set to the  learnt encoder to get the data representations and trains the MIA classifier (same as the  membership protection network) on these representations to maximally infer the membership.   

\vspace{-2mm}
\item {\bf Defense and attack testing:}
We use the utility test set to obtain the utility (i.e., test accuracy) via querying the trained encoder and utility network. Moreover, we  use 
the attack test set to obtain the MIA performance.
\vspace{-2mm}

\end{itemize}

\noindent {{\bf Privacy metric:} We measure the MIA performance via both the MIA accuracy and the true positive rate (TPR) vs. false
positive rate (FPR), suggested by the SOTA LiRA MIA \cite{carlini2022membership}. 
Specifically, the MIA accuracy is obtained by querying the trained encoder and
trained MIA classifier with the attack test set. 
Moreover, we treat the representations and membership network learnt by \texttt{\name} 
as the input data and target model for LiRA. 
We then train 16 \emph{white-box} shadow models (i.e., assume LiRA uses the exact membership network in \texttt{\name}) on the data representations of the utility training set, and report the TPR vs. FPR on the attack test set.}

\begin{table}[!t]\renewcommand{\arraystretch}{0.9}
\vspace{-2mm}
\footnotesize
\caption{\texttt{\name} results against MIAs on the three dataset. $\lambda=0$ means no privacy protection, while $\lambda=1$ means no utility preservation. Random guessing MIA accuracy is 50\%.} 
 \addtolength{\tabcolsep}{-3.5pt}
\centering
\begin{minipage}{.32\linewidth}
\begin{tabular}{ccc}
\toprule
\multicolumn{3}{c}{\bf CIFAR10}\\
          \hline
\( \lambda \) & {\bf Utility} & {\bf MIA Acc} \\
\midrule
0 & 78.9\% & 70.1\% \\
0.25 & 78.2\% & 55.9\% \\
0.5 & 78\% & 53.5\% \\
0.75 & 77.2\% & 51.1\% \\
1 & 20\% & 50\% \\
\bottomrule
\end{tabular}
\end{minipage}
\hfill
\begin{minipage}{.32\linewidth}
\begin{tabular}{ccc}
\toprule
\multicolumn{3}{c}{\bf Purchase100}\\
          \hline
\( \lambda \) & {\bf Utility} & {\bf MIA Acc} \\
\midrule
0 & 81.7\% & 68.4\% \\
0.25 & 80.9\% & 60\% \\
0.5 & 80\% & 51\% \\
0.75 & 78\% & 50\% \\
1 & 20\% & 50\% \\
\bottomrule
\end{tabular}
\end{minipage}
\hfill
\begin{minipage}{.32\linewidth}
\begin{tabular}{ccc}
\toprule
\multicolumn{3}{c}{\bf Texas100}\\
          \hline
\( \lambda \) & {\bf Utility} & {\bf MIA Acc} \\
\midrule
0 & 49.9\% & 70.2\% \\
0.25 & 49.1\% & 61\% \\
0.5   & 47\% & 53\% \\
0.75  & 46\% & 50\% \\
1     & 2\% & 50\% \\
\bottomrule
\end{tabular}
\end{minipage}
\label{tab:results_MIAs}
\vspace{-4mm}
\end{table}

\subsubsection{Experimental results}
\label{sec:results_MIAs}

\noindent {\bf Utility-privacy results:} According to Equation (\ref{eqn:task_known_MI_Guard}), \( \lambda=0 \) indicates no privacy protection. 
Increasing $\lambda$'s value enhances  \texttt{\name}'s resilience against MIAs. 
\( \lambda = 1 \) means the maximum privacy protection 
without preserving utility. 
Table~\ref{tab:results_MIAs} shows the {utility-MIA Accuracy} results of \texttt{\name}. 
We have the following observations: 
1) The MIA accuracy is the largest when $\lambda=0$, implying leaking the most membership privacy by MIAs. 
2) When only protecting privacy ($\lambda=1$), the MIA accuracy reaches to the optimal random guessing, but the utility is the lowest. 
3) When $0<\lambda<1$, \texttt{\name} obtains reasonable utility and MIA accuracy. Especially, when $\lambda=0.75$, the utility loss is marginal (i.e., $<4\%$), while the MIA accuracy is (close to) random guessing. The results show the learnt privacy-preserving encoder/representations are effective against MIAs, and maintain utility as well.

{Further, Figure~\ref{fig:LiRA} shows the TPR vs FPR of \texttt{\name} against LiRA. 
Similarly, we observe that the TPR at low FPRs is relatively large (strong membership inference) in case of no privacy protection, but it can be largely reduced by increasing $\lambda$. This implies that \texttt{\name} indeed learns the representations that can defend against LiRA to some extent.}

\begin{figure*}[!t]
\vspace{-2mm}
	\centering
	\subfigure[Utility w/o. defense (78.9\%)]
{\centering\includegraphics[width=0.22\linewidth]{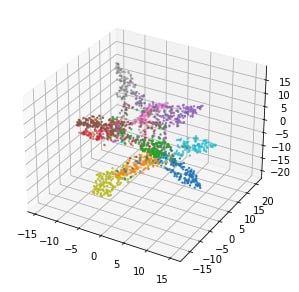}}
	\subfigure[MIA Acc w/o. defense (70.1\%)]
{\centering\includegraphics[width=0.22\linewidth]{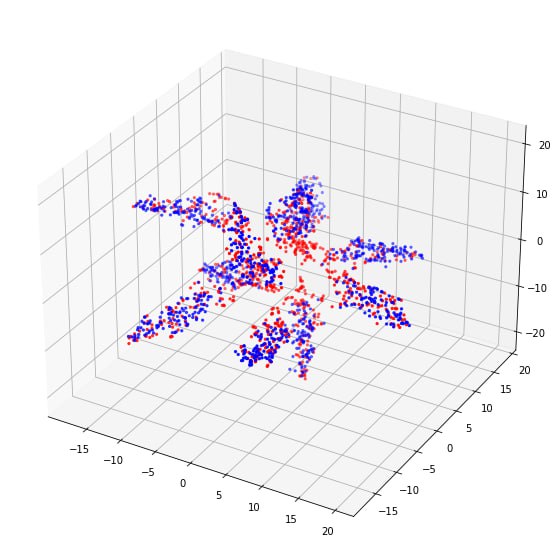}}
\subfigure[Utility w. defense (77.2\%)]
{\centering\includegraphics[width=0.22\linewidth]{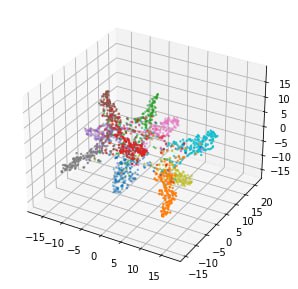}}
		\subfigure[MIA Acc w. defense (51.1\%)]
{\centering\includegraphics[width=0.22\linewidth]{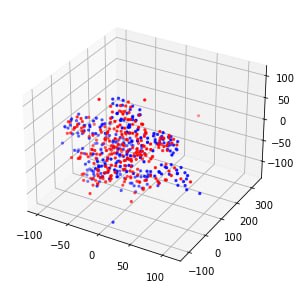}}
	\vspace{-4mm}
	\caption{\texttt{\name} against MIAs: 3D t-SNE embeddings results on the learnt representation of on CIFAR10.} 
	\label{fig:vis_cifar10}
	\vspace{-4mm}
\end{figure*}

\begin{figure*}[!t]
\vspace{-2mm}
	\centering
	\subfigure[Utility w/o. defense (81.7\%)]
{\centering\includegraphics[width=0.22\linewidth]{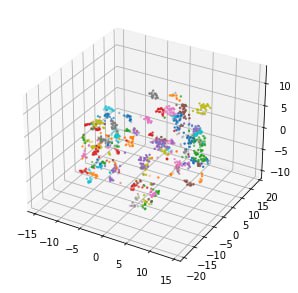}}
	\subfigure[MIA Acc w/o. defense (68.4\%)]
{\centering\includegraphics[width=0.22\linewidth]{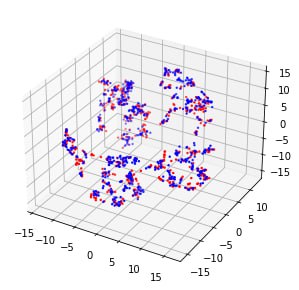}}
\subfigure[Utility w. defense (80\%)]
{\centering\includegraphics[width=0.22\linewidth]{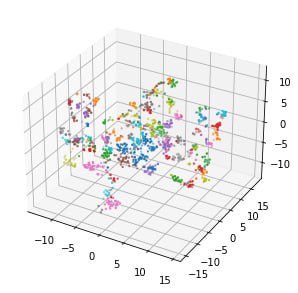}}
		\subfigure[MIA Acc w. defense (51\%)]
{\centering\includegraphics[width=0.22\linewidth]{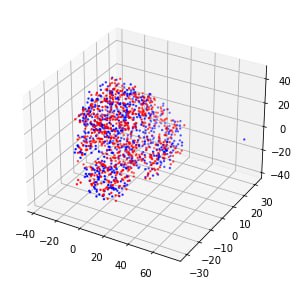}}
	\vspace{-4mm}
	\caption{\texttt{\name} against MIAs: 3D t-SNE embeddings results on the learnt representation on Purchase100.} 
	\label{fig:vis_purchase100}
	\vspace{-4mm}
\end{figure*}

\begin{figure*}[!t]
\vspace{-2mm}
	\centering
	\subfigure[Utility w/o. defense (49.8\%)]
{\centering\includegraphics[width=0.22\linewidth]{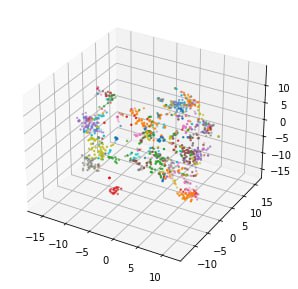}}
	\subfigure[MIA Acc w/o. defense (70.2\%)]
{\centering\includegraphics[width=0.22\linewidth]{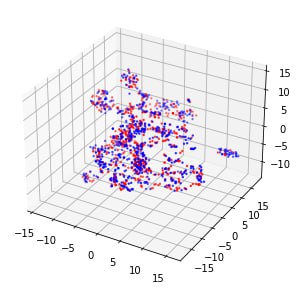}}
\subfigure[Utility w. defense (46\%)]
{\centering\includegraphics[width=0.22\linewidth]{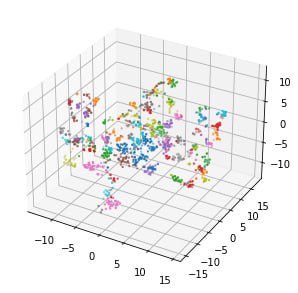}}
		\subfigure[MIA Acc w. defense (50\%)]
{\centering\includegraphics[width=0.22\linewidth]{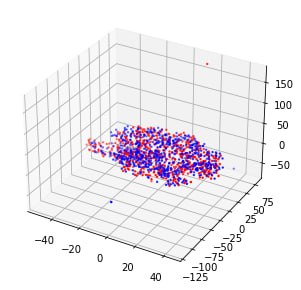}}
	\vspace{-4mm}
	\caption{\texttt{\name} against MIAs: 3D t-SNE embeddings results on the learnt representation on Texas100.} 
	\label{fig:vis_texas100}
	\vspace{-4mm}
\end{figure*}

\vspace{+0.01in}
\noindent {\bf Visualizing the learnt representations:} 
To better understand the learnt representations by \texttt{\name}, 
we adopt the t-SNE  algorithm~\cite{van2008visualizing} to visualize the low-dimensional embeddings of them. 
$\lambda$ is chosen in Table~\ref{tab:results_MIAs} that achieves the best utility-privacy tradeoff. 
We also compare with the case without privacy protection.  
Figures~\ref{fig:vis_cifar10}-\ref{fig:vis_texas100} show the 3D t-SNE embeddings, where each color corresponds to a label in the learning task or (non)member in the privacy task, and each point is a data sample. 
We can observe the t-SNE embeddings of the learnt representations without privacy protection for members and non-members are separated to some extent, meaning the  membership can be inferred via the learnt MIA classifier.  
On the contrary, the t-SNE embeddings of the learnt representations by our \texttt{\name} for members and non-members are mixed---hence making it difficult for the (best) MIA classifier to infer the membership from these learnt representations.

\begin{table}[!t]\renewcommand{\arraystretch}{0.85}
\footnotesize
\centering
\vspace{-3mm}
\caption{Comparing \texttt{\name} with existing defenses against MIAs on the three datasets. {DP methods are under the same/close defense performance as  \texttt{\name}.}
} 
 
\addtolength{\tabcolsep}{-4pt}
\centering
\begin{tabular}{|c|cc|cc|cc|}
\hline
\multirow{2}{*}{\bf Defense} & \multicolumn{2}{c|}{\bf CIFAR10}    & \multicolumn{2}{c|}{\bf Purchase100}    & \multicolumn{2}{c|}{\bf Texas100}    \\ \cline{2-7} 
& \multicolumn{1}{c|}{\bf Utility} & {\bf MIA Acc} & \multicolumn{1}{c|}{\bf Utility} & {\bf MIA Acc} & \multicolumn{1}{c|}{\bf Utility} & {\bf MIA Acc} \\ \hline
{\bf DP-SGD}          
& \multicolumn{1}{c|}{48\%} & 51\% & \multicolumn{1}{c|}{40\%} & 52\% & \multicolumn{1}{c|}{11\%} & 51\% \\ \hline
{\bf DP-enc}               & \multicolumn{1}{c|}{45\%} & 51\% & \multicolumn{1}{c|}{32\%} & 51\% & \multicolumn{1}{c|}{10\%} & 50\% \\ \hline
{\bf AdvReg}                 & \multicolumn{1}{c|}{75\%} & 53\% & \multicolumn{1}{c|}{75\%} & 51\% & \multicolumn{1}{c|}{44\%} & 52\% \\ \hline
{\bf NeuGuard}                 & \multicolumn{1}{c|}{74\%} & 56\% & \multicolumn{1}{c|}{77\%} & 53\% & \multicolumn{1}{c|}{43\%} &  52\% \\ \hline
{\texttt{\name}}                 & \multicolumn{1}{c|}{77\%} & 51\% & \multicolumn{1}{c|}{80\%} & 51\% & \multicolumn{1}{c|}{46\%} &  50\% \\ \hline
\end{tabular}
\label{tab:results_MIAs_comp}
\vspace{-6mm}
\end{table}

\begin{figure*}[!t]
\vspace{-2mm}
	\centering
\subfigure[CIFAR10]{\centering\includegraphics[width=0.29\linewidth]{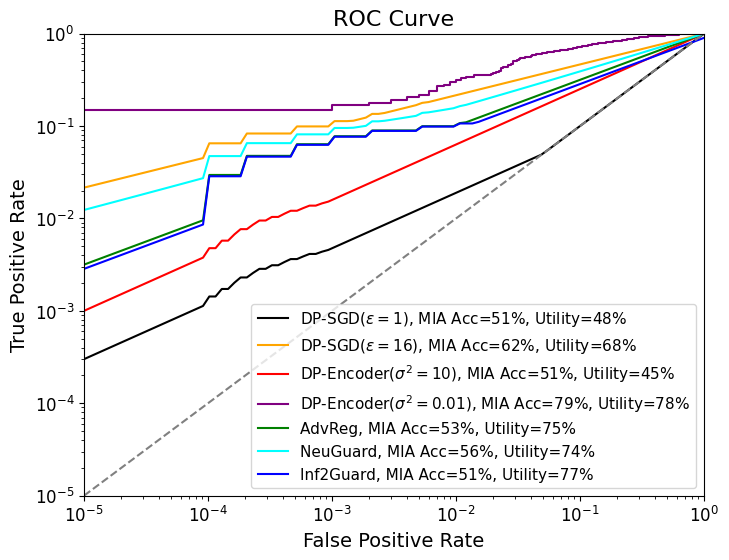}}
\subfigure[Purchase100]{\centering\includegraphics[width=0.29\linewidth]{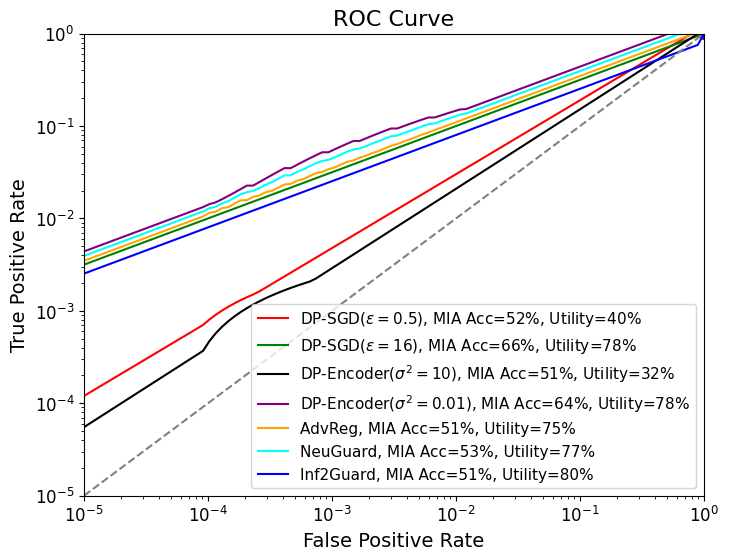}}
\subfigure[Texas100]{\centering\includegraphics[width=0.29\linewidth]{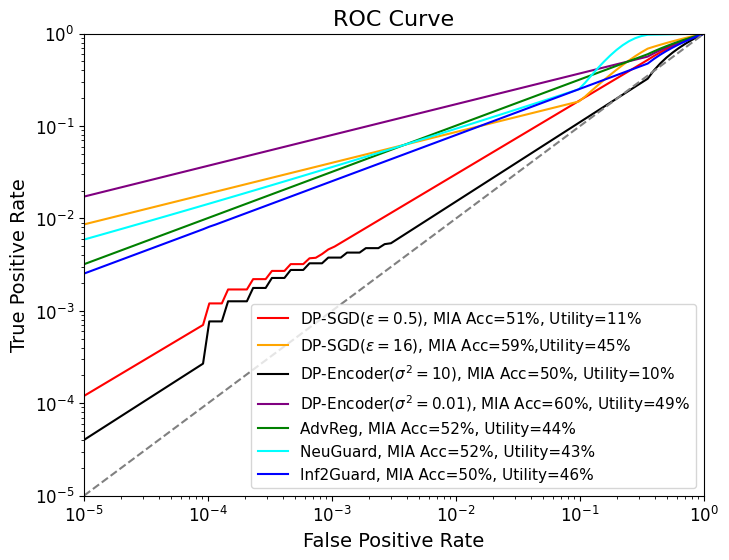}}
	\vspace{-4mm}
	\caption{{Comparing \texttt{\name} ($\lambda=0.75$) with the existing defenses against LiRA.} } 
	\label{fig:compareLiRA}
	\vspace{-4mm}
\end{figure*}

\vspace{-1mm}
\vspace{+0.01in}
\noindent {\bf Comparing with the existing defenses against MIAs:}  
All empirical defenses 
are broken by stronger  attacks~\cite{choquette2021label,song2021systematic}, except adversarial training-based AdvReg~\cite{nasr2018machine} (a special case of \texttt{\name}). NeuGuard~\cite{xu2022neuguard} is a recent empirical defense and shows better performance than, e.g., \cite{jia2019memguard,shokri2015privacy}.  Differential privacy 
is the only defense with privacy guarantees. We propose to use two DP variants, i.e., DP-SGD~\cite{abadi2016deep} and DP-encoder (details in Appendix~\ref{app:setup}). 
The comparison results of these defenses are shown in Table~\ref{tab:results_MIAs_comp} (more DP results in Table~\ref{tab:moreDPres} in Appendix~\ref{app:moreresults}) {and Figure~\ref{fig:compareLiRA}}. 
{From Table~\ref{tab:results_MIAs_comp},} we observe DP methods have bad utility when ensuring the same level {defense performance} (w.r.t. MIA accuracy) as \texttt{\name}.   
AdvReg and NeuGuard also perform 
worse than \texttt{\name}. 

{Figure~\ref{fig:compareLiRA} shows the TPR vs. FPR of these defenses against LiRA under the results in Table~\ref{tab:results_MIAs_comp}. For DP methods, we also plot the TPR vs FPR when their utility is close to \texttt{\name}.  
With an MIA accuracy close to random guessing (but low utility), 
we see DP methods have the smallest TPR at a given low FPR. This means DP methods can most reduce the attack effectiveness of LiRA, which is also verified in \cite{carlini2022membership}. 
However, if DP methods have a close utility as \texttt{\name}, their TPRs are much higher than  \texttt{\name}'s at a low FPR.
Besides, \texttt{\name} has smaller TPRs than AdvReg and NeuGuard.}

\vspace{+0.01in}
\noindent {{\bf Overhead comparison:} All MIA defenses train a task classifier. 
AdvReg trains a task classifier and membership inference network. 
NeuGuard trains a task classifier with two regularizations. 
DP-SGD trains the task classifier on noisy models, while DP-encoder normally trains the encoder first and then trains the utility network on (Gaussian) noisy  representations. 
In the experiments, we define the task classifier of the compared defenses as the concatenation of our encoder and utility network. 
In our platform (NVIDIA GeForce RTX 3070 Ti), it took \texttt{\name} (72,7,6), AdvReg (66,6,6), NeuGuard (62,5,5), DP-SGD (60,4,3) and DP-encoder (59,4,3) seconds to run each  iteration on the three datasets, respectively\footnote{We have similar conclusions on defending against the other two attacks.}.}

\subsection{Defense Results on PIAs}

\subsubsection{Experimental setup}

{\bf Datasets:} Following  
recent works~\cite{suri2022formalizing,chaudhari2023snap}, we use 
three datasets (Census~\cite{suri2022formalizing}, RSNA~\cite{suri2022formalizing}, and CelebA\cite{liu2015faceattributes}) and treat the  female ratio 
as the private dataset property.

\vspace{+0.01in}
\noindent {\bf Defense/attack training and testing:}
We first predefine a (different) female ratio set in each dataset. 
For each female ratio, we generate a number of subsets
from the training set and test set with different subset sizes. 
The generated training/test subsets  
and all data samples in these subsets are treated as the attack training/test set and the utility training/test set, respectively. More details are in Table~\ref{tbl:dataset_PIAs} in Appendix~\ref{app:setup}. 

\begin{itemize}[leftmargin=*]
\vspace{-2mm}
\item {\bf Defense training:} We use the utility training set and attack training set to train the encoder, utility preservation network, and property protection network simultaneously. Then, the learnt encoder is frozen and published as an API. 

\vspace{-2mm}
\item {\bf Attack training:} 
We mimic the strongest possible PIA, where the attacker knows the exact property protection network, the aggregator, and attack training set used in defense training. Specifically, s/he feeds each subset in the attack training set to the  learn encoder to get the subset representation, applies the aggregator to obtained the aggregated representation, and trains the PIA classifier (same as the property protection network) on these aggregated representations to maximally infer the private female ratio.   

\vspace{-2mm}
\item {\bf Defense/attack testing:}
We utilize the utility test set to obtain the utility via querying the trained encoder and utility network, and the attack test set to obtain the PIA accuracy via querying the trained encoder and trained PIA classifier. 

\vspace{-2mm}
\end{itemize}

\begin{table}[!t]\renewcommand{\arraystretch}{0.9}
\vspace{-2mm}
\footnotesize
\caption{\texttt{\name} results against PIAs with a {\bf mean}-aggregation. $\lambda=0$ means no privacy protection, while $\lambda=1$ means no utility preservation. Random guessing PIA accuracy on the three datasets are 25\%, 14.3\%, and 9.1\%, respectively.} 
\addtolength{\tabcolsep}{-3.5pt}
\centering
\begin{minipage}{.32\linewidth}
\begin{tabular}{ccc}
\toprule
\multicolumn{3}{c}{\bf Census}\\
          \hline
\( \lambda \) & {\bf Utility} & {\bf PIA Acc} \\
\midrule
0 & 85\% & 68\% \\
0.25 & 80\% & 61\% \\
0.5 & 78\% & 52\% \\
0.75 & 76\% & 34\% \\
1 & 45\% & 26\% \\
\bottomrule
\end{tabular}
\end{minipage}
\hfill
\begin{minipage}{.32\linewidth}
\begin{tabular}{ccc}
\toprule
\multicolumn{3}{c}{\bf RSNA}\\
          \hline
\( \lambda \) & {\bf Utility} & {\bf PIA Acc} \\
\midrule
0 & 83\% & 52\%  \\
0.25 & 82\% & 25\% \\
0.5 & 82\% & 24\% \\
0.75 & 80\% & 19\% \\
1 & 50\% & 15\% \\
\bottomrule
\end{tabular}
\end{minipage}
\hfill
\begin{minipage}{.32\linewidth}
\begin{tabular}{ccc}
\toprule
\multicolumn{3}{c}{\bf CelebA}\\
          \hline
\( \lambda \) & {\bf Utility} & {\bf PIA Acc} \\
\midrule
0 & 91\% & 50\%  \\
0.25 & 91\% & 28\% \\
0.5 & 91\% & 17\% \\
0.75 & 89\% & 11\% \\
1 & 53\% & 10\% \\
\bottomrule
\end{tabular}
\end{minipage}
\label{tab:results_PIAs_mean}
\vspace{-2mm}
\end{table}

\subsubsection{Experimental results}
\label{sec:results_PIAs}

{\bf Utility-privacy results:}
Table~\ref{tab:results_PIAs_mean} 
shows the utility-privacy results of \texttt{\name}, 
where the encoder uses a mean-aggregator (i.e., average the representations of a subset of data. Note different subsets have different sizes). 
We have similar observations as in defending against MIAs:    
1) The PIA accuracy can be as large as 68\% without privacy protection ($\lambda=0$ in Equation (\ref{eqn:task_known_PIA_Guard})), implying the PIA is effective; 
2) When focusing on protecting privacy  ($\lambda=1$), the PIA performance can be largely reduced.  
However, the utility is also significantly decreased, e.g., from 85\% to 45\%.  
3) Utility and privacy show a tradeoff w.r.t. $0<\lambda<1$. In most of the cases, the best tradeoff is obtained when $\lambda=0.75$. Again, the results show the learnt privacy-preserving encoder/representations are effective against PIAs, and also maintain utility.

\begin{table}[!t]\renewcommand{\arraystretch}{0.9}
\vspace{-2mm}
\footnotesize
\caption{\texttt{\name} results against PIAs with a {\bf max}-aggregation. Random guessing PIA accuracy on the three datasets are 25\%, 14.3\%, and 9.1\%, respectively.} 
\addtolength{\tabcolsep}{-3.5pt}
\centering
\begin{minipage}{.32\linewidth}
\begin{tabular}{ccc}
\toprule
\multicolumn{3}{c}{\bf Census}\\
          \hline
\( \lambda \) & {\bf Utility} & {\bf PIA Acc} \\
\midrule
0     & 85\% & 65\%  \\
0.25  & 83\% & 51\% \\
0.5   & 79\% & 49\% \\
0.75  & 77\% & 37\% \\
1     & 47\% & 30\% \\
\bottomrule
\end{tabular}
\end{minipage}
\hfill
\begin{minipage}{.32\linewidth}
\begin{tabular}{ccc}
\toprule
\multicolumn{3}{c}{\bf RSNA}\\
          \hline
\( \lambda \) &  {\bf Utility} & {\bf PIA Acc} \\
\midrule
0     & 83\% & 52\%  \\
0.25  & 82\% & 24\% \\
0.5  & 82\% & 24\% \\
0.75  & 81\% & 21\% \\
1      & 50\% & 16\% \\
\bottomrule
\end{tabular}
\end{minipage}
\hfill
\begin{minipage}{.32\linewidth}
\begin{tabular}{ccc}
\toprule
\multicolumn{3}{c}{\bf CelebA}\\
          \hline
\( \lambda \) & {\bf Utility} & {\bf PIA Acc} \\
\midrule
0     & 91\% & 50\%  \\
0.25  & 91\% & 25\% \\
0.5  & 91\% & 15\% \\
0.75  & 88\% & 15\% \\
1  & 53\% & 9\% \\
\bottomrule
\end{tabular}
\end{minipage}
\label{tab:results_PIAs_max}
\vspace{-4mm}
\end{table}

\begin{table}[!t]\renewcommand{\arraystretch}{0.9}
\footnotesize
\centering
\caption{Comparing \texttt{\name} with DP against PIAs.} 
\addtolength{\tabcolsep}{-3pt}
\centering
\begin{tabular}{|c|cc|cc|cc|}
\hline
\multirow{2}{*}{\bf Defense} & \multicolumn{2}{c|}{\bf Census}    & \multicolumn{2}{c|}{\bf RSNA}    & \multicolumn{2}{c|}{\bf CelebA}    \\ \cline{2-7} 
& \multicolumn{1}{c|}{\bf Utility} & {\bf PIA Acc} & \multicolumn{1}{c|}{\bf Utility} & {\bf PIA Acc} & \multicolumn{1}{c|}{\bf Utility} & {\bf PIA Acc} \\ \hline
{\bf DP-encoder}               & \multicolumn{1}{c|}{52\%} & 34\% & \multicolumn{1}{c|}{57\%} & 19\% & \multicolumn{1}{c|}{66\%} & 11\% \\ \hline
                
{\texttt{\name}}                 & \multicolumn{1}{c|}{76\%} & 34\% & \multicolumn{1}{c|}{80\%} & 19\% & \multicolumn{1}{c|}{89\%} &  11\% \\ \hline
\end{tabular}
\label{tab:results_PIAs_comp}
\vspace{-4mm}
\end{table}

\vspace{+0.01in}
\noindent {\bf Visualizing the learnt representations:} 
Figures~\ref{fig:vis_census}-\ref{fig:vis_celebA} in Appendix~\ref{app:moreresults} show the 3D t-SNE embeddings of the learnt representations with $\lambda=0, 0.75$. 
Similarly, we observe the t-SNE embeddings of the aggregated representations without privacy protection can be separated to a large extent, while those with privacy protection by \texttt{\name} are mixed. 
This again verifies it is  difficult for the (best) PIA to infer the private female ratio from the representations learnt by  \texttt{\name}.

\vspace{+0.01in}
\noindent {\bf Impact of the aggregator used by the encoder:} 
In this experiment, we test the impact of the aggregator and choose a max-aggregator for evaluation, where we select the element-wise maximum value of the representations of each subset of data.  Table~\ref{tab:results_PIAs_max} shows the results. We have similar conclusions as those with the mean-aggregator. In addition, \texttt{\name} with the max-aggregator has slightly worse utility-privacy tradeoff, compared with the mean-aggregator. A possible reason could be the mean-aggregator uses more information of the subset representations than the max-aggregator.

\vspace{+0.01in} 
\noindent {\bf Comparing with the DP-based defense:}
There exists no effective defense against PIAs, and \cite{suri2022formalizing} shows DP-SGD~\cite{abadi2016deep} does not work well. Here, we propose to use a DP variant called DP-encoder, similar to that against MIAs. More details about DP-encoder are in Appendix~\ref{app:setup}.   
The compared results are shown in Table~\ref{tab:results_PIAs_comp}. We can see that, with the same level privacy protection as \texttt{\name}, DP has much worse utility.

\subsection{Defense Results on DRAs}

\subsubsection{Experimental setup}
\vspace{-1mm}

{\bf Datasets:} We select two  image datasets: CIFAR10~\cite{Krizhevsky09learningmultiple} and CIFAR100~\cite{Krizhevsky09learningmultiple}, and one human activity recognition dataset Activity~\cite{misc_human_activity_recognition_using_smartphones_240} to evaluate \texttt{\name} against DRAs.

\vspace{+0.01in}
\noindent {\bf Defense/attack training and testing:}
Table~\ref{tbl:dataset_DRAs} in Appendix shows the statistics of the utility/attack training and test sets. 

\begin{itemize}[leftmargin=*]
\vspace{-2mm}
\item {\bf Defense training:} 
We use the training set to train the encoder, utility preservation network, reconstruction protection network, and update the perturbation distribution parameters, simultaneously. Then, the learnt encoder and perturbation distribution are 
published.

\vspace{-2mm}
\item {\bf Attack training:} 
We mimic the strongest DRA, where the attacker knows the  reconstruction protection network, training set, {and perturbation distribution}. S/he feeds each training data to the learnt encoder + perturbation distribution to get the 
perturbed representation. 
Then the attacker trains 
the reconstruction network (using the pair of input data and its perturbed representation) to infer the training data. 

\vspace{-2mm}
\item {\bf Defense/attack testing:}
We use the utility test set to obtain the utility via querying the encoder 
and utility network; and use the attack test set to obtain the DRA performance by querying the trained encoder and  reconstruction network. 
\vspace{-2mm}
\end{itemize}

\noindent {\bf Privacy metric:} 
  For image datasets, we use the common Structural Similarity Index Measure (SSIM) and PSNR metrics~\cite{he2019model}. 
A larger SSIM (or PSNR) between two images indicate they  look more similar. An effective attack aims to achieve a large SSIM (or PSNR), while the defender does the opposite. 
For human activity dataset,  we use the mean-square error (MSE) between two samples to measure similarity. 
 A smaller/larger MSE indicates a more effective attack/defense.

\subsubsection{Experimental results}
\label{sec:results_DRAs}

{\bf Utility-privacy results:} 
Table~\ref{tbl:results_DRAs} shows the defense results of \texttt{\name} with the Gaussian perturbation distribution,  
where $\lambda=0.4$ in Equation (\ref{eqn:task_known_DRA_Guard}). 
We can observe $\epsilon$ acts a utility-privacy tradeoff. A larger $\epsilon$ implies adding more perturbation to the  representation during defense training. This makes the DRA more challenging, but also sacrifice the utility more.     

We also test the impact of $\lambda$ and the results are shown in Table~\ref{tab:impact_lambda_DRAs}. 
We can see $\lambda$ also acts as a tradeoff---a larger $\lambda$ can protect data privacy more, while having larger utility loss.

\begin{table}[!t]
\vspace{-2mm}
\caption{\texttt{\name} results against DRAs. A smaller SSIM or PSNR indicates better defense performance (\(\lambda\) =0.4). }
\addtolength{\tabcolsep}{-5pt}
\centering
\footnotesize
\begin{minipage}{0.27\linewidth}
\begin{tabular}{ccc}
\toprule
\multicolumn{3}{c}{\bf CIFAR10}\\
          \hline
 Scale $\epsilon$ & Utility & SSIM/PSNR\\
\midrule
0 & 89.5\% & 0.78 / 15.97\\ 
\hline
0.75 & 85.2\% & 0.42 / 12.09\\
1.25 & {\bf 78.0\%} & {\bf 0.21 / 11.87}\\
1.75 & 68.9\% & 0.17 / 11.21\\
\bottomrule
\end{tabular}
\end{minipage}
\hfill
\begin{minipage}{0.27\linewidth}
\begin{tabular}{ccc}
\toprule
\multicolumn{3}{c}{\bf CIFAR100}\\
          \hline
 $\epsilon$ &  Utility & SSIM/PSNR\\
\midrule
0     & 52.7\% & 0.92 / 22.79\\ 
\hline
0.75  &  49.1\% & 0.36 / 13.36\\
1.00  &  {\bf 46.5\%} & {\bf 0.19 / 12.70}\\
1.25  &  43.3\% & 0.14 / 12.29\\
\bottomrule
\end{tabular}
\end{minipage}
\hfill
\begin{minipage}{0.2\linewidth}
\begin{tabular}{ccc}
\toprule
\multicolumn{3}{c}{\bf Activity}\\
          \hline
 $\epsilon$ & Utility & MSE\\
\midrule
0   & 95.1\% & 0.81\\ 
\hline
0.5     & 90.1\% & 1.06\\
1.0     & {\bf 85.6\%} & {\bf 1.32}\\
1.5     & 81.0\% & 1.64\\
\bottomrule
\end{tabular}
\end{minipage}
\label{tbl:results_DRAs}
\vspace{-4mm}
\end{table}

\begin{table}[!t]
\centering
\caption{Impact of $\lambda$ on \texttt{\name} against DRAs ($\epsilon=1.25$). 
}
\addtolength{\tabcolsep}{-5pt}
\centering
\small
\footnotesize
\begin{minipage}{0.27\linewidth}
\begin{tabular}{ccc}
\toprule
\multicolumn{3}{c}{\bf CIFAR10}\\
          \hline
 $\lambda$& Utility & SSIM/PSNR\\
\midrule
0.1 & 83.2\% & 0.52 / 12.79\\
0.4 & 78.0\% & 0.21 / 11.87\\
0.7 & 67.9\% & 0.15 / 11.43\\
\bottomrule
\end{tabular}
\end{minipage}
\hfill
\begin{minipage}{0.27\linewidth}
\begin{tabular}{ccc}
\toprule
\multicolumn{3}{c}{\bf CIFAR100}\\
          \hline
 $\lambda$& Utility & SSIM/PSNR\\
\midrule
0.1 & 46.8\% & 0.46 / 14.75\\
0.4 & 46.5\% & 0.21 / 12.70\\
0.7 & 46.5\% & 0.20 / 12.42\\
\bottomrule
\end{tabular}
\end{minipage}
\hfill
\begin{minipage}{0.2\linewidth}
\begin{tabular}{ccc}
\toprule
\multicolumn{3}{c}{\bf Activity}\\
          \hline
$\lambda$ & Utility & MSE\\
\midrule
0.1 &90.0\% & 1.25\\
0.4 &85.6\% & 1.32\\
0.7 &85.0\% & 1.62\\
\bottomrule
\end{tabular}
\end{minipage}
\label{tab:impact_lambda_DRAs}
\vspace{-4mm}
\end{table}

\begin{table}[!t]
\vspace{-2mm}
\caption{DP-SGD defense results against DRAs. A smaller SSIM or PSNR indicates better defense performance. 
}
\addtolength{\tabcolsep}{-5pt}
\centering
\footnotesize
\begin{minipage}{0.27\linewidth}
\begin{tabular}{ccc}
\toprule
\multicolumn{3}{c}{\bf CIFAR10}\\
          \hline
 Scale $\epsilon$  & Utility & SSIM/PSNR\\
\midrule
0 & 89.5\% & 0.78 / 15.97\\ 
\hline
0.75 & 85.6\% & 0.50 / 13.21\\
1.25 & {\bf 77.4\%} & {\bf 0.37 / 12.45}\\ 
1.75 & 65.3\% & 0.36 / 12.35\\
\bottomrule
\end{tabular}
\end{minipage}
\hfill
\begin{minipage}{0.27\linewidth}
\begin{tabular}{ccc}
\toprule
\multicolumn{3}{c}{\bf CIFAR100}\\
          \hline
 $\epsilon$ &  Utility & SSIM/PSNR\\
\midrule
0     & 52.7\% & 0.92 / 22.79\\ 
\hline
0.75  &  49.5\% & 0.43 / 13.77\\
1.00  &  {\bf 46.7\%} & {\bf 0.24 / 12.87}\\
1.25  &  43.3\% & 0.18 / 12.54\\
\bottomrule
\end{tabular}
\end{minipage}
\hfill
\begin{minipage}{0.22\linewidth}
\begin{tabular}{ccc}
\toprule
\multicolumn{3}{c}{\bf Activity}\\
          \hline
 $\epsilon$  & Utility & MSE\\
\midrule
0   & 95.1\% & 0.81\\ 
\hline
0.5     & 91.8\% & 0.88\\
1.0     & 90.1\% & 0.90\\
1.5     & {\bf 84.1\%} & {\bf 1.01}\\
\bottomrule
\end{tabular}
\end{minipage}
\label{tbl:results_DRAs_DP}
\vspace{-3mm}
\end{table}

\vspace{+0.01in}
\noindent {\bf Comparing with the DP-based defense:} 
All empirical defenses against DRAs are broken are by an advanced attack~\cite{balunovic2022bayesian}. 
A few papers~\cite{balle2022reconstructing,salem2023sok} show  if a randomized algorithm satisfies DP, it can defend against DRAs with provable guarantees. 
We compare \texttt{\name} with DP 
and  Table~\ref{tbl:results_DRAs_DP} shows the DP results.  
Viewing with results in Table~\ref{tbl:results_DRAs}, we  see \texttt{\name} obtains better utility-privacy tradeoffs than DP-SGD. 

\begin{figure}[!t]
	\centering
	\subfigure[Raw images]
{\centering\includegraphics[width=0.22\linewidth]{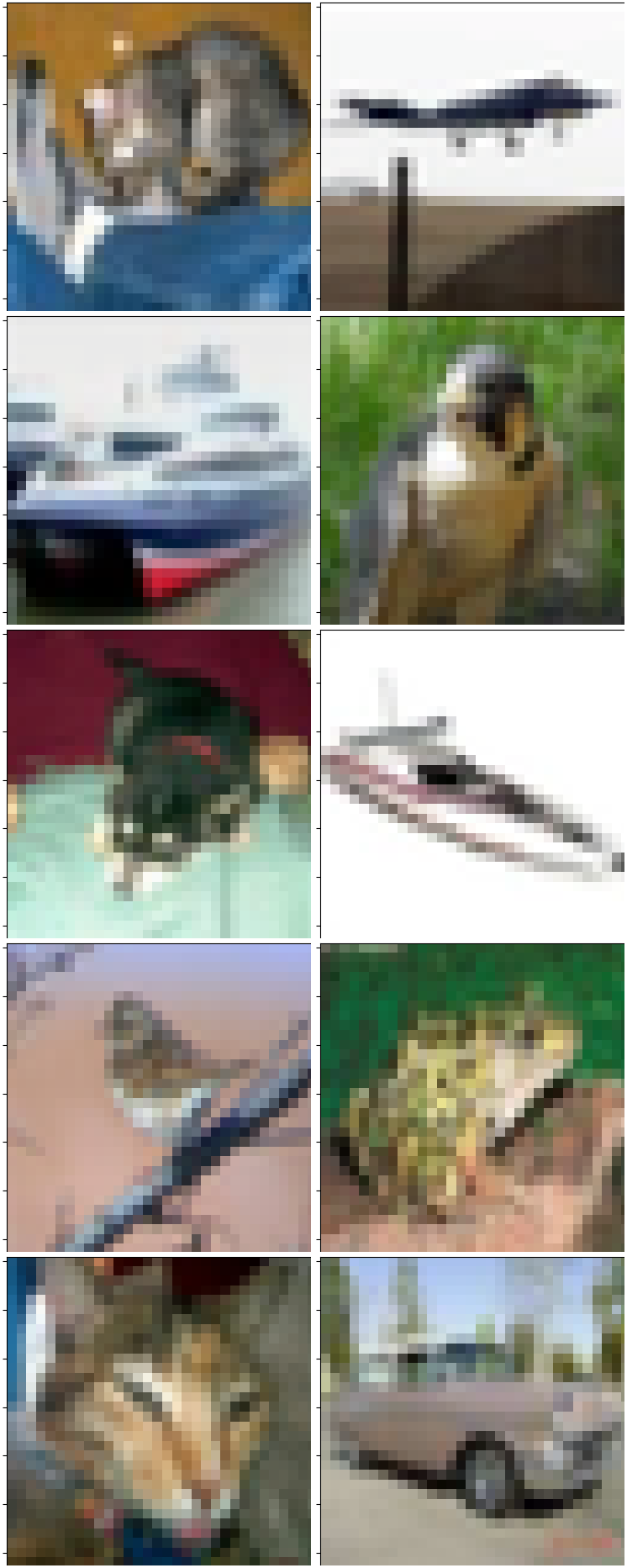}}
	\subfigure[No defense]
{\centering\includegraphics[width=0.22\linewidth]{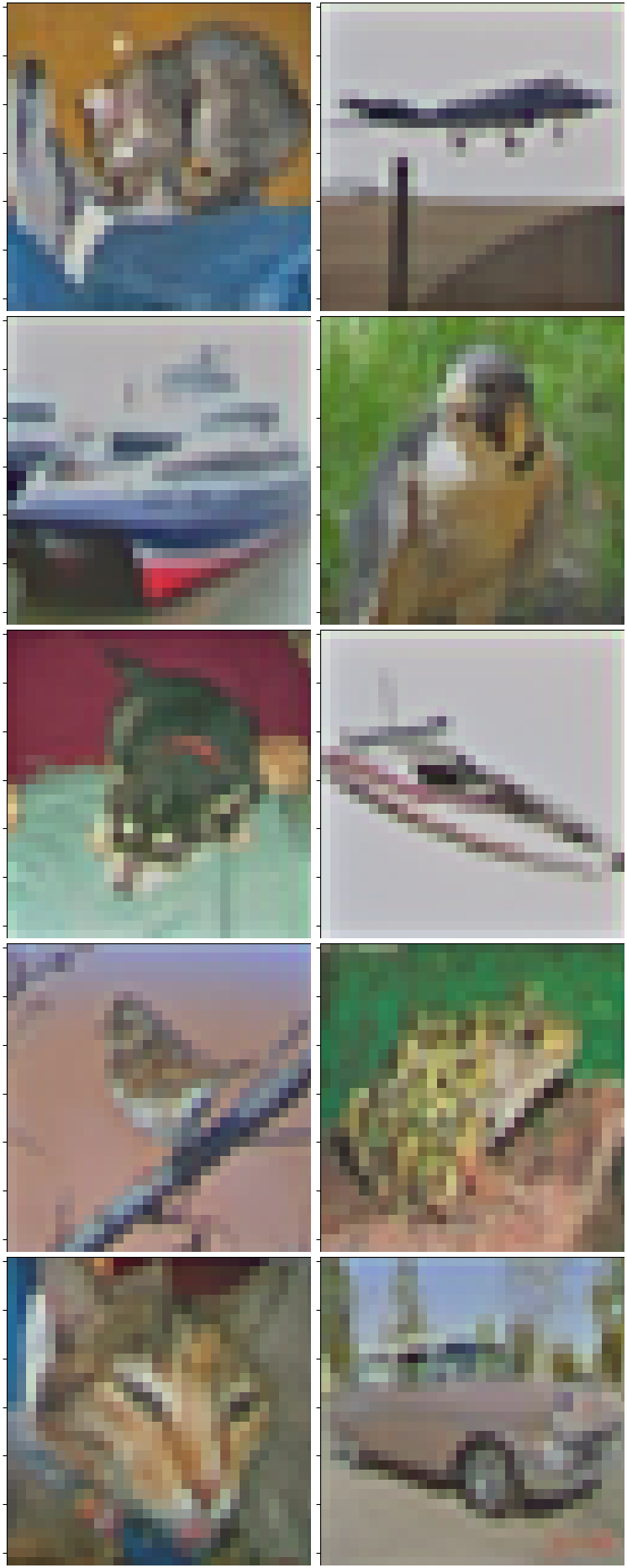}}
\subfigure[DP]
{\centering\includegraphics[width=0.22\linewidth]{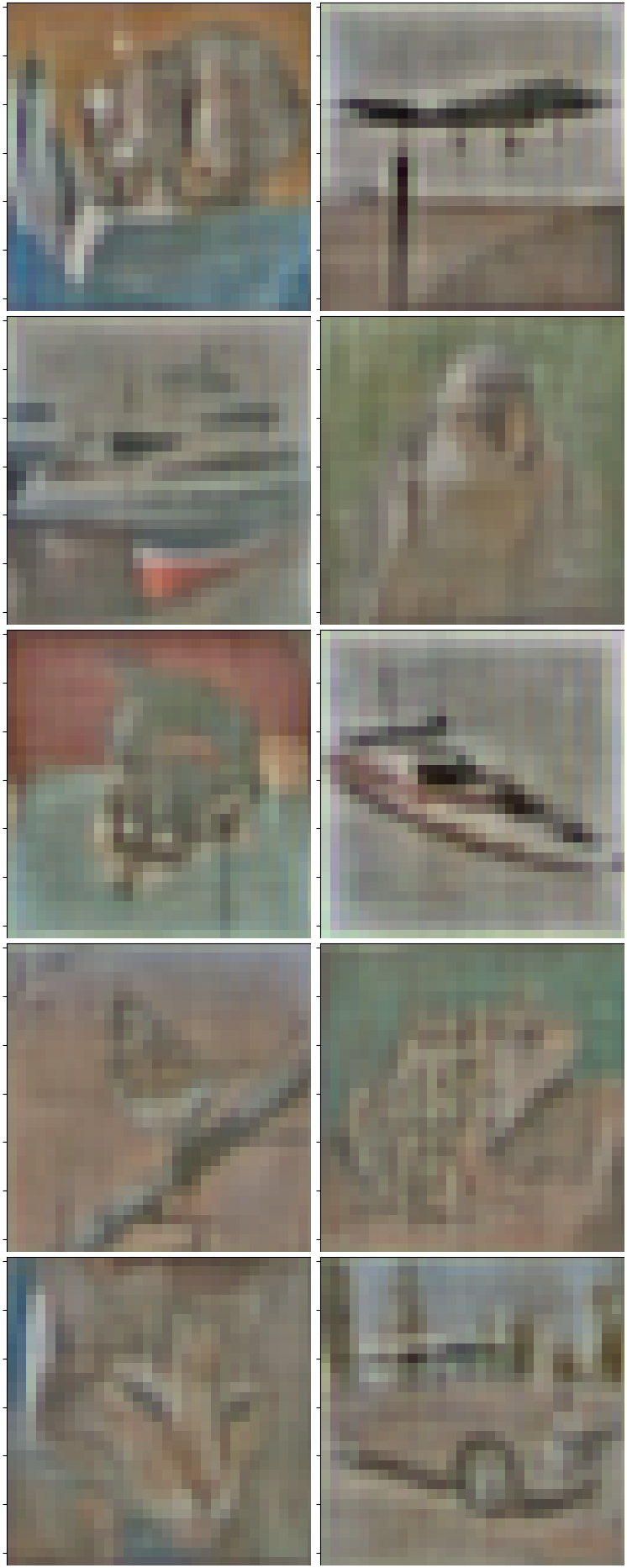}}
		\subfigure[\texttt{\name}]
{\centering\includegraphics[width=0.22\linewidth]{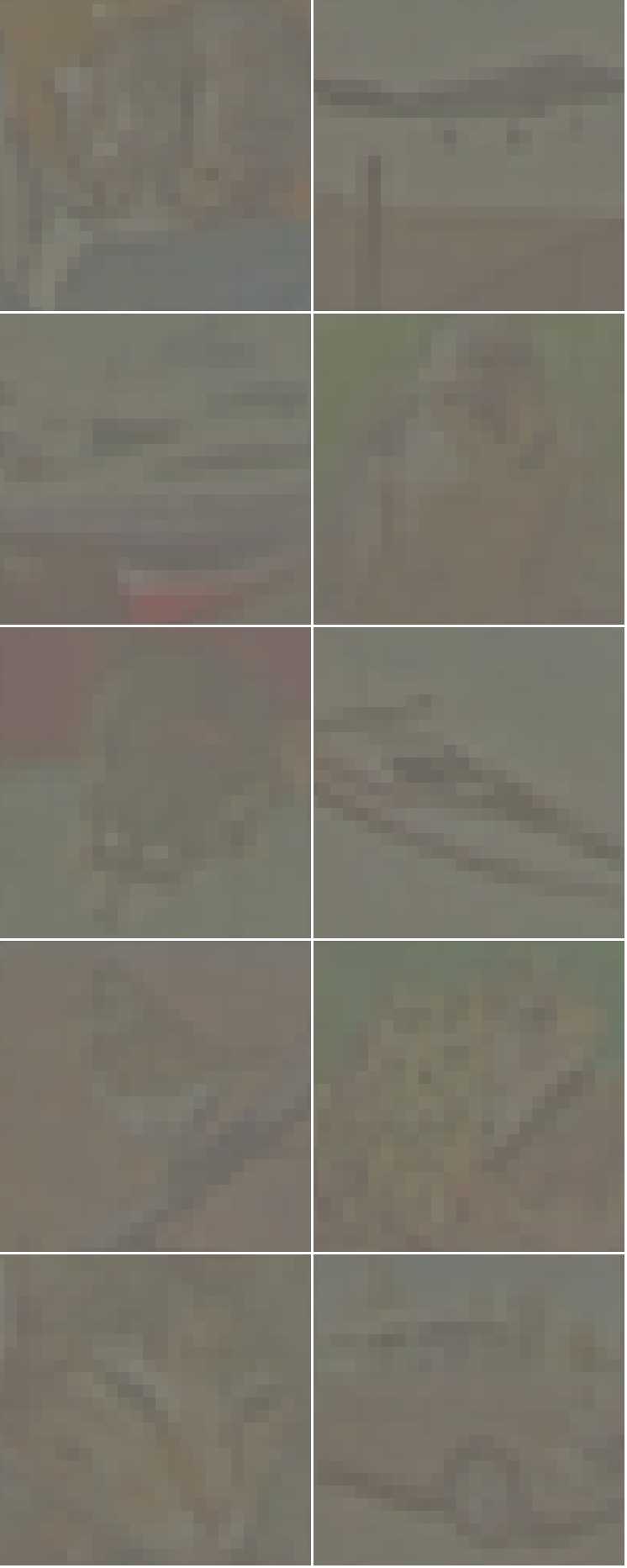}}
	\vspace{-4mm}
	\caption{Raw images vs. reconstructions on CIFAR10. DP Utility: 77\%,  \texttt{\name} Utility: 78\%.} 
	\label{fig:rec_cifar10}
	\vspace{-5mm}
\end{figure}

\begin{figure}[t]
	\centering
	\subfigure[Raw images]
{\centering\includegraphics[width=0.22\linewidth]{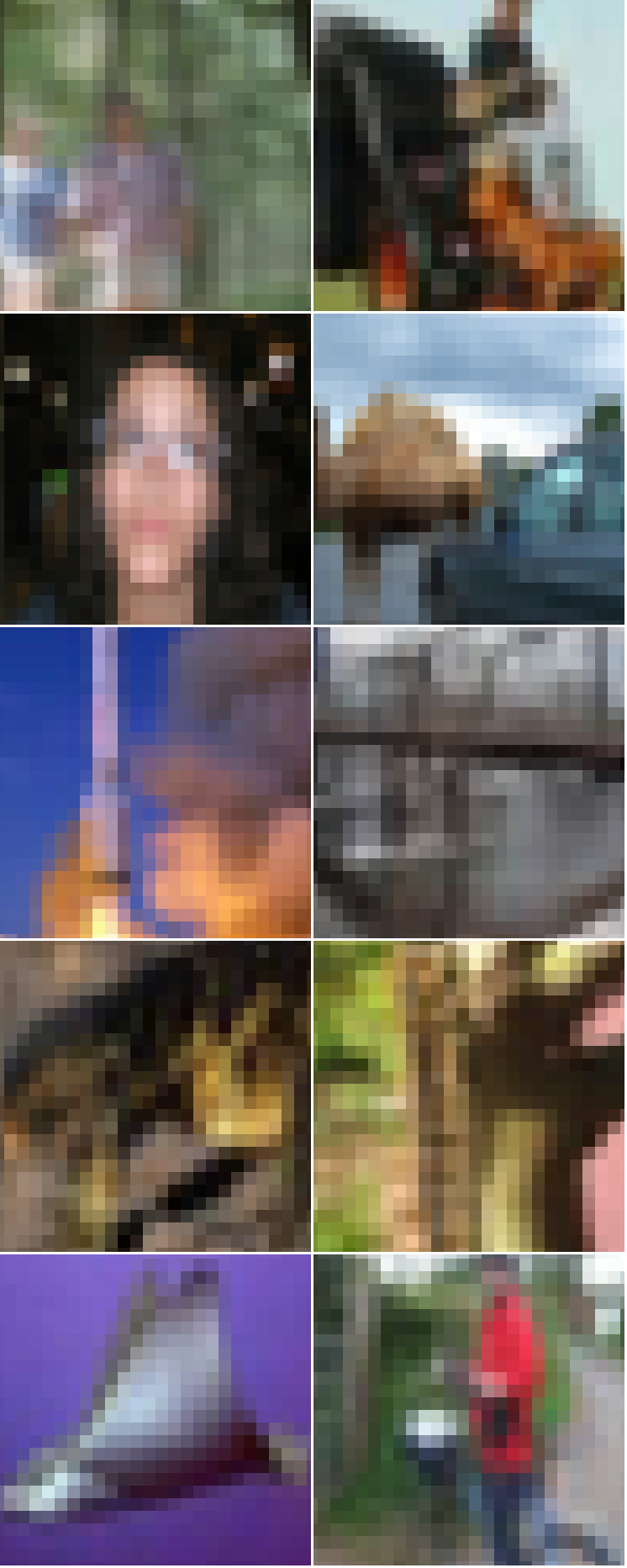}}
	\subfigure[No defense]
{\centering\includegraphics[width=0.22\linewidth]{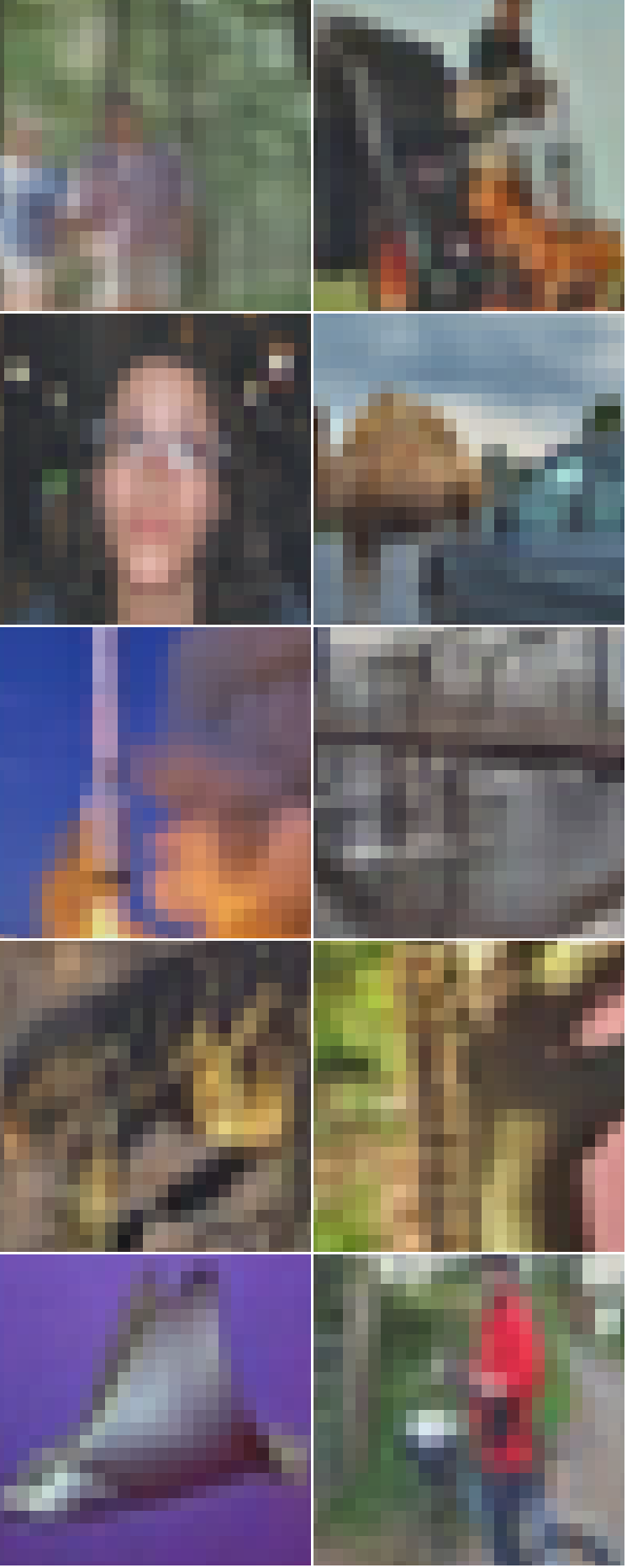}}
\subfigure[DP]
{\centering\includegraphics[width=0.22\linewidth]{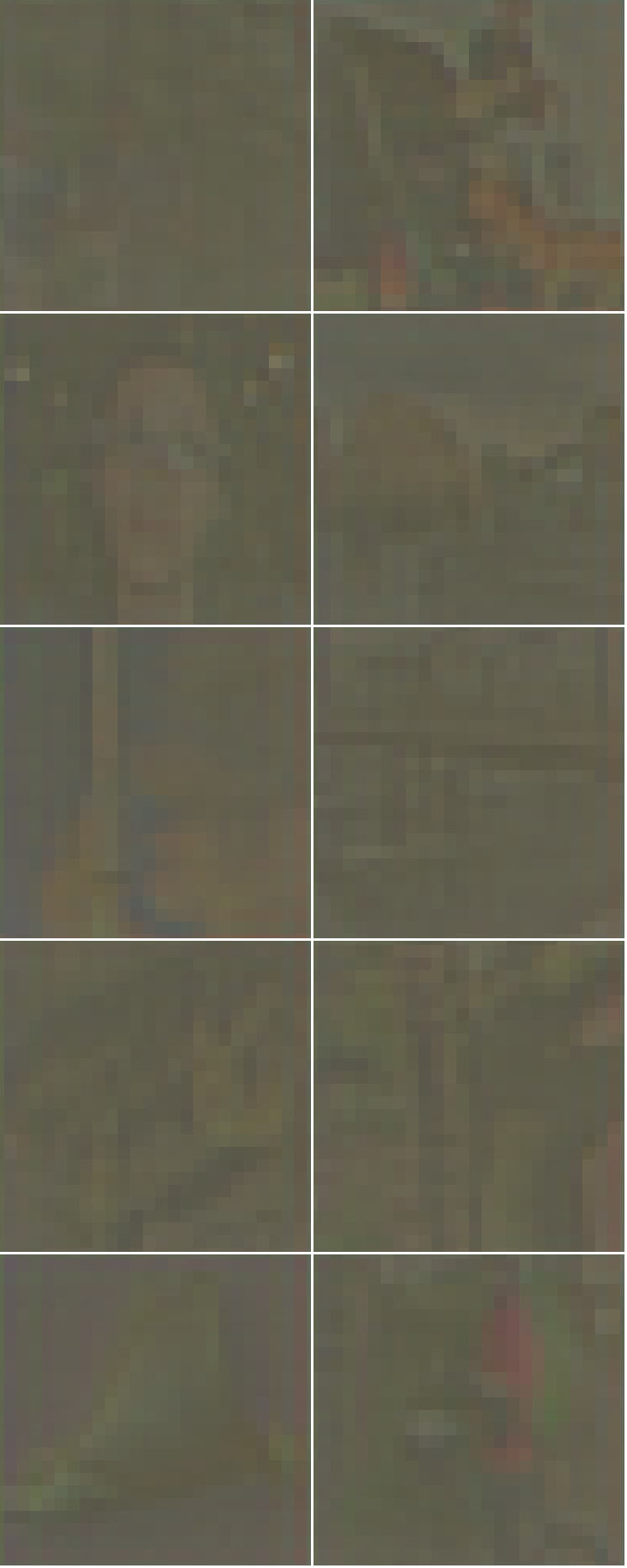}}
		\subfigure[\texttt{\name}]
{\centering\includegraphics[width=0.22\linewidth]{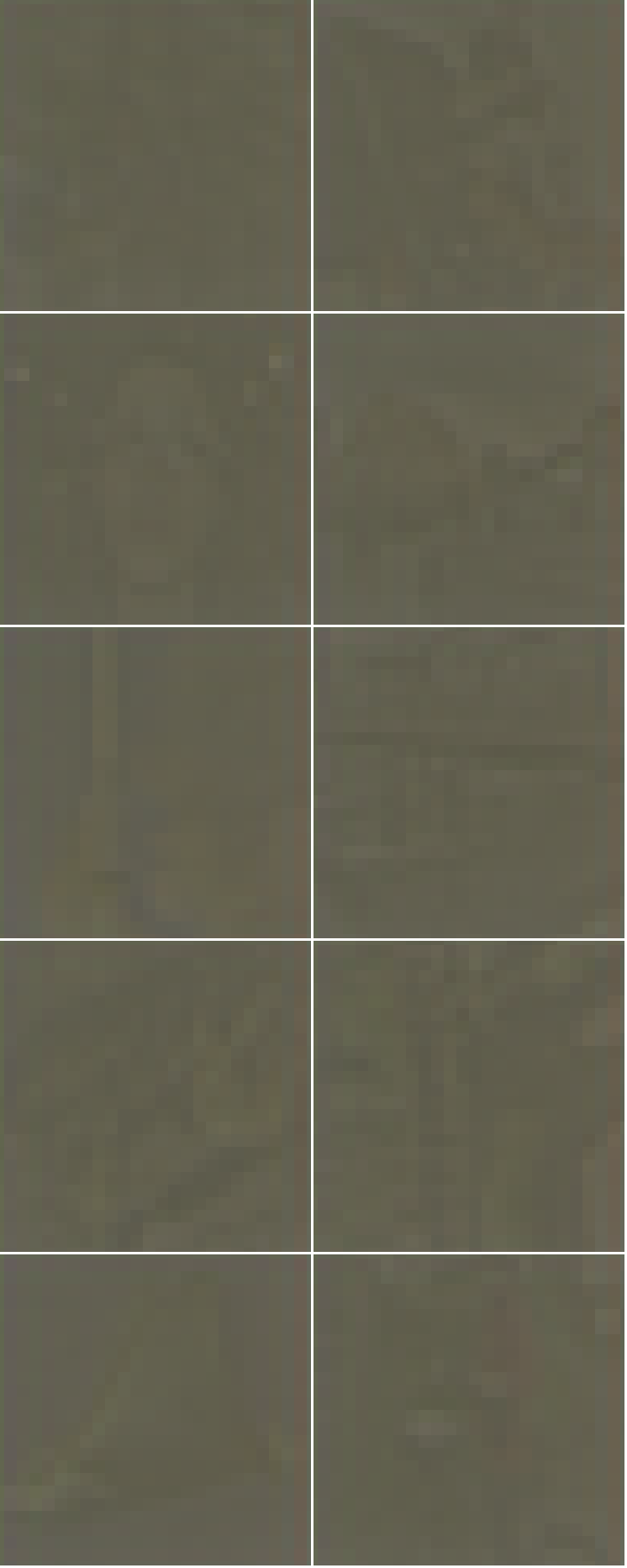}}
	\vspace{-4mm}
	\caption{Raw images vs. reconstructions on CIFAR100. DP Utility: 47\%,  \texttt{\name} Utility: 47\%. Better zoom in.} 
	\label{fig:rec_cifar100}
	\vspace{-4mm}
\end{figure}

\begin{figure}[!t]
    \centering
    \includegraphics[width=0.32\textwidth]{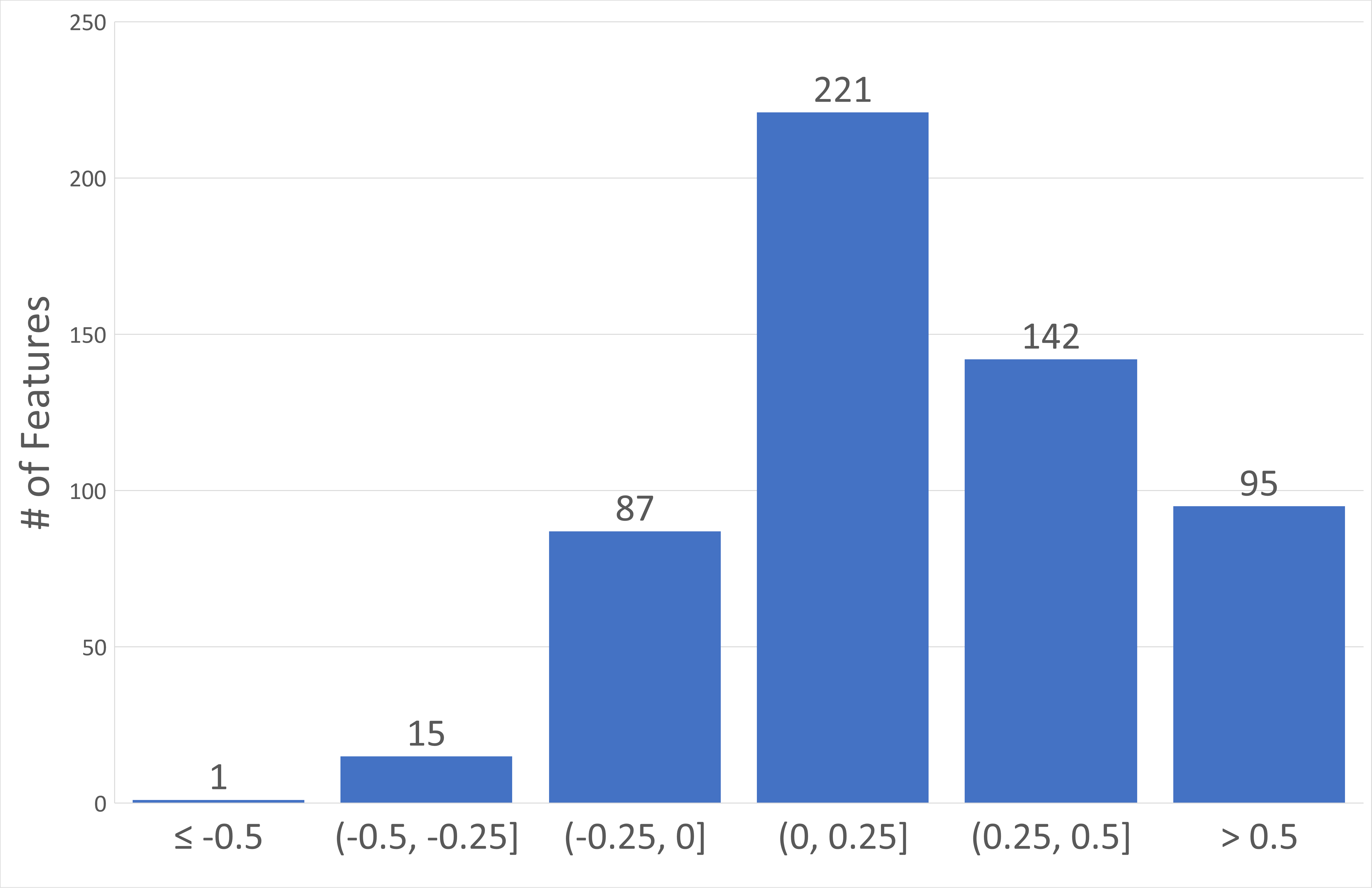}
   \vspace{-4mm}
    \caption{\texttt{\name} vs. DP on 50 samples in Activity (both utility=86\%). 
    We count \#features located in each bin. The value range in each bin means the difference between the reconstructed feature by \texttt{\name} and that by DP to the true feature. A positive value implies \texttt{\name} produces more dissimilar reconstruction than DP.    
    }
    \label{fig:rec_activity}
    \vspace{-4mm}
\end{figure}

\vspace{+0.01in}
\noindent {\bf Visualizing data reconstruction results:}
Figure~\ref{fig:rec_cifar10} and 
Figure~\ref{fig:rec_cifar100} show the reconstruction results on some CIFAR10 and CIFAR100 images, respectively. 
We see that, without defense, the attacker can accurately reconstruct the raw images. 
With a similar utility, visually, \texttt{\name} can better defend against image reconstruction than DP. 
Figure~\ref{fig:rec_activity} summarizes the reconstruction results on 50 samples in Activity, where we report the difference between each reconstructed feature by \texttt{\name} and that by DP to the true feature. A (larger) positive value implies \texttt{\name} is (more) dissimilar than DP to the true feature. We can see \texttt{\name} has better defense results than DP in most (413 out of 516) of the features.

\section{Discussion and Future Work}
\label{sec:discussion}
\vspace{-2mm}

{\textbf{\texttt{\name} and DP:} 
Essentially, 
\texttt{\name} and DP are two different provable privacy mechanisms, and they complement each other. First, DP mainly measures the user or sample-level privacy risks in the worst case while \texttt{\name} can accurately measure the average privacy risks at the dataset level with the derived bounds. Second, DP has been shown to provide some resilience transferability across some inference attacks~\cite{salem2023sok} (but not all of them). It is also interesting to study the resilience transferability for the proposed \texttt{\name}, which we will explore in the future. More importantly, our \texttt{\name} can complement DP. For instance, we can use the learnt (deterministic) data representations by \texttt{\name} as input to DP-SGD or add (Gaussian) noise to the representations to ensure DP guarantees against MIAs.}

\vspace{+0.01in}
\noindent {\bf Task-agnostic representation learning:} Our current MI formulation for utility preservation  knows the labels of the learning task (e.g., see Equation (\ref{eqn:MIA_maxutil_known})). A more promising solution  would be task-agnostic, i.e., learning task-agnostic representations that can benefit many (unknown) downstream tasks. We note that our framework can be easily extended to this scenario. For instance, in MIAs, we now require 
the learnt representation 
${\bf r}$ includes as much information about the \emph{training} sample  ${\bf x}$ as possible (i.e., $u=1$). Intuitively, when 
${\bf r}$ retains all information about 
${\bf x}$, the model trained on 
${\bf r}$ will have the same performance as trained on the raw 
${\bf x}$, despite the learning task. Formally, the MI objective 
becomes $\max_f I({\bf x} ; {\bf r}| u=1) $.

\vspace{+0.01in}
\noindent {{\bf Defending against multiple inference attacks simultaneously:} We design the customized MI objectives to defend against each inference attack in the paper. A natural 
solution to defend against multiple inference attacks is unifying their training objectives (by summarizing them with tradeoff hyperparameters). 
While this is possible, we emphasize that the learnt encoder is weak against all attacks. This is because the encoder should balance the defense effectiveness among these attacks, and cannot be optimal against all of them.}

\vspace{+0.01in}
\noindent {\bf Generalizing our theoretical results:} 
Our theoretical results assume the learning task is binary classification and dataset property is binary-valued. 
We will generalize our 
theoretical results to multiclass classification and other types of learning such as regression, and multi-valued dataset property.

\vspace{+0.01in}
\noindent {\bf Generalizing our framework against security attacks:} In our current framework, each privacy protection task is formalized via an MI objective. An important future work would be generalizing our framework to design customized MI objectives to learn robust representations against security attacks such as evasion, poisoning, and backdoor attacks.

\section{Related Work}
\label{sec:related}

\subsection{MIAs and Defenses}
\label{sec:MIAsD}

\noindent {\bf MIAs~\cite{shokri2017membership,yeom2018privacy,salem2018ml,song2019privacy,chen2020gan,song2021systematic,leino2020stolen,choquette2021label,carlini2022membership,yuan2022membership,hui2021practical,wen2023canary}.}
Existing MIAs can be classified as \textit{training based}~\cite{shokri2017membership,yeom2018privacy,salem2018ml,song2019privacy,sablayrolles2019white,chen2020gan,leino2020stolen,choquette2021label,ye2022enhanced,carlini2022membership}   and \textit{non-training based}~\cite{song2021systematic,choquette2021label}. 
Given a (non)training sample and its output by a target ML model, training based MIAs use the (sample, output) pair to train a binary classifier, which is then used to determine whether a testing sample belongs to the training set or not. 
For instance, \cite{shokri2017membership} introduces multiple shadow models to perform training. 
In contrast, non-training based MIAs directly  
use the samples' predicted  score/label to make decisions. 
For instance, 
\cite{song2021systematic} designs a metric  \emph{prediction correctness}, which infers the membership based on whether a given  sample is correctly classified by the target model or not. 
Overall, an MIA that has more information is often more effective than that has less information. 

\vspace{+0.01in}
\noindent {\bf 
Defenses~\cite{srivastava2014dropout,shokri2017membership,salem2018ml,nasr2018machine,jia2019memguard,song2021systematic,shejwalkar2021membership,xu2022neuguard}.} 
They can be categorized as 
\textit{training time} based defense (e.g.,  dropout ~\cite{salem2018ml}, $L_2$ norm regularization~\cite{shokri2017membership}, model stacking~\cite{salem2018ml}, adversary regularization~\cite{nasr2018machine}, loss variance deduction ~\cite{xu2022neuguard}, DP~\cite{abadi2016deep,yu2019differentially,iyengar2019towards}, early stopping~\cite{song2021systematic}, knowledge distillation~\cite{shejwalkar2021membership}) and \textit{inference time} based defense (e.g., MemGuard~\cite{jia2019memguard}). 
 Almost all of them are empirical and broken by stronger  attacks~\cite{choquette2021label,song2021systematic}. DP is only defense offering privacy guarantees. 
Its main idea is to add noise to the gradient~\cite{abadi2016deep,yu2019differentially}
or objective function~\cite{iyengar2019towards} during training. The main drawback of current DP methods is that they have significant utility losses~\cite{jayaraman2019evaluating,shejwalkar2021membership}. 

\subsection{PIAs and Defenses}
\label{sec:PIAsD}

\noindent {\bf PIAs~\cite{ateniese2015hacking,ganju2018property,gopinath2019property,zhang2021leakage,maini2021dataset,wang2022group,mahloujifar2022property,zhou2022property,suri2022formalizing,chaudhari2023snap,taskagaaa24}.}
Ateniese et al.~\cite{ateniese2015hacking} are the first to describe the problem of the PIA (against support vector machines and hidden Markov models), where the attack is performed in the the white-box setting and consists of training a meta-classifier on top of many shadow models.  
Ganju et al.~\cite{ganju2018property} extend PIAs to neural networks, particularly fully connected neural networks (FCNNs). 
Zhang et al.~\cite{zhang2021leakage} propose PIAs 
 in the black-box setting and train a meta-classifier based on shadow models. 
Mahloujifar et al.~\cite{mahloujifar2022property} observe that data poisoning attacks can be incorporated into training the shadow model and increase the  effectiveness of PIAs.
Suri and Evans~\cite{suri2022formalizing} are the first to formally formalize PIAs as a cryptographic game, inspired by the way to formalize MIAs~\cite{yeom2018privacy}. 
They also extend the white-box attack on FCNNs~\cite{ganju2018property} to convolutional neural networks  (CNNs). 
Zhou et al.~\cite{zhou2022property} develop the first PIA against generative models, i.e., generative adversarial networks (GANs)~\cite{goodfellow2014generative}, under the black-box setting. 
Chaudhari et al.~\cite{chaudhari2023snap} propose a data poisoning strategy to perform the efficient private property inference.

\vspace{+0.01in}
\noindent {\bf Defenses}.  
To our best knowledge, there exist no known effective defenses against PIAs. DP cannot mitigate PIAs since it obfuscates individual samples, while PIAs care about the entire datasets~\cite{suri2022formalizing}. 
\cite{suri2022formalizing} also shows that DP does not work as a potential defense (also verified in Section~\ref{sec:eval}).

\subsection{DRAs and Defenses}
\label{sec:DRAsD}

\noindent {\bf DRAs~\cite{hitaj2017deep,he2019model,zhu2019deep,zhao2020idlg,wang2019beyond,geiping2020inverting,yin2021see,jeon2021gradient,balunovic2022bayesian,fowl2022robbing,balle2022reconstructing,balunovic2022lamp,wang2023reconstructing}.} 
Existing DRAs mainly reconstruct the training data from the model parameters or representations. 
They are formulated as an optimization problem that minimizes 
the difference between gradient from the raw training data and that from the reconstructed data. 
For instance, Zhu et al.~\cite{zhu2019deep} proposed a DLG attack method which relies entirely on minimization of the difference of gradients. 
Furthermore, several methods~\cite{hitaj2017deep,wang2019beyond,jeon2021gradient,geiping2020inverting,yin2021see} propose to incorporate prior knowledge (e.g., total variation regularization
~\cite{geiping2020inverting,yin2021see}, batch normalization statistics~\cite{yin2021see}) into the 
training data, or introduce an auxiliary dataset to simulate the training data distribution~\cite{hitaj2017deep, wang2019beyond,jeon2021gradient} 
(e.g., via GANs~\cite{goodfellow2014generative}). 
A few works~\cite{geiping2020inverting,zhu2021r} 
derive close-formed solutions to reconstruct the data, 
by constraining the neural networks to be fully connected~\cite{geiping2020inverting} or convolutional~\cite{zhu2021r}.

\vspace{+0.01in}%
\noindent {\bf Defenses~\cite{pathak2010multiparty,hamm2016learning,wei2020federated,zhu2019deep,sun2020provable,gao2021privacy,lee2021digestive,scheliga2022precode}.} 
Most of these defenses have none/little privacy guarantees. For instance, Zhu et al.~\cite{zhu2019deep} propose to prune model parameters with smaller magnitudes. 
Sun et al.~\cite{sun2020provable} propose to obfuscate the gradient for a single layer (called defender layer) such that the reconstructed data and the original data are dissimilar. Gao et al.~\cite{gao2021privacy} propose to generate augmented images that, when they are used to train the network, produce non-invertible gradients. 
These defenses are broken by an advanced attack
based on Bayesian learning~\cite{balunovic2022bayesian}. 
Only defenses based on DP-SGD~\cite{abadi2016deep}, a version of SGD with clipping and adding Gaussian noise, provide formal privacy guarantees.

\section{Conclusion}
\label{sec:conclusion}

We propose a unified information-theoretic framework, dubbed \texttt{\name}, to learn privacy-preserving representations against the three major types of inferences attacks (i.e., membership inference, property inference, and data reconstruction attacks). 
The framework formalizes the utility preservation and privacy protection against each attack via customized mutual information objectives. 
The framework also enables deriving theoretical results, e.g., inherent utility-privacy tradeoff, and guaranteed privacy leakage against each attack. 
    Extensive evaluations verify the effectiveness of \texttt{\name} for learning privacy-preserving representations and show the superiority over the compared baselines.

\section*{Acknowledgement} We thank all the anonymous reviewers and our shepherd for the valuable feedback and constructive comments. Wang is partially supported by the National Science Foundation (NSF) under grant Nos. ECCS-2216926, CNS-2241713, and CNS-2339686. Hong is partially
supported by the National Science Foundation (NSF) under grant Nos. CNS-2302689, CNS-2308730, CNS-2319277 and CMMI-2326341. Any opinions, findings and conclusions or recommendations expressed in this material are those of the author(s) and do not necessarily reflect the views of the funding agencies.

{
\small
\bibliographystyle{plain}
\bibliography{ref}
}

\appendix
\begin{algorithm}[!t]
\footnotesize
\caption{\texttt{\name} against MIAs}
\begin{flushleft}
{\bf Input:} Dataset $D_{1}$ of members and dataset $D_{0}$ of non-members, tradeoff hyperparameter $\lambda \in [0,1]$, learning rates $lr_1, lr_2, lr_3$;
\#local gradients $I$,
\#global rounds $T$. \\
{\bf Output:} Network parameters: $\Theta,\Psi,\Omega$.
\end{flushleft}
\vspace{-4mm}
\begin{algorithmic}[1]
\STATE Initialize $\Theta,\Psi,\Omega$ for the encoder $f$, 
 membership protection network $g_\Psi$, and utility preservation network $h_\phi$; 
\FOR{$t=1$ to $T$}
    \STATE $L_1 = \sum_{({\bf x}_j,u_j) \in D_1 \cup D_0 } H(u_j,g_{\Psi}(f({\bf x} _j)))$; 
    \STATE $L_2 = \sum_{({\bf x}_j, y_j) \in D_1} H(y_j,h_{\Omega}(f({\bf x} _j)))$;
    
    \FOR{$i=1$ to $I$}
        \STATE $\Psi\leftarrow \Psi - lr_1 \cdot \frac{\partial L_1}{\partial\Psi}$; 
        \STATE 
        $\Omega\leftarrow \Omega- lr_2 \cdot \frac{\partial L_2}{\partial \Omega}$;
        \STATE 
        $\Theta\leftarrow \Theta + lr_3 \cdot \frac{\partial (\lambda L_1 - (1-\lambda) L_2)}{\partial \Theta} $;
    \ENDFOR
\ENDFOR 
\end{algorithmic}
\label{alg:MIGuard_MIA}
\end{algorithm}

\begin{algorithm}[!t]
\footnotesize
\caption{\texttt{\name} against PIAs}
\begin{flushleft}
{\bf Input:} 
$N$ datasets $\{D_j\}_{j=1}^N$ sampled from a reference dataset $D_r$ with each $D_j$ having a property value $u_j$, 
tradeoff hyperparameter $\lambda\in [0,1]$, learning rates $lr_1, lr_2, lr_3$;
\#local gradients $I$,
\#global rounds $T$. \\
{\bf Output:} Network parameters: $\Theta,\Omega, \Psi$.
\end{flushleft}
\vspace{-4mm}
\begin{algorithmic}[1]
\STATE Initialize $\Theta,\Psi,\Omega$ for the encoder $f$, 
 property protection network $g_\Psi$, and utility preservation network $h_\phi$; 

\FOR{round $t=1$ to $T$}
    \STATE $L_1 =   \sum_{\{({\bf X}_j, {\bf y}_j) = D_j\}_j} H(u_j,g_{\Psi}(f({\bf X}_j)))$;

    \STATE $L_2 = \sum_{({\bf x}_i, y_i) \in \bigcup_j D_j} H({y}_i,h_{\Omega}(f({\bf x}_i)))$;

    \FOR{$i=1$ to $I$}
        \STATE $\Psi \leftarrow \Psi - lr_1 \cdot \frac{\partial L_1}{\partial\Psi}$;
        \STATE 
        $\Omega \leftarrow \Omega- lr_2 \cdot \frac{\partial L_2}{\partial \Omega}$;
        \STATE 
        ${\Theta\leftarrow \Theta + lr_3 \cdot \frac{\partial (\lambda L_1 - (1-\lambda) L_2}{\partial \Theta} }$.
    \ENDFOR
\ENDFOR  
\end{algorithmic}
\label{alg:MIGuard_PIA}
\end{algorithm}

\begin{algorithm}[!t]
 \footnotesize
\caption{Update perturbation distribution parameter $\Phi$}
\begin{flushleft}
{\bf Input:} $K$ Monte Carlo samples, the encoder $f_\Theta$ in the previous round, objective function Eqn (\ref{eqn:task_known_DRA_Guard}). 
learning rate $lr$, 
\#epochs $I_l$ \\
{\bf Output:} Perturbation distribution parameters $\Phi$ 
\end{flushleft}
\vspace{-4mm}
\begin{algorithmic}[1]
\STATE Initialize $\Phi = (\bm{\mu},\bm{\sigma})$.

\FOR{$i=1$ to $I_l$}
\FOR{$j=1$ to $K$}
       \STATE Sample ${\bf z}_j$ from $\mathcal{N}(0,1)$ 
        and compute $\bm{\delta}_j = \bm{\mu}+\bm{\sigma z}_j$; 
        \STATE Calculate the gradient ${\bf g}_\Phi$ of Eqn (\ref{eqn:genet}) w.r.t. $\Phi$;
        \STATE Update $\Phi$ by: $\Phi \leftarrow \Phi - lr \cdot {\bf g}_\Phi$.
    \ENDFOR
\ENDFOR 
\end{algorithmic}
\label{alg:update_perdist}
\end{algorithm}

\begin{algorithm}[!t]
 \footnotesize
\caption{\texttt{\name} against DRAs}
\begin{flushleft}
{\bf Input:} A dataset $D = \{{\bf x}_n, y_n \}$, 
hyperparameters $\lambda \in [0,1]$, 
learning rates $lr_1, lr_2, lr_3$, 
\#local gradients $I$,
\#global rounds $T$. \\
{\bf Output:} Network parameters: $\Omega, \Psi, \Theta$. 
\end{flushleft}
\vspace{-4mm}
\begin{algorithmic}[1]
\STATE Initialize $\Theta,\Psi,\Omega, \Phi$ for the encoder $f$, 
 data reconstruction network $g_\Psi$,  utility preservation network $h_\Omega$, and perturbation distribution parameter.

\FOR{round $t=1$ to $T$}
\FOR{each batch $bs \subset D$}
\STATE {\bf Update $\Phi$} via Algorithm~\ref{alg:update_perdist}; 

    \STATE {\bf Update $g_\Psi$} (given $\Theta$ and $\{\bm{\delta}_i\}$): Calculate $I^{(JSD)}_{\Theta, \Psi}$ on $bs$ with $\{\bm{\delta}_i\}$ via Eqn (\ref{eqn:MI_JSD}); $\Psi \leftarrow \Psi + lr_1 \cdot {\partial I^{(JSD)}_{\Theta, \Psi}}/{\partial\Psi}$; 
    
    \STATE {\bf Update $h_\Omega$} (given $\Theta$ and $\{\bm{\delta}_i\}$): Calculate CE loss $L_{1}$ on $bs$ with $\{\bm{\delta}_i\}$ via Eqn (\ref{eqn:CE_pert});
    Calculate CE loss $L_{2}$ on $bs$ with clean data via Eqn (\ref{eqn:CE_clean}); 
    $\Omega\leftarrow \Omega- lr_2 \cdot {\partial (L_1 +  L_2)}/{\partial \Omega}$;
    
    \STATE {\bf Update $f_\Theta$} (given $\Psi$, $\Omega$, and $\{\bm{\delta}_i\}$): 
    $\Theta \leftarrow \Theta - lr_3 \cdot \frac{\partial}{\partial \Theta} (\lambda I^{(JSD)}_{\Theta, \Psi} + (1-\lambda) ( L_{1} +  L_{2}))$; 

\ENDFOR
\ENDFOR 
\end{algorithmic}
\label{alg:MIGuard_DRA}
\end{algorithm}

\section{Proofs of Privacy Guarantees}
\label{app:PrivGuarantees}

The following lemmas will be useful in the proofs.
\begin{lemma}[\cite{calabro2009exponential} Theorem 2.2]
\vspace{-2mm}
\label{lem:inveren}
Let $H_2^{-1}(p)$ be the inverse binary entropy function for $p \in [0,1]$, then $H_2^{-1}(p) \geq \frac{p}{2 \log_2({6}/{p})}$.
\vspace{-2mm}
\end{lemma}

\begin{lemma}[Data processing inequality] 
\label{lem:datainq}
Given random variables $X$, $Y$, and $Z$ that form a Markov chain in the
order $X \rightarrow Y \rightarrow Z$, then the mutual information between $X$ and $Y$ is greater than or equal to the mutual information between $X$ and $Z$. That is $I(X;Y) \geq I(X;Z)$.
\vspace{-2mm}
\end{lemma}

\subsection{Proof of Theorem \ref{thm:provprivacy_MI} for MIAs}
\label{supp:provprivacy_MI}

\provprivacymi*

\begin{proof} 
For brevity, we define the optimal MIA as $A^*$.
Let $s$ be an indicator that takes value 1 if and only if 
$A^*(\mathbf{r}) \neq u$, and 0 otherwise, i.e., $s = 1[A^*(\mathbf{r}) \neq u]$. Now consider the conditional entropy $H(s, u|A^*(\mathbf{r}))$.
By decomposing it via two different ways, we have 
\begin{align}
\label{eqn:entdecompo}
H(s, u|A^*(\mathbf{r})) & = H(u|A^*(\mathbf{r})) + H(s | u,  A^*(\mathbf{r})) \notag \\ 
& =  H(s | A^*(\mathbf{r})) + H(u| s, A^*(\mathbf{r})),
\end{align} 

Note that $H(s | u, A^*(\mathbf{r}))=0$ as when $u$ and $A^*(\mathbf{r})$ are known, $s$ is also known.
Moreover, 
\begin{align}
& H(u| s, A^*(\mathbf{r})) \notag \\
& = Pr(s=1) H(u| s=1, A^*(\mathbf{r})) 
+ Pr(s=0) H(u| s=0, A^*(\mathbf{r})) \notag \\ 
& = 0+0 = 0,
\end{align} 
because when knowing $s$ and $A^*(\mathbf{r})$, we also 
know $u$.

Thus, Equation~\ref{eqn:entdecompo} reduces to $H(u|A^*(\mathbf{r}))=  H(s | A^*(\mathbf{r}))$. 
As conditioning does not increase entropy, i.e.,  
$H(s | A^*(\mathbf{r})) \leq H(s)$, we further have
\begin{align}
\label{eqn:finalone}
H(u|A^*(\mathbf{r})) \leq H(s).
\end{align}

On the other hand, using mutual information and entropy properties, we have
$I(u; A^*(\mathbf{r})) = H(u) - H(u|A^*(\mathbf{r}))$ and
$I(u; \mathbf{r}) = H(u) - H(u|\mathbf{r})$. Hence, 
\begin{align}
\label{eqn:mi_en}
    I(u; A^*(\mathbf{r})) + H(u|A^*(\mathbf{r})) = I(u; \mathbf{r}) + H(u|\mathbf{r}).
\end{align}

Note that 
$u \perp A^*(\mathbf{r}) | {\bf r}$. Hence, we have the Markov chain $u \rightarrow {\bf r} \rightarrow A^*(\mathbf{r})$.
Based on the data processing inequality in Lemma~\ref{lem:datainq}, $I(u; A^*(\mathbf{r})) \leq I(u; \mathbf{r})$. 
Combining it with Equation~\ref{eqn:mi_en}, we have 
\begin{align}
\label{eqn:finaltwo}
H(u|A^*(\mathbf{r})) \geq H(u|\mathbf{r}).
\end{align}

Combing Equations~\ref{eqn:finalone} and~\ref{eqn:finaltwo}, we have $H(s) = H_2(Pr(s=1) )\geq H(u|\mathbf{r})$,
which implies 
\begin{align}
    \label{eqn:final}
    & Pr(A^*(\mathbf{r}) \neq u) = Pr(s=1) \geq H_2^{-1} (H(u|\mathbf{r})),
\end{align} 
where $H_2 (t) = -t \log_2 t - (1-t) \log_2 (1-t)$. 

Finally, by applying Lemma~\ref{lem:inveren}, we have 
\begin{align}
Pr(A^*(\mathbf{r}) \neq u) \geq \frac{H(u|{\bf r})}{2 \log_2 ({6}/{H(u|{\bf r})})}.
\end{align} 
Hence the membership privacy leakage is bounded by $Pr(A^*(\mathbf{r}) = u) \leq 1 - \frac{H(u|{\bf r})}{2 \log_2 ({6}/{H(u|{\bf r})})}.$
\end{proof}

\subsection{Proof of Theorem \ref{thm:provprivacy_PI} for PIAs}
\label{supp:provprivacy_PI}

\provprivacypi*

\begin{proof}
The proof is identical to that for Theorem~\ref{thm:provprivacy_MI}. 
The only differences are: 1) replace ${\bf r}$ to be ${\bf R}$; and 2) $s$ is  an indicator that takes value 1 if
and only if $A({\bf R}) \neq u$, and 0 otherwise, i.e., $s = 1[A({\bf R}) \neq u]$, where $u$ is the private dataset property.
\end{proof}

\subsection{Proof of Theorem \ref{thm:provprivacy_DR_appr} for DRAs}
\label{supp:provprivacy_DR}

Different from MIAs and PIAs where the adversary makes decisions in a discrete space, i.e., inferring member or non-member and property value, DRAs aim to infer the \emph{continuous} data. 
To deal with this challenging scenario, we need to first introduce the following lemma.

Recall that for a set $\mathcal{S}$ in a $d$-dimension space, we denote $\partial \mathcal{S}$ as its boundary in the $(d-1)$-dimensional space, and denote the volume of $\partial \mathcal{S}$ as $\textrm{Vol}(\partial \mathcal{S})$. For a ${\bf v} \in \mathcal{V}$, an $l_p$-norm ball centered at ${\bf v}$ with a radius $t$ as $\mathcal{B}_p({\bf v},t) = \{ {\bf v}' \in \mathbb{R}^d: \|{\bf v}' -  {\bf v}\|_p \le t \}$. 

\begin{lemma}[Proposition 2 in \cite{duchi2013distance}]
\label{lem:dwlowerbound}
Assume the set $\mathcal{V} \subset \mathbb{R}^d$ has a non-zero and finite volume. 
Then, if ${\bf v}$ is uniform over $\mathcal{V}$ and for any Markov chain 
${\bf v} \rightarrow \cdots \rightarrow  {\bf w} \rightarrow \cdots \rightarrow {\bf v}'$, we have $\textrm{Pr}(\|{\bf v}' - {\bf v} \|_p \geq t) \geq 1 - \frac{I({\bf v}; {\bf w}) + \log 2}{\log \text{Vol}(\partial \mathcal{V}) - \log \text{Vol}(\partial \mathcal{V}(t))}$, where $\text{Vol}(\partial \mathcal{V}(t)) = {\max_{{\bf v} \in \mathcal{V}} \textrm{Vol}(\partial \mathcal{B}_p({\bf v},t)\cap \mathcal{V})}$. 
\end{lemma}

Now we prove the Theorem~\ref{thm:provprivacy_DR_appr} restated as below. 

\provprivacydrappr*

\begin{proof}
We have the following Markov chain: $
{\bf x} \rightarrow  {\bf r} = f({\bf x}) \rightarrow {\bf r} + \bm{\delta} \rightarrow A_{DRA}({\bf r}+ \bm{\delta})$. 
W.l.o.g., we assume ${\bf x}$ is uniform over the input set $\mathcal{X}$. We let ${\bf x} = {\bf v}$, ${\bf r} + \bm{\delta} = {\bf w}$, and $A_{DRA}({\bf r}+\bm{\delta}) = {\bf v}' $. Then by applying Lemma~\ref{lem:dwlowerbound}, we reach out Theorem~\ref{thm:provprivacy_DR_appr}. 
\end{proof}

\section{Theoretical Utility-Privacy Tradeoff}
\label{app:UPTradeoff}

\subsection{Tradeoff of \texttt{\name} against MIAs}
\label{sec:uptradeoff_mi}

Let $\mathcal{A}_{MIA}$ be the set of all MIAs, i.e., $\mathcal{A}_{MIA}=\{A_{MIA}: {\bf r} = f({\bf x}) \in \mathcal{Z} \rightarrow u \in \{0,1\}\}$, with data ${\bf x}$ randomly sampled from the data distribution $\mathcal{D}$.
W.l.o.g, we assume the representation space $\mathcal{Z}$ is bounded by $R$, i.e., $\max_{{\bf r} \in \mathcal{Z}} \|{\bf r} \| \leq R$.
Remember the learning task classifier $C: \mathcal{Z} \rightarrow \mathcal{Y} = \{0,1\}$ is on top of the representations ${\bf r}$. 
We further define the \emph{advantage} of any 
MIA with respect to the data distribution $\mathcal{D}$ 
as:
\begin{align}
    \label{eqn:MIadv}
    & \textrm{Adv}_\mathcal{D} (\mathcal{A}_{MIA}) = \max_{A_{MIA} \in \mathcal{A}_{MIA}} | \textrm{Pr}(A_{MIA} ({\bf r}) =a | u=a) \notag \\ 
    & \qquad - \textrm{Pr} (A_{MIA} ({\bf r}) =a | u=1-a) |, \, \forall a \in \{0,1\}, 
\end{align}
where $\textrm{Adv}_\mathcal{D}(\mathcal{A}_{MIA}) =1$ means the strongest MIA can \emph{completely} infer the privacy membership  through the learnt representation. In contrast, an MIA obtains a \emph{random guessing} 
MIA performance if $\textrm{Adv}_\mathcal{D}(\mathcal{A}_{MIA}) =0$.  

\begin{restatable}[]{theorem}{uptradeoffmi}
\label{thm:uptradeoff_mi}
\vspace{-2mm}
Let ${\bf r} = f({\bf x})$ be the representation with a bounded norm $R$ 
outputted by the encoder $f$ by Equation (\ref{eqn:task_known_MI_Guard}) on ${\bf x}$ with label $y$, $u$ be ${\bf x}$'s membership.  
Assume the task classifier $C$ is $C_L$-Lipschitz. 
Then the utility 
loss/risk  
induced by all MIAs $\mathcal{A}_{MIA}$ is bounded as below:  
\begin{align}
\label{eqn:uptradeoff_mi}
    \mathrm{Risk}_{MIA}(C \circ f)
    \geq \Delta_{y|u} - 2R \cdot C_L \cdot \textrm{Adv}_\mathcal{D}(\mathcal{A}_{MIA}), 
\end{align}
where 
$\Delta_{y|u} = |\textrm{Pr}(y=1|u=0) - \textrm{Pr}(y=1|u=1)|$ is a (task-dependent) constant. 
\vspace{-2mm}
\end{restatable}
\noindent \emph{Remark.} 
Theorem~\ref{thm:uptradeoff_mi} states any 
 learning task classifier using  representations learnt by 
the encoder incurs a risk, at the cost of membership protection---The larger the advantage $\textrm{Adv}_\mathcal{D}(\mathcal{A}_{MIA})$, the smaller the lower bound risk, and vice versa. Note that the lower bound is {independent} of the adversary.
Hence, Theorem~\ref{thm:uptradeoff_mi} 
reflects an inherent trade-off between  utility preservation and membership protection.

\vspace{-2mm}
\subsection{Tradeoff of \texttt{\name} against PIAs}
\label{sec:uptradeoff_pi}

Let $\mathcal{A}_{PIA}$ be the set of all PIAs that have access to the representations ${\bf R}$ of a dataset ${\bf X} = \{ {\bf x}_i \} $ sampled from the data distribution $\mathcal{D}$, i.e., $\mathcal{A}_{PIA}=\{A_{PIA}: {\bf R} = f({\bf X})\in \mathcal{Z}
\rightarrow u=\{0,1\}\}$. 
Let the dataset representation space is bounded, i.e., $\max_{\mathbf{R} \in \mathcal{Z}} \|{\bf R} \| \leq R$.
We further define the {advantage} of any 
PIA with respect to the data distribution $\mathcal{D}$ as:
\begin{align}
    \label{eqn:PIadv}
    & \textrm{Adv}_\mathcal{D} (\mathcal{A}_{PIA}) = \max_{A_{PIA} \in \mathcal{A}_{PIA}} | \textrm{Pr}(A_{PIA}({\bf R}) =a | u=a) \notag \\ 
    & \quad \, - \textrm{Pr} (A_{PIA}({\bf R}) =a | u=1-a) |, \, \forall a \in \{0,1\}, 
\end{align}
where $\textrm{Adv}_\mathcal{D}(\mathcal{A}_{PIA}) =1$ means the strongest PIA can \emph{completely} infer the privacy dataset property through the learnt dataset representations, while $\textrm{Adv}_\mathcal{D}(\mathcal{A}_{PIA}) =0$ implies 
a \emph{random guessing} PI performance.  

\begin{restatable}[]{theorem}{uptradeoffpi}
\label{thm:uptradeoff_pi}
Let $({\bf X}, {\bf y})$ be a dataset randomly sampled from the data  distribution $\mathcal{D}$ and ${\bf R} = f({\bf X})$ be the dataset representation outputted by the encoder $f$ by Equation (\ref{eqn:task_known_PIA_Guard}) on ${\bf X}$. Assume the representation space is bounded by $R$ and  task classifier $C$ is $C_L$-Lipschitz. Let 
$u$ be ${\bf X}$'s private property.
Then, the risk induced by all PIAs $\mathcal{A}_{PIA}$  is  bounded as below:  
\begin{align}
\label{eqn:uptradeoff_pi}
    \mathrm{Risk}_{PIA}(C \circ f) 
    \geq \Delta_{{y} | u} - 2R \cdot C_L \cdot \textrm{Adv}_\mathcal{D}(\mathcal{A}_{PIA}). 
\end{align}
where $\Delta_{{\bf y} | u}$ is same as in Equation~\ref{eqn:uptradeoff_mi}.
\vspace{-2mm}
\end{restatable}

\noindent \emph{Remark.} 
Similarly, Theorem~\ref{thm:uptradeoff_pi} states any learning task classifier using representations learnt by the encoder incurs a utility loss. The larger/smaller the advantage $\textrm{Adv}_\mathcal{D}(\mathcal{A}_{PIA})$, the smaller/larger the lower bound risk. Also, the lower bound is {independent} of the adversary, and thus covers the strongest PIA. Hence, Theorem~\ref{thm:uptradeoff_pi} shows an inherent tradeoff between  utility preservation and dataset property protection.

\vspace{-2mm}
\subsection{Tradeoff of \texttt{\name} against DRAs}
\label{sec:uptradeoff_dr}
Let $\mathcal{A}_{DRA}$ be the set of all DRAs that have access to the perturbed data representations, i.e., $\mathcal{A}_{DRA}=\{A_{DRA}: {\bf r} + \bm{\delta} \in \mathcal{Z} \times \mathcal{P} \rightarrow {\bf x} \in \mathcal{X}\}$. 
We assume the perturbed representation space $\mathcal{Z}$ is bounded, i.e., $\max_{{\bf r} + \bm{\delta} \in \mathcal{Z} \times \mathcal{P}} \|{\bf r} + \bm{\delta} \| \leq R'$. 
We also denote the perturbed representation as ${\bf r}' = {\bf r} + \bm{\delta}$ for short. 
We further define the {advantage}\footnote{Note that our defined advantage is different from that in \cite{salem2023sok}.} of any DRA 
with respect to the data distribution $\mathcal{D}$ as: 
\begin{align}
    \label{eqn:DRadv}
    & \textrm{Adv}_\mathcal{D} (\mathcal{A}_{DRA}) = \max_{A_{DRA} \in \mathcal{A}_{DRA}} | \textrm{Pr}(A_{DRA}({\bf r}') = {\bf x} | y=0) \notag \\ 
    & \qquad - \textrm{Pr} (A_{DRA}({\bf r}') = {\bf x} | y=1) |, 
\end{align}
where $\textrm{Adv}_\mathcal{D}(\mathcal{A}_{DRA}) =1$ means the strongest DRA can \emph{completely} reconstruct the private data through the perturbed representation. In contrast, $\textrm{Adv}_\mathcal{D}(\mathcal{A}_{DRA}) =0$ means the adversary cannot infer any raw data information.

\begin{restatable}[]{theorem}{uptradeoffdr}
\label{thm:uptradeoff_dr}
\vspace{-2mm}
Let $f$ and ${\bf r}'$ be the learnt encoder and perturbed representation with a bounded norm $R'$ outputted by Equation (\ref{eqn:task_known_DRA_Guard}) on ${\bf x}$ with label $y$, respectively. 
Assume the task classifier $C$ is $C_L$-Lipschitz. 
Then the utility loss 
induced by all DR adversaries $\mathcal{A}_{DRA}$ is bounded as below:  
\begin{align}
\label{eqn:uptradeoff_dr}
    \mathrm{Risk}_{DRA}(C \circ f)
    & \geq \Delta_{y} - 2R' \cdot C_L \cdot \textrm{Adv}_\mathcal{D}(\mathcal{A}_{DRA}),
\end{align}
where $\Delta_{y} = |\textrm{Pr}(y=1) - \textrm{Pr}(y=0)|$ is a 
constant. 
\vspace{-2mm}
\end{restatable}
\noindent \emph{Remark.} 
Theorem~\ref{thm:uptradeoff_dr} states that, given a  task-dependent constant $\Delta_{y}$, any learning task classifier using  representations learnt by 
the encoder $f$ incurs a risk. 
The larger the advantage $\textrm{Adv}_\mathcal{D}(\mathcal{A}_{DRA})$, the smaller the lower bound risk, and vice versa. Note that the lower bound covers the strongest  DRA. Hence, Theorem~\ref{thm:uptradeoff_dr} reflects an inherent trade-off between  utility preservation and data sample protection.

\subsection{Proof of Theorem \ref{thm:uptradeoff_mi} for MIAs}
\label{supp:uptradeoff_mi}

We first introduce the following definitions and lemmas.

\begin{definition}[Lipschitz function and Lipschitz norm]
A function $f: A \rightarrow \mathbb{R}^m$ is $L$-Lipschitz continuous, if for any $a, b \in A$, $\|f(a)-f(b)\| \leq L \cdot \|a-b\|$. Lipschitz norm of $f$, i.e., $\|f\|_L$, is defined as $\|f\|_L = \max \frac{\|f(a)-f(b)\|_L}{\|a-b\|_L}$.
\end{definition}

\begin{definition}[Total variance (TV) distance]
\label{def:TV}
Let $\mathcal{D}_1$ and $\mathcal{D}_2$ be two distributions over the same sample space $\Gamma$,  the TV distance between $\mathcal{D}_1$ and $\mathcal{D}_2$ is defined as: $d_{TV}(\mathcal{D}_1, \mathcal{D}_2) =  \max_{E \subseteq \Gamma} |\mathcal{D}_1 (E) - \mathcal{D}_2(E)|$.   
\end{definition} 

\begin{definition}[1-Wasserstein distance]
\label{def:wassdis}
Let $\mathcal{D}_1$ and $\mathcal{D}_2$ be two distributions over the same sample space $\Gamma$, the 1-Wasserstein distance between $\mathcal{D}_1$ and $\mathcal{D}_2$ is defined as $W_1(\mathcal{D}_1, \mathcal{D}_2) = \max_{\|f\|_L \leq 1} |\int_{\Gamma} f d\mathcal{D}_1 - \int_{\Gamma} f d\mathcal{D}_2 |$, where $\|\cdot\|_L$  is the Lipschitz norm of a real-valued function. 
\end{definition} 

\begin{definition}[Pushforward distribution]
\label{def:pushforDist}
Let $\mathcal{D}$ be a distribution over a sample space and $g$ be a function of the same space. Then we call $g(\mathcal{D})$ the induced pushforward distribution 
of $\mathcal{D}$.
\end{definition}

\begin{lemma}[Contraction of the 1-Wasserstein distance]
\label{lem:contWass}
Let $g$ be a function defined on a space and $C_L$ be the constant such that $\|g\|_L \leq C_L$. 
Then for any two distributions $\mathcal{D}_1$ and $\mathcal{D}_2$ over this space, 
$ W_1(g(\mathcal{D}_1), g(\mathcal{D}_2))  \leq C_L \cdot W_1(\mathcal{D}_1, \mathcal{D}_2)$.
\end{lemma}

\begin{lemma}[1-Wasserstein distance over two Bernoulli random scalars]
\label{lem:berprob}
Let $y_1$ and $y_2$ be two Bernoulli random scalars with distributions $\mathcal{D}_1$ and $\mathcal{D}_2$, respectively. Then, 
$W_1(\mathcal{D}_1, \mathcal{D}_2)  = |\textrm{Pr}(y_1=1) - \textrm{Pr}(y_2=1)|$.
\end{lemma}

\begin{lemma}
[Relationship between the 1-Wasserstein distance and the TV distance~\cite{gibbs2002choosing}]
\label{lem:tvWass}
Let $g$ be a function defined on a norm-bounded space $\mathcal{Z}$, where $\max_{{\bf r} \in \mathcal{Z}} \|{\bf r} \| \leq R$, and $\mathcal{D}_1$ and $\mathcal{D}_1$ are two distributions over the space $\mathcal{Z}$.  Then $W_1(g(\mathcal{D}_1), g(\mathcal{D}_2)) \leq 2R \cdot d_{TV}(g(\mathcal{D}_1), g(\mathcal{D}_2))$.  
\end{lemma}

We now prove Theorem~\ref{thm:uptradeoff_mi}.

\begin{proof}
We denote $\mathcal{D}_{{\bf x}|u}$ as the conditional data distribution of $\mathcal{D}$ given $u$, i.e., ${\bf x}|u \sim \mathcal{D}_{{\bf x}|u}$,  
and $\mathcal{D}_{y|u}$ as the conditional label distribution 
 given $u$, i.e., $y|u \sim \mathcal{D}_{y|u}$.  
As $C$ is a binary task classifier on top of the encoder $f$, 
it follows that the pushforward 
$C \circ f(\mathcal{D}_{{\bf x} |u=0})$ and $C \circ f(\mathcal{D}_{{\bf x} |u=1})$ induce two distributions over $\{0,1\}$. 
We denote $Cf$ as $C\circ f$ for short. 
By leveraging the triangle inequalities of 1-Wasserstein distance, we have 
\begin{small}
\begin{align}
& W_1 (\mathcal{D}_{y|u=0}, \mathcal{D}_{y|u=1}) 
\leq 
W_1 (\mathcal{D}_{y|u=0}, Cf(\mathcal{D}_{{\bf x} |u=0})) \nonumber \\ 
& + 
W_1 (Cf(\mathcal{D}_{{\bf x} |u=0}), Cf(\mathcal{D}_{{\bf x} |u=1})) 
+  W_1 (Cf(\mathcal{D}_{{\bf x} |u=1}), \mathcal{D}_{y|u=1})
\label{eqn:keytri}
\end{align}
\end{small}
Using Lemma~\ref{lem:berprob} on Bernoulli r.v. 
$y|u=a$, we have 
\begin{small}
\begin{align}
\label{eqn:berprob}
& W_1 (\mathcal{D}_{y|u=0}, \mathcal{D}_{y|u=1}) = 
|\textrm{Pr}_{\mathcal{D}}(y=1|u=0) - \textrm{Pr}_{\mathcal{D}}(y=1|u=1) | \notag \\
& = \Delta_{y|u}.
\end{align}
\end{small}%

Using Lemma~\ref{lem:contWass} on the contraction of the 1-Wasserstein distance and that $\|C\|_L \leq C_L$, we have 
\begin{small}
\begin{align}
\label{eqn:contWass_MI}
& W_1 (Cf(\mathcal{D}_{{\bf x} |u=0}), Cf(\mathcal{D}_{{\bf x} |u=1})) 
\leq C_L \cdot W_1(f(\mathcal{D}_{{\bf x} |u=0}), f(\mathcal{D}_{{\bf x} |u=1})).
\end{align}
\end{small}
Using Lemma~\ref{lem:tvWass} with $\max_{{\bf r}} \|{\bf r}\| \leq R$, we have 
\begin{small}
\begin{align}
\label{eqn:tvWass_MI}
W_1(f(\mathcal{D}_{{\bf x} |u=0}), f(\mathcal{D}_{{\bf x} |u=1})) \leq 2R \cdot d_{TV}(f(\mathcal{D}_{{\bf x} |u=0}), f(\mathcal{D}_{{\bf x} |u=1})). 
\end{align}
\end{small}%
We next show $d_{TV}(f(\mathcal{D}_{{\bf x} |u=0}), f(\mathcal{D}_{{\bf x} |u=1})) = \textrm{Adv}_{\mathcal{D}}(\mathcal{A}_{MIA})$. 
{
\small 
\begin{align}
& d_{TV}(f(\mathcal{D}_{{\bf x} |u=0}), f(\mathcal{D}_{{\bf x} |u=1})) 
= \max_{E} |\textrm{Pr}_{f(\mathcal{D}_{{\bf x} |u=0})}(E) - \textrm{Pr}_{f(\mathcal{D}_{{\bf x} |u=1})}(E)| \nonumber \\
& = \max_{A_E \in \mathcal{A}_{MIA}} | \textrm{Pr}_{{\bf r} \sim f(\mathcal{D}_{{\bf x} |u=0})}(A_E({\bf r})=1) 
- \textrm{Pr}_{{\bf r} \sim f(\mathcal{D}_{{\bf x} |u=1})}(A_E({\bf r})=1) | \nonumber \\
& = \max_{A_E \in \mathcal{A}_{MIA}} | \textrm{Pr}(A_E({\bf r})=1 | u=0) - \textrm{Pr}(A_E({\bf r})=1 | u=1) | \nonumber \\
& = \textrm{Adv}_{\mathcal{D}}(\mathcal{A}_{MIA}), \label{eqn:tvadv_MI}
\end{align}
}%
where the first equation uses the definition of TV distance, and  $A_E(\cdot)$ is the characteristic function of the event $E$ in the second equation.
With Equations~\ref{eqn:contWass_MI}-\ref{eqn:tvadv_MI}, we have
\begin{small}
\begin{align}
W_1 (Cf(\mathcal{D}_{{\bf x} |u=0}), Cf(\mathcal{D}_{{\bf x} |u=1})) \leq 2R \cdot  C_L \cdot \textrm{Adv}_{\mathcal{D}}(\mathcal{A}_{MIA}). 
\end{align}
\end{small}%
Furthermore, using Lemma~\ref{lem:berprob} on Bernoulli random variables $y$ and $Cf({\bf x})$, we have 
\begin{small}
\begin{align}
& W_1 (\mathcal{D}_{y|u=a}, Cf(\mathcal{D}_{{\bf x}|u=a})) 
\notag \\
& = |\textrm{Pr}_{\mathcal{D}}(y=1 | u=a) - \textrm{Pr}_{\mathcal{D}}({Cf({\bf x}})=1 | u=a)) | \nonumber \\
& = |\mathbb{E}_{\mathcal{D}}[y|u=a] - \mathbb{E}_{\mathcal{D}}[Cf({\bf x})| u=a]| \nonumber \\
& \leq \mathbb{E}_{\mathcal{D}}[|y-Cf({\bf x})| |u=a] 
= \textrm{Pr}_{\mathcal{D}}(y \neq Cf({\bf x}) | u=a) 
\label{eqn:CEloss}
\end{align}
\end{small}%
Finally, by combining Equations~\ref{eqn:keytri}-\ref{eqn:CEloss}, we have:
\begin{small}
\begin{align}
    \Delta_{y|u} & \leq \textrm{Pr}_{\mathcal{D}}(y \neq Cf({\bf x}) | u=0)  + \textrm{Pr}_{\mathcal{D}}(y \neq Cf({\bf x}) | u=1) \notag \\ 
    & \quad + 2R \cdot C_L \cdot \textrm{Adv}_{\mathcal{D}}(\mathcal{A}_{MIA}).
\end{align}
\end{small}%
Hence, 
$
\textrm{Risk}_{MI}(C \circ f) = 
\textrm{Pr}_{\mathcal{D}}(y \neq Cf({\bf x}) | u=0)  + \textrm{Pr}_{\mathcal{D}}(y \neq Cf({\bf x}) | u=1) \geq \Delta_{y|u} - 2R \cdot C_L \cdot \textrm{Adv}_{\mathcal{D}}(\mathcal{A}_{MIA}).
$
\vspace{-2mm}
\end{proof}

\subsection{Proof of Theorem \ref{thm:uptradeoff_pi} for PIAs}
\label{supp:uptradeoff_pi}

\begin{proof}
We follow the 
similar strategy for proving Theorem~\ref{thm:uptradeoff_mi}. 
To differential 
the conditional \emph{data sample} distribution $\mathcal{D}_{{\bf x} | u}$ in MIAs, we denote $\mathcal{D}_{{\bf X} | u}$ as the conditional \emph{dataset} distribution of $\mathcal{D}$ given the dataset property $u$, and $\mathcal{D}_{{\bf y}|u}$ as the conditional distribution of dataset labels ${\bf y}$ given $u$. 
Also, the pushforward 
$Cf(\mathcal{D}_{{\bf X} | u=0})$ and $Cf(\mathcal{D}_{{\bf X} | u=1})$ induce two distributions over $\{0,1\}^{|{\bf y}|}$. 
First, by leveraging the triangle inequalities of the 1-Wasserstein distance, we have
\begin{small}
\begin{align}
& W_1 (\mathcal{D}_{{\bf y}|u=0}, \mathcal{D}_{{\bf y}|u=1}) 
\leq W_1 (\mathcal{D}_{{\bf y}|u=0}, Cf(\mathcal{D}_{{\bf X} | u=0})) \label{eqn:keytri_PI} \\ & 
\qquad + 
W_1 (Cf(\mathcal{D}_{{\bf X} | u=0}), Cf(\mathcal{D}_{{\bf X} | u=1})) 
+  W_1 (Cf(\mathcal{D}_{{\bf X} | u=1}), \mathcal{D}_{{\bf y}|u=1}) \nonumber
\end{align} 
\end{small}

Using Lemma~\ref{lem:berprob} on Bernoulli random variables ${y_i}|u=a$ and that ${y_i} \in {\bf y}$ are independent, we have 
\begin{small}
\begin{align}
\label{eqn:berprob_PI}
& W_1 (\mathcal{D}_{{\bf y}|u=0}, \mathcal{D}_{{\bf y}|u=1}) \notag 
= {\frac{1}{|{\bf y}|}\sum_{{y_i} \in {\bf y}} W_1 (\mathcal{D}_{{y_i}|u=0}, \mathcal{D}_{{y_i}|u=1})} \notag \\ 
& = {\frac{1}{|{\bf y}|} \sum_{{y_i} \in {\bf y}} |\textrm{Pr}_{\mathcal{D}}({y_i}=1|u=0) - \textrm{Pr}_{\mathcal{D}}({y_i}=1|u=1) |} \notag \\
& = {|\textrm{Pr}_{\mathcal{D}}({y}=1|u=0) - \textrm{Pr}_{\mathcal{D}}({y}=1|u=1) |} 
\notag \\
&  
= \Delta_{{y}|u},  
\end{align}
\end{small}%
where in the first equality and third equality we use the independence of $\{y_i\}$'s and second equality uses Equation \ref{eqn:berprob}.

Using Lemma~\ref{lem:contWass} on the contraction of the 1-Wasserstein distance and that $\|c\|_L \leq C_L$, we have 
\begin{small}
\begin{align}
\label{eqn:contWass_PI}
& W_1 (Cf(\mathcal{D}_{{\bf X} | u=0}), Cf(\mathcal{D}_{{\bf X} | u=1})) 
\leq C_L \cdot W_1(f(\mathcal{D}_{{\bf X} | u=0}), f(\mathcal{D}_{{\bf X} | u=1})).
\end{align}
\end{small}

Using Lemma~\ref{lem:tvWass} with $\max_{{\bf R}} \|{\bf R}\| \leq R$, we have 
\begin{small}
\begin{align*}
W_1(f(\mathcal{D}_{{\bf X} | u=0}), f(\mathcal{D}_{{\bf X} | u=1})) \leq 2R \cdot d_{TV}(f(\mathcal{D}_{{\bf X} | u=0}), f(\mathcal{D}_{{\bf X} | u=1})). 
\end{align*}
\end{small}%
We next show $d_{TV}(f(\mathcal{D}_{{\bf X} | u=0}), f(\mathcal{D}_{{\bf X} | u=1})) = \textrm{Adv}_{\mathcal{D}}(\mathcal{A}_{PIA})$. 
\begin{small}
\begin{align}
& d_{TV}(f(\mathcal{D}_{{\bf X} | u=0}), f(\mathcal{D}_{{\bf X} | u=1})) 
= 
\max_{E} |\textrm{Pr}_{f(\mathcal{D}_{{\bf X} | u=0})}(E) - \textrm{Pr}_{f(\mathcal{D}_{{\bf X} | u=1})}(E)| \nonumber \\
& = \max_{A_E \in \mathcal{A}_{PIA}} | \textrm{Pr}_{{\bf R} \sim f(\mathcal{D}_{{\bf X} | u=0})}(A_E({\bf R})=1) 
- \textrm{Pr}_{{\bf R} \sim f(\mathcal{D}_{{\bf X} | u=1})}(A_E({\bf R})=1) | \nonumber \\
& = \max_{A_E \in \mathcal{A}_{PIA}} | \textrm{Pr}(A_E({\bf R})=1 | u=0) - \textrm{Pr}(A_E({\bf R})=1 | u=1) | \nonumber \\
& = \textrm{Adv}_{\mathcal{D}}(\mathcal{A}_{PIA}). \label{eqn:tvadv_PI}
\end{align}
\end{small}%

With Equations~\ref{eqn:contWass_PI}-\ref{eqn:tvadv_PI}, we thus have 
\begin{small}
\begin{align*}
W_1 (Cf(\mathcal{D}_{{\bf X} | u=0}), Cf(\mathcal{D}_{{\bf X} | u=1})) \leq 2R \cdot  C_L \cdot \textrm{Adv}_{\mathcal{D}}(\mathcal{A}_{PIA}).
\end{align*}
\end{small}%

Further, 
as $\{{y_i}\}$'s are independent and using Lemma~\ref{lem:berprob} on Bernoulli random variables ${y_i}$ and $Cf({\bf x}_i)$, we have 
\begin{small}
\begin{align}
& W_1 (\mathcal{D}_{{\bf y}|u=a}, Cf(\mathcal{D}_{{\bf X} | u=a})) \notag \\
& = {\frac{1}{|{\bf y}|} \sum_{({\bf x}_i, {y_i}) \in ({\bf X}, {\bf y})} W_1 (\mathcal{D}_{{y_i} |u=a}, Cf(\mathcal{D}_{{\bf x}_i | u}))} \notag \\ 
& = {\frac{1}{|{\bf y}|} \sum_{({\bf x}_i, {y_i}) \in ({\bf X}, {\bf y})} |\textrm{Pr}_{\mathcal{D}}({y_i}=1 | u=a) - \textrm{Pr}_{\mathcal{D}}({Cf({\bf x}_i})=1 | u=a)) |} \nonumber \\
& = {\frac{1}{|{\bf y}|} \sum_{({\bf x}_i, {y_i}) \in ({\bf X}, {\bf y})} |\mathbb{E}_{\mathcal{D}}[{y_i}|u=a] - \mathbb{E}_{\mathcal{D}}[Cf({\bf x}_i)| u=a]|} \nonumber \\
& \leq {\frac{1}{|{\bf y}|} \sum_{({\bf x}_i, {y_i}) \in ({\bf X}, {\bf y})} \mathbb{E}_{\mathcal{D}}[|{y_i}-Cf({\bf x}_i)| |u=a]} \nonumber \\
& = {\frac{1}{|{\bf y}|} \sum_{({\bf x}_i, {y_i}) \in ({\bf X}, {\bf y})} \textrm{Pr}_{\mathcal{D}}({y_i} \neq Cf({\bf x}_i) | u=a)}, 
\label{eqn:CEloss_PI}
\end{align}
\end{small}%
where again the first equality uses independence of $\{y_i\}$'s.

Finally, by combining Equations~\ref{eqn:keytri_PI}-\ref{eqn:CEloss_PI}, 
\begin{small}
\begin{align}
    \Delta_{{\bf y}|u} & \leq \frac{1}{|{\bf y}|} \sum_{({\bf x}_i, {y_i}) \in ({\bf X}, {\bf y})} \textrm{Pr}_{\mathcal{D}}({y_i} 
    \neq Cf({\bf x}_i) | u=0) \notag \\ 
    & + \frac{1}{|{\bf y}|} \sum_{({\bf x}_i, {y_i}) \in ({\bf X}, {\bf y})} \textrm{Pr}_{\mathcal{D}}({y_i} \neq Cf({\bf x}_i) | u=1) 
    \notag \\  
    & 
    + 2R \cdot C_L \cdot \textrm{Adv}_{\mathcal{D}}(\mathcal{A}_{PIA}). 
\end{align}
\end{small}%
Hence, 
$
    \mathrm{Risk}_{PI}(C \circ f) = \frac{1}{|{\bf y}|} \sum_{({\bf x}_i, {y_i}) \in ({\bf X}, {\bf y})} \textrm{Pr}_{\mathcal{D}}({y_i} 
    \neq Cf({\bf x}_i)) 
    \geq \Delta_{{\bf y}|u} - 2R \cdot C_L \cdot \textrm{Adv}_{\mathcal{D}}(\mathcal{A}_{PIA}).
$ 

\end{proof}

\subsection{Proof of Theorem \ref{thm:uptradeoff_dr} for DRAs}
\label{supp:uptradeoff_dr}

\begin{proof}

We denote $\mathcal{D}_{{\bf x}|y}$ as the conditional data sample distribution of $\mathcal{D}$ given $y$ and $\mathcal{P}_{\bm{\delta}}$ as the perturbation distribution on $\bm{\delta}$. We further denote $\mathcal{D}_{{\bf x}|y} \oplus \mathcal{P}_{\bm{\delta}}$ as the combined distribution of the two. 
Accordingly, ${\bf r}' = {\bf r} + \bm{\delta}$ is sampled from this combined distribution. 
By leveraging the triangle inequalities of the 1-Wasserstein distance, we have
{
\small
\begin{align} 
& W_1 (\mathcal{D}_{y=0|\mathbf{x}}, \mathcal{D}_{y=1|\mathbf{x}}) 
\leq W_1 (\mathcal{D}_{y=0|\mathbf{x}}, Cf(\mathcal{D}_{\mathbf{x}|y=0} \oplus \mathcal{P}_{\bm{\delta}})) \nonumber \\ 
& \quad + W_1 (Cf(\mathcal{D}_{\mathbf{x}|y=0} \oplus \mathcal{P}_{\bm{\delta}}), Cf(\mathcal{D}_{\mathbf{x}|y=1} \oplus \mathcal{P}_{\bm{\delta}})) \nonumber \\ 
& \quad +  W_1 (Cf(\mathcal{D}_{\mathbf{x}|y=1} \oplus \mathcal{P}_{\bm{\delta}}), \mathcal{D}_{y=1|\mathbf{x}}).
\label{eqn:keytri_DR}
\end{align}
}%
Using Lemma~\ref{lem:berprob} on Bernoulli r.v. $y$, we have 
\begin{small}
{
\begin{align} 
\label{eqn:berprob2}
& W_1 (\mathcal{D}_{y=0|{\bf x}}, \mathcal{D}_{y=1|{\bf x}}) = |\textrm{Pr}_{\mathcal{D}}({y=0}) - \textrm{Pr}_{\mathcal{D}}({y=1}) | = \Delta_{y}.
\end{align} 
}%
\end{small} %
Using Lemma~\ref{lem:contWass} on the contraction of the 1-Wasserstein distance and that $\|C\|_L \leq C_L$, we have 
\begin{small}
\begin{small} \begin{align} 
\label{eqn:contWass}
& W_1 (Cf(\mathcal{D}_{{\bf x}|y=0} \oplus \mathcal{P}_{\bm{\delta}} ), Cf(\mathcal{D}_{{\bf x}|y=1} \oplus \mathcal{P}_{\bm{\delta}})) \notag \\
& \leq C_L \cdot W_1(f(\mathcal{D}_{{\bf x}|y=0} \oplus \mathcal{P}_{\bm{\delta}}), f(\mathcal{D}_{{\bf x}|y=1} \oplus \mathcal{P}_{\bm{\delta}})).
\end{align} \end{small} %
\end{small}
Using Lemma~\ref{lem:tvWass} with $\max_{{\bf r}'} \|{\bf r}' \| \leq R'$, we have 
\begin{small}
\begin{small} \begin{align} 
\label{eqn:tvWass}
& W_1(f(\mathcal{D}_{{\bf x}|y=0} \oplus \mathcal{P}_{\bm{\delta}}), f(\mathcal{D}_{{\bf x}|y=1} \oplus \mathcal{P}_{\bm{\delta}})) \notag \\
& \leq 2R' \cdot d_{TV}(f(\mathcal{D}_{{\bf x}|y=0} \oplus \mathcal{P}_{\bm{\delta}}), f(\mathcal{D}_{{\bf x}|y=1} \oplus \mathcal{P}_{\bm{\delta}})). 
\end{align} \end{small} %
\end{small}%
We further show that $d_{TV}(f(\mathcal{D}_{{\bf x}|y=0} \oplus \mathcal{P}_{\bm{\delta}}), f(\mathcal{D}_{{\bf x}|y=1} \oplus \mathcal{P}_{\bm{\delta}})) = \textrm{Adv}_{\mathcal{D}}(\mathcal{A}_{DRA})$.
Specifically, 
\begin{small}
{
\begin{align} 
& d_{TV}(f(\mathcal{D}_{{\bf x}|y=0} \oplus \mathcal{P}_{\bm{\delta}}), f(\mathcal{D}_{{\bf x}|y=1} \oplus \mathcal{P}_{\bm{\delta}})) \notag \\
& = \max_{E} |\textrm{Pr}_{f(\mathcal{D}_{{\bf x}|y=0} \oplus \mathcal{P}_{\bm{\delta}})}(E) - \textrm{Pr}_{f(\mathcal{D}_{{\bf x}|y=1} \oplus \mathcal{P}_{\bm{\delta}})}(E)| \nonumber \\
& = \max_{A_E \in \mathcal{A}_{DRA}} | \textrm{Pr}_{{\bf r}' \sim f(\mathcal{D}_{{\bf x}|y=0} \oplus \mathcal{P}_{\bm{\delta}})}(A_E({\bf r}')= {\bf x}) 
- \textrm{Pr}_{{\bf r}' \sim f(\mathcal{D}_{{\bf x}|y=1} \oplus \mathcal{P}_{\bm{\delta}})}(A_E({\bf r}')= {\bf x}) | \nonumber \\
& = \max_{A_E \in \mathcal{A}_{DRA}} | \textrm{Pr}(A_E({\bf r}')= {\bf x} | {y=0}) - \textrm{Pr}(A_E({\bf r}')= {\bf x} | {y=1}) | \nonumber \\
& = \textrm{Adv}_{\mathcal{D}}(\mathcal{A}_{DRA}), \label{eqn:tvadv}
\end{align}
}%
\end{small}%
where the first equation uses the definition of TV distance, and  $A_E(\cdot)$ is the characteristic function of the event $E$ in the second equation.
With Equations~\ref{eqn:contWass}-\ref{eqn:tvadv}, we have 
$$W_1 (Cf(\mathcal{D}_{{\bf x}|y=0} \oplus \mathcal{P}_{\bm{\delta}}), Cf(\mathcal{D}_{{\bf x}|y=1} \oplus \mathcal{P}_{\bm{\delta}})) \leq 2R' \cdot  C_L \cdot \textrm{Adv}_{\mathcal{D}}(\mathcal{A}_{DRA}).$$ 
Furthermore, using Lemma~\ref{lem:berprob} on Bernoulli r.v.s $y$ and $Cf({\bf x})$: 
\begin{small} \begin{align} 
& W_1 (\mathcal{D}_{y=1|{\bf x}}, Cf(\mathcal{D}_{{\bf x}|y=1}) \oplus \mathcal{P}_{\bm{\delta}}) \notag \\
& = |\textrm{Pr}_{\mathcal{D}}(y=1) - \textrm{Pr}_{\mathcal{D} \times \mathcal{P}} ({C(f({\bf x}) +\bm{\delta})}=1|y=1) | \nonumber \\
& = |\mathbb{E}_{\mathcal{D}}[y] - \mathbb{E}_{\mathcal{D} \times \mathcal{P}}[C(f({\bf x}) +\bm{\delta})|y=1]| \nonumber \\
& \leq \mathbb{E}_{\mathcal{D} \times \mathcal{P}}[|y-C(f({\bf x}) +\bm{\delta})|y=1|]. 
\nonumber \\ & 
= \textrm{Pr}_{\mathcal{D}\times \mathcal{P}}(y \neq C(f({\bf x}) +\bm{\delta})|y=1).
\label{eqn:CEloss_DR_1}
\end{align} 
\end{small}%
Similarly, 
\begin{small} \begin{align} 
& W_1 (\mathcal{D}_{y=0|{\bf x}}, Cf(\mathcal{D}_{{\bf x}|y=0}) \oplus \mathcal{P}_{\bm{\delta}}) 
\leq \textrm{Pr}_{\mathcal{D}\times \mathcal{P}}(y \neq C(f({\bf x}) +\bm{\delta})|y=0).
\label{eqn:CEloss_DR_2}
\end{align} 
\end{small} %
Finally, by combining Equation~\ref{eqn:keytri_DR}-Equation~\ref{eqn:CEloss_DR_2}, 
\begin{small} 
\begin{align} 
    \Delta_{y} & \leq \textrm{Pr}_{\mathcal{D} \times \mathcal{P}}(y \neq C(f({\bf x}) +\bm{\delta}|y=1) + 2R' \cdot C_L \cdot \textrm{Adv}_{\mathcal{D}}(\mathcal{A}_{DRA}) \notag \\ 
    & \quad +\textrm{Pr}_{\mathcal{D} \times \mathcal{P}}(y \neq C(f({\bf x}) +\bm{\delta})|y=0)
\end{align} 
\end{small} %
Hence, 
$
    \mathrm{Risk}_{DR}(C \circ f) = \textrm{Pr}_{\mathcal{D} \times \mathcal{P}}(y \neq C(f({\bf x}) +\bm{\delta})|y=1)  + \textrm{Pr}_{\mathcal{D} \times \mathcal{P}}(y \neq C(f({\bf x}) +\bm{\delta})|y=0)
    \geq \Delta_{y} - 2R' \cdot C_L \cdot \textrm{Adv}_{\mathcal{D}}(\mathcal{A}_{DRA}).
$ 
\end{proof}

\section{Datasets and Network Architectures}
\label{appendix:Datasets}

\subsection{Detailed Dataset Description}
\label{app:datades}

\textbf{CIFAR-10~\cite{Krizhevsky09learningmultiple}.}
It contains 60,000 colored images of 32x32 resolution. 
The dataset consists of images belonging to 10 classes: airplane, automobile, bird, cat, deer, dog, frog, horse, ship and truck. There are 6,000 images per class.  
In this dataset, the primary task is to predict the label of the image. 

\noindent  \textbf{Purchase100~\cite{nasr2018machine}.} The dataset includes 197,324 shopping records, where 
each data record corresponds to one costumer and has 600 binary features (each corresponding to one item). Each feature reflects if the item is purchased by the costumer or not. The data is clustered into 100 classes and the task is to predict the class for each costumer. 

\noindent  \textbf{Texas100~\cite{shokri2017membership}.} This dataset includes hospital discharge data
published by the Texas Department of State Health Services. 
Data records have features about the external causes of injury (e.g., suicide, drug misuse), the diagnosis (e.g., schizophrenia, illegal abortion), the procedures the patient underwent (e.g., surgery), and generic information such as gender, age, race, hospital ID, and length of stay. 
The dataset contains 67,330 records
and 6,170 binary features which represent the 100 most frequent medical procedures. The records are clustered into 100 classes, each representing a type of patient.

\noindent  \textbf{Census~\cite{suri2022formalizing}}. It consists of several categorical and numerical attributes like age, race, education level to predict whether a
person makes over \$50K a year. We focus on the ratios of females (sex) as property.

\noindent  \textbf{RSNA Bone Age~\cite{suri2022formalizing}}. It contains X-Ray images of hands, with the task being predicting the patients’ age in months. We
convert the task to binary classification based on an age threshold, 132 months, as a binary primary task. We focus on the ratios of the females (available as metadata) as properties.

 \noindent  \textbf{CelebA~\cite{liu2015faceattributes}}. It is a large-scale wild face attributes dataset that consists of 200,000 RGB celebrity images. Each image has 40 annotated 
 binary attributes  such as ``gender", ``race". We focus on smile detection as a primary task and the ratios of females (sex) as the private property.

\noindent {\bf CIFAR100~\cite{Krizhevsky09learningmultiple}.} This dataset has 100 classes containing 600 colored images each, with size 32x32. There are 500 training images and 100 testing images per class. The learning task is classifying the 100 classes.

\noindent \textbf{Activity~\cite{misc_human_activity_recognition_using_smartphones_240}.} This dataset contains data collected from smartphones carried by a person while performing 1 of 6 activities. There is a total of 516 features in a single row format. All features are pre-normalized and bounded by [-1, 1]. The data collected include statistical summaries of sensory data collected by the phone, including acceleration, orientation, etc.

\begin{table}[!t]  
  \centering
  \footnotesize 
  \caption{Training and test sets for primary and MIA tasks.}
  \begin{tabular}{|c|c|c|c|}
    \cline{2-4}
    \multicolumn{1}{c|}{} & {\bf CIFAR10} & {\bf Purchase100} & {\bf Texas100} \\ \hline
    Utility training set & 25,000    & 98,662   & 33,665  \\ \hline
    Utility test set & 25,000   & 98,662   & 33,665  \\ \hline
    Attack training set & 40,000   & 157,859   &  53,864 \\ \hline
    Attack test set & 10,000  & 39,465  & 13,466 \\ \hline
  \end{tabular}%
  \label{tbl:dataset_MIAs}
  \vspace{-2mm}
\end{table}

\begin{table}[!t]  
  \centering
  \footnotesize 
  \addtolength{\tabcolsep}{-3.5pt}
  \caption{Training and test sets for primary and PIA tasks.}
  
  \begin{tabular}{|c|c|c|c|}
    \cline{2-4}
    \multicolumn{1}{c|}{} & {\bf Census} & {\bf RSNA} & {\bf CelebA} \\ \hline
    Subset size & [2,32k]   & [2,100]   & [2,20]  \\ \hline
    female ratios & $\{0.2,0.3,\cdots,0.5\}$  & $\{0.2,0.3,\cdots,0.8\}$  & $\{0.0,0.1,\cdots,1.0\}$  \\ \hline
     Attack train set & 8k subsets & 14k subsets & 22k subsets \\ \hline
    Attack test set & 2k subsets & 3.5k subsets & 5.5k subsets \\ \hline
    Utility train set & data in 8k subsets &  in 14k subsets   &   in 22k subsets \\ \hline
    Utility test set  & data in 2k subsets  &  in 3.5k subsets  & in 5.5k subsets \\ \hline
  \end{tabular}
  \label{tbl:dataset_PIAs}
   \vspace{-2mm}
\end{table}

\begin{table}[!t]  
  \centering
  \footnotesize 
  \caption{Training and test sets for primary and DRA tasks.}
  \begin{tabular}{|c|c|c|c|}
    \cline{2-4}
    \multicolumn{1}{c|}{} & {\bf 
    CIFAR10} & {\bf CIFAR100} & {\bf Activity} \\ \hline
   Utility/Attack training set & 50,000  & 50,000  & 7,352 \\ \hline
    Utility test set & 10,000   & 10,000  & 2,947  \\ \hline
    Attack test set & 50  & 50  & 50  \\ \hline
  \end{tabular}%
  \label{tbl:dataset_DRAs}
\end{table}

\subsection{More Experimental Setup}
\label{app:setup}

{\bf Training and testing:}
Table~\ref{tbl:dataset_MIAs}-Table~\ref{tbl:dataset_DRAs} show the utility training/test and attack training/test sets.  

\vspace{+0.01in} 
\noindent {\bf Differential Privacy (DP) against MIAs:} 
DP provides an upper bound on the success of any
MIA. We can 
add noise in several ways (e.g., to input data, model parameters, gradients, latent features, output scores) to ensure DP. 
Note that there exists an inherent trade-off between utility and privacy: a larger added noise often leads to a higher level of privacy protection, but incurs a larger utility loss. 
Here, we propose to use the below two ways. 

\begin{itemize}[leftmargin=*]

\vspace{-2mm}

\item {\bf DP-SGD~\cite{abadi2016deep}:} 
\emph{1) DP-SGD training:}   
It clips gradients (with a gradient norm bound) and adds Gaussian noise to the gradient in each SGD round when training the ML model (i.e., encoder + utility network). More details can be seen in Algorithm 1 in \cite{abadi2016deep}.  
After training, the model ensures DP guarantees and the encoder is published.  
\emph{2) Attack training:} The attacker obtains the representations of the attack training data via querying the trained encoder and uses these representations to train the MIA classifier.  
\emph{3) Defense/attack testing:} 
The utility test set is used to
obtain the utility via querying the trained ML model; and the attack test set to obtain the MIA accuracy via querying the trained encoder and trained MIA classifier. 

We used the Opacus library ({\url{https://opacus.ai/}}),
a PyTorch extension that enables training models with DP-SGD {and dynamically tracks privacy budget and utility. In the experiments, we tried $\epsilon$ in DP-SGD from 0.5 to 16.}

\vspace{-2mm}

\item {\bf DP-encoder:} \emph{1) Normal training:} It first trains the encoder + utility network using the (utility) training set. The encoder is then frozen and can be used to produce data representations when queried by data samples.  
\emph{2) Defense via adding noise to the representations:} We add Gaussian noise (i.e., $\mathcal{N}(0,\sigma^2)$) to the representations by querying the encoder with the attack training data to produce the noisy representations. \emph{Notice that, since the Gaussian noises are injected to the matrix-outputs (data representations), if needed, the actual DP guarantee (i.e., privacy bounds) can be derived via Rényi Differential Privacy \cite{Mironov17}, similar to the theoretical studies in \cite{WangSFSH22,r2dpccs220}. We skip the details here since this work does not focus on the derivation for the privacy bounds of DP-encoder.} 
\emph{3) Attack training:} 
The attacker 
uses the noisy representations of attack training data to train the MIA classifier.  
\emph{4) Defense/attack testing:} 
We use the utility test set to
obtain the utility via querying the trained encoder
and utility network; and use the attack test set to obtain the MIA accuracy on the trained encoder and trained MIA classifier. 
We call this \emph{DP-encoder} as we add noise to the representations outputted by the well-trained encoder. 
\vspace{-2mm}
\end{itemize}

\vspace{+0.01in} 
\noindent {\bf DP-encoder against PIAs:} We follow the strategy in DP against MIAs and choose the DP-encoder, as DP-SGD is ineffective in this setting~\cite{suri2022formalizing}.  
The only difference is that we now add Gaussian noise to the mean-\emph{aggregated} representation of a subset, instead of the individual representation.  

\vspace{+0.01in} 
\noindent {\bf Data reconstruction attack/defense on shallow encoder:} As shown in \cite{he2019model}, when the encoder is deep, it is difficult  for the attacker to reconstruct the input data from the representation. To ensure DRAs be  effective, we use a shallow 2-layer encoder. As a result, this makes the defense more challenging.

\subsection{Network Architectures} 
\label{app:netarch}

Network architectures for the neural networks used in defense training are detailed in Table~\ref{tab:arch_MIAs}-Table \ref{tab:arch_DRAs}.

\begin{table}[!t]\renewcommand{\arraystretch}{0.9}
\footnotesize
\addtolength{\tabcolsep}{-5pt}
\caption{Network architectures for the used datasets (MIAs)}
  \centering
    \begin{tabular}[t]{c|c|c}
      \hline
    \centering\textbf{Encoder} &\centering\textbf{Membership Network} &\textbf{Utility Network}\\
    \hline
    \hline
    \multicolumn{3}{c}{\bf CIFAR10}\\
    \hline
    \hline
    conv3-64 (ResNet Block 6) & 2xconv3-512 & 2xconv3-512 
    \\ BN-64 & (ResNet Block 6) & avgpool \& Flatten \\
    MaxPool& avgpool \& Flatten \\
    \cline{1-3}
    2xconv3-64  (ResNet Block) & Linear-512 & Linear-512\\
    \cline{1-3}
    2xconv3-64  (ResNet Block)&  Linear-2 classes  & Linear-$\#$labels \\
    \cline{1-3}
    3xconv3-64  (CustomBlock)&  &  \\
    \cline{1-1}
    2xconv3-128 (ResNet Block) & 
    & \\
    \cline{1-1}
    2xconv3-128 (ResNet Block)&  &  \\
    \cline{1-1}
    3xconv3-128  (CustomBlock)&  &  \\
    \cline{1-1}
    2xconv3-256 (ResNet Block)&  &  \\
    \cline{1-1}
    conv3-512 (ResNet Block) &  & \\
    \cline{1-3}
    \hline
    \hline
    \multicolumn{3}{c}{\bf Purchase100}\\
    \hline
    \hline
     Linear-600 \& Tanh & Linear-1024  \& ReLU &Linear-128 \& ReLU \\ 
     
    \cline{1-3}
    Linear-512 \& Tanh  & Linear-512 \& ReLU & Linear-256 \& ReLU \\ 
    \cline{1-3}
    Linear-256 \& Tanh & Linear-256 \& Tanh & Linear-128 \\ 
    
    \cline{2-3}
    & Linear-128 \& Tanh & 3xLinear-256\\ 
    \cline{2-3}
    
    & Linear-2 classes &Linear-$\#$labels\\
    \cline{1-3}
    \hline
    \hline
    \multicolumn{3}{c}{\bf Texas 100}\\
    \hline
    \hline
     Linear-6169  \& ReLU & Linear-128  \& ReLU & Linear-128\\ 
     \cline{1-3}
    Linear-1024  \& ReLU & Linear-256  \& ReLU & Linear-$\#$labels \\ 
    \cline{1-3}
    Linear-2048  \& ReLU & Linear-128  \& ReLU & \\
    \cline{1-2}
    Linear-1024  \& ReLU &Linear-128 &  \\ 
    \cline{1-2}
    Linear-512  \& ReLU & Linear-2 classes  & \\ 
     \cline{1-1}
    Linear-256  \& ReLU &  & \\ 
     \cline{1-1}
    Linear-128  \& ReLU  &  & \\ 
    \hline
   \end{tabular}
   \label{tab:arch_MIAs}
  \vspace{-2mm}
\end{table}

\begin{table}[!t]\renewcommand{\arraystretch}{0.9}
\footnotesize
\addtolength{\tabcolsep}{-5pt}
\caption{Network architectures for the used datasets (PIAs). BN: BatchNorm, SC: Shortcut Connection}
  \centering
    \begin{tabular}[t]{c|c|c}
      \hline
    \centering\textbf{Encoder} &\centering\textbf{Property Network} &\textbf{Utility Network}\\
    \hline
    \hline
    \multicolumn{3}{c}{\bf RSNA Bone Age}\\
    \hline
    \hline
    conv2-64 \& BN-64 & Linear-1024&con2d-1024 \\ 
      ReLU \& MaxPool & ReLU& ReLU\\ 
    \cline{1-3}
    DenseBlock-64& Linear-512& AdaptiveAvgPool2d \\ TransitionLayer-64 & ReLU& \\
    \cline{1-3}
    DenseBlock-128& Linear-256 & \\ TransitionLayer-128& ReLU &Linear-$\#$labels\\ 
    \cline{1-3}
    DenseBlock-256& Linear-128& \\ TransitionLayer-256 & ReLU & \\
    \cline{1-3}
    DenseBlock-512& Linear-$\#$property values & \\ AdaptiveAvgPool & & \\ 
    \cline{1-3}
    \hline
    \hline
    \multicolumn{3}{c}{\bf Census}\\
    \hline
    \hline
     conv1d-256 \& ReLU & Linear-128  \& ReLU & Conv1d-64  \& ReLU \\ 
    \cline{1-3}
    conv1d-128  \& ReLU & Linear-64  \& ReLU &Conv1d-32  \& ReLU \\
    \cline{1-3}
    AdaptiveMaxPool1d-1 & Linear-$\#$property values & AdaptiveAvgPool1d-1  \\ 
    \cline{1-3}
    Linear-128 &  & AdaptiveAvgPool1d-1 \\
    \cline{1-3}
    AdaptiveMaxPool1d-1 &  & Linear-$\#$labels \\ 
    \cline{1-3}
    \hline
    \hline
    \multicolumn{3}{c}{\bf CelebA}\\
    \hline
    \hline
     conv3-64 \&  BN-64 & FCResidualBlock & Linear-512\\
     ReLU \& MaxPool&  (512->256)& \\ 
     \cline{1-3}
     \bf{Residual Block 1\& 2} & FCResidualBlock & Linear-$\#$labels\\
    conv3-64 \& BN-64 &  (256->128) & \\
    ReLU \& conv3-64 & & \\ 
    BN-64 \& SC \& ReLU & & \\ 
    \cline{1-2}
    \bf{Residual Block 3\& 4} & Linear-128& \\
    conv3-128 \& BN-128 &  & \\
    ReLU \& conv3-128 & & \\
    BN-128 \& SC \& ReLU & & \\
    \cline{1-2}
    \bf{Residual Block 5\& 6} & Linear-$\#$property values & \\
    conv3-256 \& BN-256 & & \\
    ReLU \& conv3-256 & & \\ 
    BN-256 \& SC \& ReLU & & \\
    \cline{1-1}
    \bf{Residual Block 7\& 8} & & \\
    conv3-512 \& BN-512 & & \\ 
    ReLU \& conv3-512 & & \\ 
    BN-512 \& SC \& ReLU & & \\
    \hline
   \end{tabular}
   \label{tab:arch_PIAs}
\end{table}

\begin{table}[!t]\renewcommand{\arraystretch}{0.85}
\footnotesize
\addtolength{\tabcolsep}{-5pt}
\caption{Network architectures for the used datasets (DRA)}
  \centering
    \begin{tabular}[t]{c|c|c}
      \hline
    \centering\textbf{Encoder} &\centering\textbf{Reconstruction Network} &\textbf{Utility Network}\\
    \hline
    \hline
    \multicolumn{3}{c}{\bf CIFAR10 (CIFAR100)}\\
    \hline
    \hline
    conv2-32 \& BN-32 & conv2-32 \& BN-32 & conv2-64 \& BN-64 \\
    \cline{1-3}
    conv2-32 \& BN-32 & & conv2-128 \& BN-128 \\
     & & ReLU \\
    \cline{1-3}
    & & conv2-128 \& BN-128 U\\ 
     & & ReLU \\
    \cline{3-3}
    & & conv2-128 \& BN-128 \\ 
     & & ReLU \\
    \cline{3-3}
    & & conv2-128 \& BN-128 \\
     & & ReLU \\
    \cline{3-3}
    & & conv2-64 \& BN-64 \\
     & & ReLU \\
    \cline{3-3}
    \cline{3-3}
    & & Flatten\\
    \cline{3-3}
    & & Linear-1024 \& BN-1024 \\ 
    & & ReLU\\
    \cline{3-3}
    & & Linear-64 (256)  \\ & & BN-64 (256) \\ & & ReLU\\
    \cline{3-3}
    & & Linear-$\#$labels\\
    \cline{3-3}
    \hline
    \hline
    \multicolumn{3}{c}{\bf Activity}\\
    \hline
    \hline
     Linear-256  & Linear-128  & Linear-32 \\ 
     BN-256 \& ReLU & BN-128 \& ReLU & BN-32\& ReLU \\
    \cline{1-3}
     Linear-128& Linear-32 & Linear-12\\BN-128 \& ReLU & BN-32 \& ReLU & BN-12 \& ReLU \\ 
    \cline{1-3}
     Linear-32& & Linear-\#labels \\BN-32 \& ReLU & & \\ 
    \hline
   \end{tabular}
   \label{tab:arch_DRAs}
  \vspace{-4mm}
\end{table}

\begin{table}[!t]
\footnotesize
\centering
\caption{Impact of the size of the encoder on \texttt{\name} defense results against MIAs (we fix the utility classifier and privacy protection/attack classifier). A larger encoder can have (marginally) better MIA performance without defense.}   
\addtolength{\tabcolsep}{-2pt}
\centering
\vspace{+2mm}
\begin{minipage}{.75\linewidth}
\begin{tabular}{|l|c|c|c|}
\multicolumn{4}{c}{\bf CIFAR10}\\
          \hline
\textbf{ \#parameters} & \( \lambda \) & \textbf{Utility} & \textbf{MIA Acc} \\
\hline
Encoder: 4,898,112 & 0 & 77.83\% & 63.62\% \\
Utility: 4,725,770 & 0.25 & 77.53\% & 56.2\% \\
Pri/Atk: 4,721,153 & 0.5 & 77\% & 52.08\% \\
               & 0.75 & 76\% & 50\% \\
               & 1 & 10\% & 49.99\% \\
\hline
Encoder: 6,448,192 & 0 & 80.88\% & 68.78 \% \\
Utility: 4,725,770 & 0.25 & 78.66\% & 55.39\% \\
Pri/Atk: 4,721,153 & 0.5 & 77.80\% & 51.78\% \\
               & 0.75 & 76.85\% & 50.31\% \\
               & 1 & 13.44\% & 50\% \\
\hline
Encoder: 7,018,944 & 0 & 78.85\% & 70.05\% \\
Utility: 4,725,770 & 0.25 & 78.15\% & 55.88\% \\
Pri/Atk: 4,721,153 & 0.5 & 78.00\% & 53.53\% \\
               & 0.75 & 77.20\% & 51.08\% \\
               & 1 & 20\% & 50\% \\
\hline
\end{tabular}
\end{minipage}
\vfill
\begin{minipage}{.75\linewidth}
\begin{tabular}{|l|c|c|c|}
\multicolumn{4}{c}{\bf Purchase100}\\
          \hline
\textbf{\#parameters} & \( \lambda \) & \textbf{Utility} & \textbf{MIA Acc} \\
\hline
Encoder: 471,936 & 0 & 80.08\% & 59.31\% \\
Utility: 25,700 & 0.25 & 80.08\% & 51\% \\
Pri/Atk: 66,049 & 0.5 & 80\% & 50\% \\
            & 0.75 & 80\% & 50\% \\
            & 1 & 20\% & 49.99\% \\
\hline
Encoder: 734,592 & 0 & 81.65\% & 68.36\% \\
Utility: 25,700 & 0.25 & 80.87\% & 60\% \\
Pri/Atk: 66,049 & 0.5 & 80\% & 51\% \\
            & 0.75 & 78\% & 50\% \\
            & 1 & 20\% & 50\% \\
\hline
Encoder: 1,304,448 & 0 & 80.65\% & 69.65\% \\
Utility: 25,700 & 0.25 & 80.87\% & 61\% \\
Pri/Atk: 66,049 & 0.5 & 80\% & 53\% \\
            & 0.75 & 78\% & 50\% \\
            & 1 & 20\% & 50\% \\
\hline
\end{tabular}
\end{minipage}
\vfill
\begin{minipage}{.75\linewidth}
\begin{tabular}{|l|c|c|c|}
\multicolumn{4}{c}{\bf Texas100}\\
          \hline
\textbf{\#parameters} & \( \lambda \) & \textbf{Utility} & \textbf{MIA Acc} \\
\hline
Encoder: 7,532,416 & 0 & 46.52\% & 68.96\% \\
Utility: 25,700 & 0.25 & 46.20\% & 60\% \\
Pri/Atk: 66,049 & 0.5 & 46\% & 50\% \\
            & 0.75 & 46\% & 50\% \\
            & 1 & 2\% & 50\% \\
\hline
Encoder: 9,106,304 & 0 & 49.76\% & 70.21\% \\
Utility: 25,700 & 0.25 & 49.05\% & 61\% \\
Pri/Atk: 66,049 & 0.5 & 46\% & 50\% \\
            & 0.75 & 46\% & 50\% \\
            & 1 & 2\% & 50\% \\
\hline
Encoder: 11,204,480 & 0 & 51.05\% & 69.10\% \\
Utility: 25,700 & 0.25 & 51\% & 61\% \\
Pri/Atk: 66,049 & 0.5 & 51\% & 50\% \\
            & 0.75 & 46.82\% & 50\% \\
            & 1 & 2\% & 50\% \\
\hline
\end{tabular}
\end{minipage}
\label{tab:FE_depth_impact}
\end{table}

\subsection{More Experimental Results}
\label{app:moreresults}

\noindent {\bf More results on defending against MIAs:}
Table~\ref{tab:FE_depth_impact} shows the \texttt{\name} results against MIAs, with varying sizes of the encoder. We can see a larger encoder can have (marginally) better MIA performance without defense. 
One possible is that a larger model can better memorize the training data. 
Table~\ref{tab:moreDPres} shows more DP results (vs varying $\epsilon$'s) against MIAs.  
We can see \texttt{\name} obtains higher utility than DP methods under the same privacy protection performance. 

\begin{figure*}[!t]
\vspace{-2mm}
	\centering
	\subfigure[Utility w/o. defense (85\%)]
{\centering\includegraphics[width=0.23\linewidth]{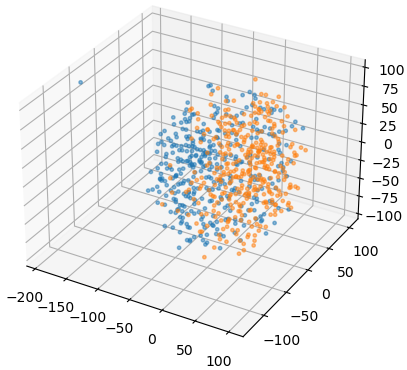}}
	\subfigure[Attack acc. w/o. defense (68\%)]
{\centering\includegraphics[width=0.23\linewidth]{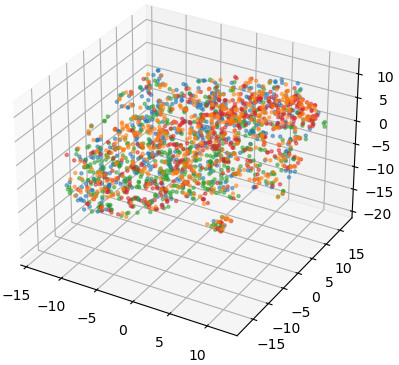}}
\subfigure[Utility w. defense (76\%)]
{\centering\includegraphics[width=0.23\linewidth]{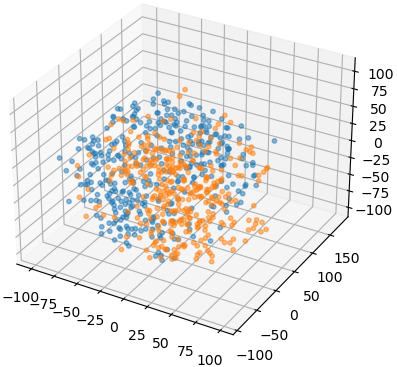}}
		\subfigure[Attack acc. w. defense (34\%)]
{\centering\includegraphics[width=0.23\linewidth]{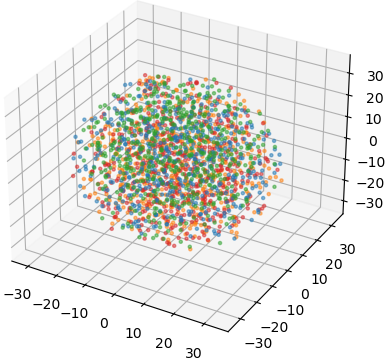}}
	\caption{\texttt{\name} against PIAs: 3D t-SNE embeddings results on the learnt representation on Census Income. 
{Each color indicates a label or a female ratio, and each point is a data representation in (a) and (c) or an aggregated representation in (b) and (d) of a subset in the learning task and in the privacy task, respectively. Same for the other two datasets.}
 } 
	\label{fig:vis_census}
\end{figure*}

\begin{figure*}[!t]
	\centering
	\subfigure[Utility w/o. defense (83\%)]
{\centering\includegraphics[width=0.23\linewidth]{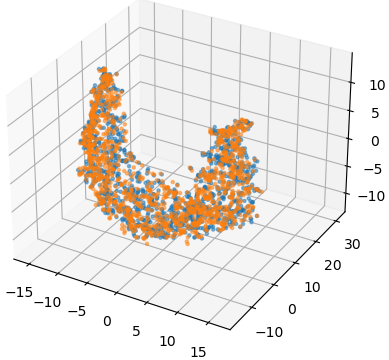}}
	\subfigure[Attack acc. w/o. defense (52\%)]
{\centering\includegraphics[width=0.23\linewidth]{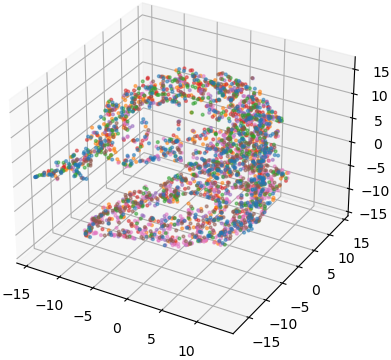}}
\subfigure[Utility w. defense (80\%)]
{\centering\includegraphics[width=0.23\linewidth]{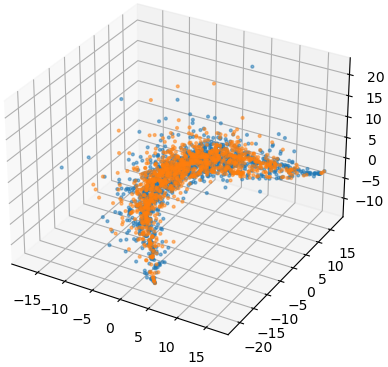}}
		\subfigure[Attack acc. w. defense (19\%)]
{\centering\includegraphics[width=0.23\linewidth]{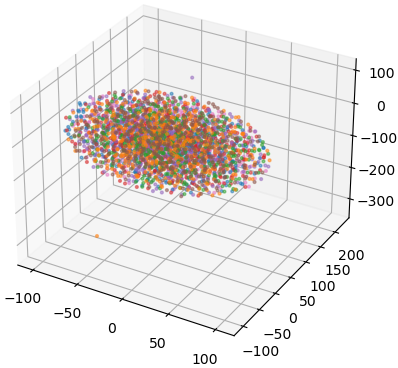}}
	\caption{\texttt{\name} against PIAs: 3D t-SNE embeddings results on the learnt representation on RSNA Bone Age.} 
	\label{fig:vis_rsna}
\end{figure*}

\begin{figure*}[!t]
	\centering
	\subfigure[Utility w/o. defense (91\%)]
{\centering\includegraphics[width=0.23\linewidth]{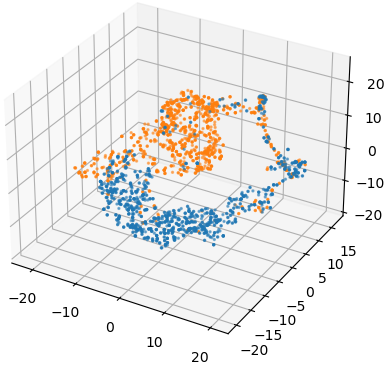}}
	\subfigure[Attack acc. w/o. defense (50\%)]
{\centering\includegraphics[width=0.23\linewidth]{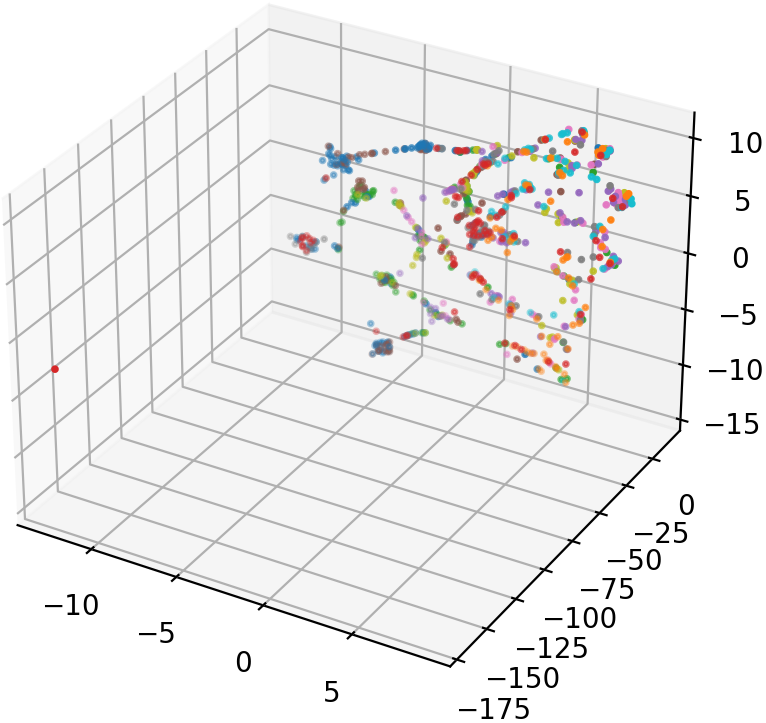}}
\subfigure[Utility w. defense (89\%)]
{\centering\includegraphics[width=0.23\linewidth]{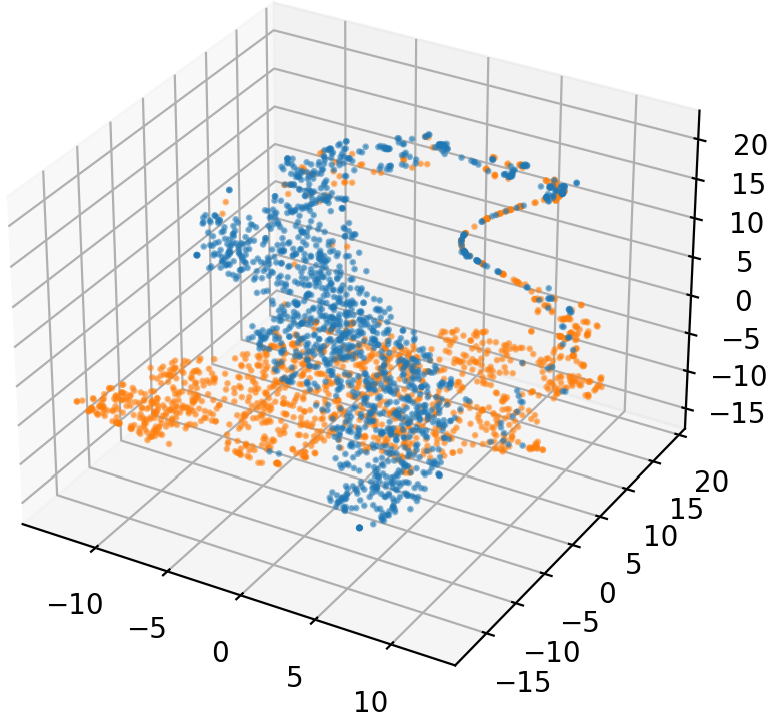}}
		\subfigure[Attack acc. w. defense (11\%)]
{\centering\includegraphics[width=0.23\linewidth]{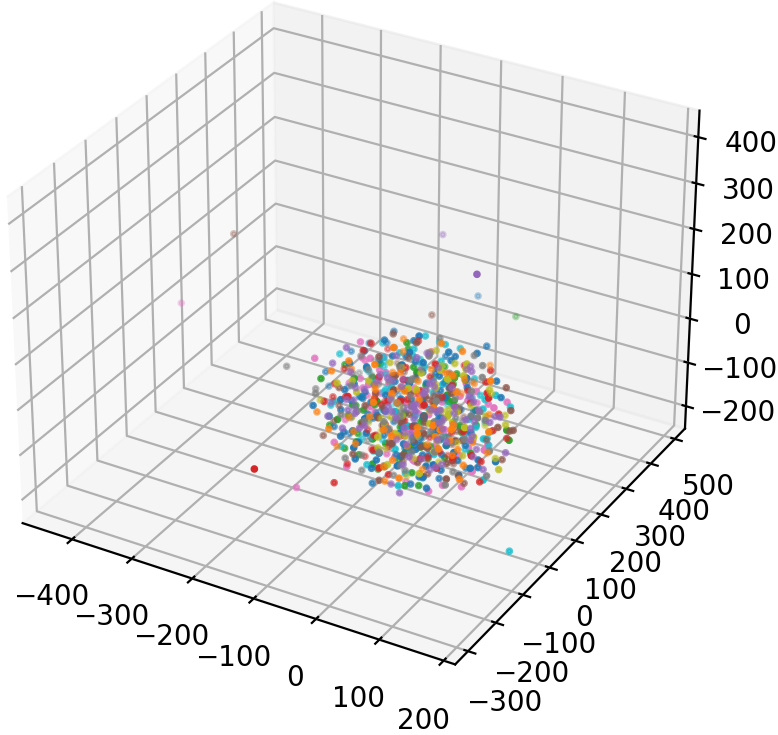}}
	\caption{\texttt{\name} against PIAs: 3D t-SNE embeddings results on the learnt representation on CelebA.} 
	\label{fig:vis_celebA}
\end{figure*}

\vspace{+0.05in}
\noindent {\bf More results on defending against PIAs:} 
Figure~\ref{fig:vis_census}-Figure~\ref{fig:vis_celebA} shows the t-SNE embeddings of \texttt{\name} against PIAs. 
Table~\ref{tab:results_PIAs_substitute} shows \texttt{\name} results against PIAs, where the attacker does not the true (mean) aggregator and use a substitute one (i.e., max-aggregator). 
We can see the attack performance is less effective and \texttt{\name} can yield (close to) random guessing attack performance (when $\gamma=0.75$), with a slight utility loss. This implies the aggregator plays a critical role in designing effective PIAs against \texttt{\name}.

\vspace{+0.05in}
\noindent {\bf More results on defending against DRAs:}
Table~\ref{tbl:more_results_DRAs} shows the \texttt{\name} results against DRAs, where the perturbation distribution is uniform distribution. We observe similar utility-privacy tradeoff in terms of the noise scale $\epsilon$.

\begin{table}[!t]
\footnotesize
\caption{More DP results against MIAs.} 
\addtolength{\tabcolsep}{-3pt}
\centering
\begin{tabular}{|c|cc|cc|cc|}
\hline
\multirow{2}{*}{\bf DP-SGD} & \multicolumn{2}{c|}{\bf CIFAR10}    & \multicolumn{2}{c|}{\bf Purchase100}    & \multicolumn{2}{c|}{\bf Texas100}    \\ \cline{2-7} 
                  & \multicolumn{1}{c|}{\bf Utility} & {\bf MIA Acc} & \multicolumn{1}{c|}{\bf Utility} & {\bf MIA Acc} & \multicolumn{1}{c|}{\bf Utility} & {\bf MIA Acc}  \\ \hline
               $\epsilon=0.5$   & \multicolumn{1}{c|}{46\%} & 50\% & \multicolumn{1}{c|}{40\%} & 52\% & \multicolumn{1}{c|}{11\%} & 51\% \\ \hline
               $\epsilon=1$   & \multicolumn{1}{c|}{48\%} & 51\% & \multicolumn{1}{c|}{48\%} &  54\% & \multicolumn{1}{c|}{15\%} & 52\% \\ \hline
               $\epsilon=2$   & \multicolumn{1}{c|}{59\%} &  55\% & \multicolumn{1}{c|}{53\%} & 57\% & \multicolumn{1}{c|}{26\%} & 54\% \\ \hline
               $\epsilon=4$   & \multicolumn{1}{c|}{61\%} & 57\% & \multicolumn{1}{c|}{60\%} &  59\% & \multicolumn{1}{c|}{33\%} & 55\% \\ \hline
              $\epsilon=8$   & \multicolumn{1}{c|}{65\%} & 59\% & \multicolumn{1}{c|}{71\%} &  62\% & \multicolumn{1}{c|}{39\%} &  57\% \\ \hline
             $\epsilon=16$   & \multicolumn{1}{c|}{68\%} &  62\% & \multicolumn{1}{c|}{78\%} & 66\% & \multicolumn{1}{c|}{45\%} &  59\% \\ \hline 
             \hline 
\multirow{2}{*}{\bf DP-encoder} & \multicolumn{2}{c|}{\bf CIFAR10}    & \multicolumn{2}{c|}{\bf Purchase100}    & \multicolumn{2}{c|}{\bf Texas100}    \\ \cline{2-7} 
                  & \multicolumn{1}{c|}{\bf Utility} & {\bf MIA Acc} & \multicolumn{1}{c|}{\bf Utility} & {\bf MIA Acc} & \multicolumn{1}{c|}{\bf Utility} & {\bf MIA Acc}  \\ \hline
               $\sigma^2=10$   & \multicolumn{1}{c|}{48\%} & 51\% & \multicolumn{1}{c|}{32\%} & 51\% & \multicolumn{1}{c|}{10\%} & 50\% \\ \hline
               $\sigma^2=1$   & \multicolumn{1}{c|}{59\%} & 56\% & \multicolumn{1}{c|}{54\%} &  56\% & \multicolumn{1}{c|}{26\%} & 54\% \\ \hline
               $\sigma^2=0.1$   & \multicolumn{1}{c|}{71\%} &  66\% & \multicolumn{1}{c|}{65\%} & 59\% & \multicolumn{1}{c|}{46\%} & 55\% \\ \hline
               $\sigma^2=0.01$   & \multicolumn{1}{c|}{78\%} & 79\% & \multicolumn{1}{c|}{78\%} & 64\% & \multicolumn{1}{c|}{49\%} & 60\% \\ \hline \hline 
              $\texttt{\name}$   & \multicolumn{1}{c|}{\bf 77\%} &  {\bf 51\%} & \multicolumn{1}{c|}{\bf 80\%} & {\bf 51\%} & \multicolumn{1}{c|}{\bf 46\%} & {\bf 50\%} \\ \hline
\end{tabular}
\label{tab:moreDPres}
 \vspace{-4mm}
\end{table}

\begin{table}[!t]\renewcommand{\arraystretch}{0.9}
\footnotesize
\caption{\texttt{\name} results against PIAs, where the attacker is unknown to the true (mean) aggregator and uses a substitute (max) one to aggregate the subset representations. 
} 
\addtolength{\tabcolsep}{-3.5pt}
\centering
\begin{minipage}{.32\linewidth}
\begin{tabular}{ccc}
\toprule
\multicolumn{3}{c}{\bf Census}\\
          \hline
\( \lambda \) & {\bf Utility} & {\bf PIA Acc} \\
\midrule
0     & 85\% & 48\%  \\
0.25  & 83\% & 45\% \\
0.5   & 82\% & 44\% \\
0.75  & 81\% & 28\% \\
1     & 60\% & 25\% \\
\bottomrule
\end{tabular}
\end{minipage}
\hfill
\begin{minipage}{.32\linewidth}
\begin{tabular}{ccc}
\toprule
\multicolumn{3}{c}{\bf RSNA}\\
          \hline
 {\bf Utility} & {\bf PIA Acc} \\
\midrule
0     & 83\% & 42\%  \\
0.25  & 82\% & 24\% \\
0.5   & 82\% & 21\% \\
0.75  & 81\% & 15\% \\
1     & 60\% & 14\% \\
\bottomrule
\end{tabular}
\end{minipage}
\hfill
\begin{minipage}{.32\linewidth}
\begin{tabular}{ccc}
\toprule
\multicolumn{3}{c}{\bf CelebA}\\
          \hline
{\bf Utility} & {\bf PIA Acc} \\
\midrule
0     & 91\% & 40\%  \\
0.25  & 91\% & 22\% \\
0.5   & 91\% & 15\% \\
0.75  & 88\% & 9\% \\
1     & 63\% & 9\% \\
\bottomrule
\end{tabular}
\end{minipage}
\label{tab:results_PIAs_substitute}
\vspace{-2mm}
\end{table}

\begin{table}[!t]
\vspace{-2mm}
\caption{\texttt{\name} results against DRAs with uniform perturbation distribution. A smaller SSIM or PSNR indicates better defense performance (\(\lambda\) =0.4). 
}
\centering
\small
\begin{tabular}{ccc}
\toprule
\multicolumn{3}{c}{\bf CIFAR10}\\
          \hline
 Scale $\epsilon$ & Utility & SSIM/PSNR\\
\midrule
0 & 89.5\% & 0.78 / 15.97\\ 
\hline
1.25 & 85.7\% & 0.28 / 13.23\\
2.25 & 77.7\% & 0.24 / 12.34\\
3.25 & 64.9\% & 0.20 / 12.93\\
\bottomrule
\end{tabular}
\label{tbl:more_results_DRAs}
\vspace{-4mm}
\end{table}

\section{Optimal Solution of \texttt{\name}}

\texttt{\name} involves training three (deep) neural networks whose loss functions are highly non-convex. To protect the data privacy, \texttt{\name} needs to solve a min-max adversary game, whose optimal solution could be  \emph{Nash Equilibrium}. 
However, finding Nash Equilibrium for highly non-convex functions can be a challenging problem, as traditional optimization techniques may not work effectively in such cases.

\end{document}